%% file: main.tex
\RequirePackage{fix-cm}

\documentclass[twocolumn]{svjour3}          
\smartqed  
\usepackage[authoryear]{natbib}

\usepackage{natbib}
\usepackage{amsmath}
\usepackage{multirow}
\usepackage{amssymb}
\usepackage{nccmath}
\usepackage{enumitem}
\usepackage{booktabs}
\usepackage{microtype} 
\usepackage{balance}    
\usepackage[pagebackref,breaklinks,colorlinks,allcolors=cvprblue]{hyperref}
\usepackage{xcolor}
\definecolor{cvprblue}{rgb}{0.21,0.49,0.74}
\makeatletter
\let\cl@chapter\undefined
\makeatother
\usepackage{graphicx}
\usepackage{caption}
\usepackage{subcaption}
\usepackage{cleveref}

\begin{document}
\newcommand{\methodName}{UIL-AQA }
\newcommand{\networkName}{\textit{Query-based transformer decoder }}
\newcommand{\problemName}{\textit{Temporal Skipping} }
\newcommand{\backbone}{\textit{Feature Extractor} }
\newcommand{\temporaldecoder}{\textit{Temporal Decoder} }
\newcommand{\regressionhead}{\textit{Difficulty-Qualtiy Regression Head} }

\title{Human–AI Divergence in Ego-Centric Action Recognition under Spatial and Spatiotemporal Manipulations}

\author{
    Sadegh Rahmaniboldaji\textsuperscript{1}\textsuperscript{*}\thanks{* Indicates these authors contributed equally.} \and
    Filip Rybansky\textsuperscript{2}\textsuperscript{*}\footnotemark[1] \and
    Quoc C. Vuong\textsuperscript{2} \and
    Anya C. Hurlbert\textsuperscript{2} \and
    Frank Guerin\textsuperscript{1} \and
    Andrew Gilbert\textsuperscript{1}
}

\institute{Sadegh Rahmani\textsuperscript{*} \at
              \email{s.rahmani@surrey.ac.uk}        
           \and
           Filip Rybansky\textsuperscript{*} \at
              \email{F.Rybansky2@newcastle.ac.uk}   
            \and
            Quoc C. Vuong \at
            \email{quoc.vuong@newcastle.ac.uk}  
            \and
            Anya C. Hurlbert \at
            \email{anya.hurlbert@newcastle.ac.uk}
            \and
            Frank Guerin \at
            \email{f.guerin@surrey.ac.uk}
            \and
            Andrew Gilbert \at
            \email{a.gilbert@surrey.ac.uk}
            \and
        \textsuperscript{1} University of Surrey, Guildford, UK
        \and
        \textsuperscript{2} Newcastle University, Newcastle upon Tyne, UK
}

\date{Received: date / Accepted: date}

\maketitle

\begin{abstract}

\input{sec/0_abstract}

\end{abstract}

\input{sec/1_intro}
\input{sec/2_Related_works}
\input{sec/3_Methodologies}
\input{sec/4_experiments}
\input{sec/5_conclusion}

\section*{Acknowledgements}
This work was funded by the Leverhulme Trust Research Project Grant RPG-2023-079.

\noindent The dataset used in this work, \textit{Epic-ReduAct}, is available 
\href{https://github.com/SadeghRahmaniB/Epic-ReduAct}{here}.

\noindent The authors declare no conflict of interest.

\bibliographystyle{apalike}
\bibliography{egbib}

\end{document}

%% file: sec/0_abstract.tex
Humans consistently outperform state-of-the-art AI models in action recognition, particularly in challenging real-world conditions involving low resolution, occlusion, and visual clutter. Understanding the sources of this performance gap is essential for developing more robust and human-aligned models. In this paper, we present a large-scale human–AI comparative study of egocentric action recognition using Minimal Identifiable Recognition Crops (MIRCs), defined as the smallest spatial or spatiotemporal regions sufficient for reliable human recognition. 
We used our previously introduced, Epic-ReduAct, a systematically spatially reduced and temporally scrambled dataset derived from 36 EPIC-KITCHENS videos, spanning multiple spatial reduction levels and temporal conditions. Recognition performance is evaluated using over 3,000 human participants and the Side4Video model. Our analysis combines quantitative metrics, Average Reduction Rate and Recognition Gap, with qualitative analyses of spatial (high-, mid-, and low-level visual features) and spatiotemporal factors, including a categorisation of actions into Low Temporal Actions (LTA) and High Temporal Actions (HTA).
Results show that human performance exhibits sharp declines when transitioning from MIRCs to sub-MIRCs, reflecting a strong reliance on sparse, semantically critical cues such as hand–object interactions. In contrast, the model degrades more gradually and often relies on contextual and mid- to low-level features, sometimes even exhibiting increased confidence under spatial reduction. Temporally, humans remain robust to scrambling when key spatial cues are preserved, whereas the model often shows insensitivity to temporal disruption, revealing class-dependent temporal sensitivities.
Overall, our findings demonstrate that strong benchmark performance can obscure fundamental human–AI misalignment and highlight limitations in current spatial and spatiotemporal representations, suggesting concrete directions for more efficient and human-aligned action recognition systems.

%% file: sec/1_intro.tex
\section{Introduction}
\label{sec:intro}

Ego-centric action recognition concerns the understanding of human actions from a first-person visual perspective and is fundamental to both human cognition and artificial intelligence (AI) models. It underpins a wide range of applications, including assistive technologies, augmented reality, human--computer interaction, and wearable AI. Effective ego-centric action understanding enables functionalities such as hands-free interaction, activity tracking, and robotic assistance.

In humans, action recognition is supported by two parallel visual processing streams: the ventral ``what'' and the dorsal ``where/how'' pathways \citep{Neural_mechanisms_for_the_recognition_of_biological_movements, GOODALE199220, SHMUELOF2005457}. These pathways are hierarchically specialised to process complementary information, including form and motion. Similarly, early artificial intelligence models such as convolutional neural networks (CNNs) \cite{726791, alexnet} adopt hierarchical architectures, although they do not explicitly disentangle different types of visual information. More recent models, including two-stream networks \cite{DBLP:journals/corr/SimonyanZ14}, long short-term memory networks (LSTMs) \cite{hochreiter1997long}, and vision transformers (ViTs) \cite{vit}, aim to integrate spatial and temporal cues in a manner loosely inspired by human action recognition. However, despite these advances, such models remain vulnerable to occlusions and generally lack predictive processing mechanisms that enable humans to infer missing or ambiguous visual information \cite{Harmonizing_the_object_recognition_strategies_of_deep_neural_networks_with_humans}.

Studies of hierarchically organised layered AI architectures have demonstrated that activations within these layers can predict neural responses in early \cite{10005}, intermediate \cite{doi:10.1073/pnas.1403112111}, and late \cite{Yamins2016} areas of both monkey and human visual cortex. On this basis, some researchers argue that the combination of high task performance and alignment with behavioural and neural responses renders these models plausible approximations of human visual recognition \cite{DBLP:journals/corr/Lindsay15, 1231244}. Others, however, contend that their correspondence with human behaviour remains only slightly above chance \cite{DBLP:journals/corr/abs-2106-07411, DBLP:journals/corr/abs-2006-16736}, raising questions about whether current models rely on the same information and strategies as humans.

Despite similarities in hierarchical architecture, the features extracted at each layer by human vision and AI models can differ substantially. For instance, in static visual recognition tasks,  humans initially process boundaries and surfaces, which can result in a shape bias \cite{doi:10.1126/science.3283936}, whereas AI models often exhibit a texture bias \cite{DBLP:journals/corr/abs-1811-12231, DBLP:journals/corr/abs-2004-07780, Malhotra2022}. Additionally, AI systems frequently fail to encode the global configuration of visual elements \cite{Baker2018}, the three-dimensional and internal structure of objects, and cues related to occlusion and depth \cite{8954212, Jacob2021, dong2022viewfoolevaluatingrobustnessvisual, 1231244, 031739}. Nonetheless, with specialised training datasets or architectural modifications, AI models can be trained to extract features that more closely resemble those utilised by humans \cite{dear, DBLP:journals/corr/abs-1811-12231, jang2021noise}.

Although both human vision and artificial models rely on hierarchical representations, humans uniquely integrate bottom-up sensory signals with top-down cognitive processes to achieve dynamic perception \cite{AHISSAR2004457}. Human perception is continually modulated by prior knowledge during inference \cite{089892903321662976, HOCHSTEIN2002791}, whereas deep learning models primarily depend on labelled data during training. Furthermore, human attention inherently prioritises salient regions based on contextual relevance \cite{torralba2006contextual}. This suggests that humans and current AI models may differ not only in the features they encode, but also in how they allocate and integrate spatial and temporal information when recognising actions.

Motivated by these findings and differences, we previously studied human performance in action recognition in \cite{Rybansky2026.02.15.705896} and conducted a range of quantitative analyses of spatial features in \cite{humanvsmachine}. In this work, we build on and extend these contributions by further investigating the roles of spatial and spatiotemporal features in action recognition. We examine the differences between human observers and AI models in recognising actions in challenging ego-centric scenarios. Ego-centric videos provide detailed information about objects within an actor's immediate affordance \cite{BORGHI201264, roche2013visually} and their interactions via the hands \cite{campanella2011visual}, while often offering limited contextual information \cite{HENDERSON200840}. This reduced context can affect both the visual features \cite{libby2011self, mcisaac2002vantage} and cognitive strategies \cite{10.1162/jocn_a_00195, rizzolatti2001neurophysiological} employed during action recognition, making ego-centric video an ideal domain for comparative analysis. We utilise videos from the standard ego-centric kitchen dataset EPIC-KITCHENS-100 \cite{EPICKITCHENS} and systematically reduce available spatial and temporal information to extend the Epic-ReduAct dataset\footnote{Epic-ReduAct is available \href{https://github.com/SadeghRahmaniB/Epic-ReduAct}{here}.} (Epic-Kitchens Reduced Action Videos) \cite{humanvsmachine, Rybansky2026.02.15.705896}, which is publicly released. 

To quantify human and AI performance, and inspired by prior work on static images \cite{A_model_for_full_local_image_interpretation} and third-person actions \cite{Minimal_videos_Trade-off_between_spatial_and_temporal_information_in_human_and_machine_vision}, we adopt the concept of Minimal Recognisable Configurations (MIRCs), defined as the smallest spatial crops of a video that remain identifiable by humans. Further spatial reduction yields smaller sub-quadrants, which, when unrecognisable, are referred to as sub-MIRCs, \cite{humanvsmachine, Rybansky2026.02.15.705896}. In parallel, to explore the spatiotemporal constraints of human and AI models, MIRCs are temporally scrambled. For simplicity and to ensure comparison to their spatial equivalents, we also refer to these as sub-MIRCs, regardless of their recognisability.

Using Epic-ReduAct and the MIRC framework, we address the central question of \emph{which spatial and spatiotemporal cues are minimally sufficient for humans and state-of-the-art models to recognise ego-centric actions}. We quantify performance disparities between humans and AI models using two complementary metrics: the Average Reduction Rate, which we introduced \cite{humanvsmachine, Rybansky2026.02.15.705896}, and the Recognition Gap, adapted from \cite{A_model_for_full_local_image_interpretation}. Beyond these quantitative measures, we conduct extensive qualitative analysis to examine feature variations across reduction levels and to introduce a novel categorisation of motion patterns. Together, these analyses provide insight into the key spatial and spatiotemporal differences between human and model-based action recognition and highlight concrete directions for improving AI models.

The main contributions of this paper are, therefore:
\begin{enumerate}
    \item Proposing A diagnostic framework and benchmark for minimal spatial and spatiotemporal information in ego‑centric action recognition, combining hierarchical spatial reduction, temporal scrambling, Epic‑ReduAct, and two quantitative metrics that support systematic human–AI comparison under degraded visual conditions.

    \item An actionable, qualitative, strategy-level spatial analysis establishing design principles for efficient and human-aligned video models. We introduce a spatial-efficiency guideline grounded in high- and mid-level feature analysis, showing that humans rely on sparse, semantically critical active object configurations, whereas a state-of-the-art video model depends primarily on distributed contextual objects and mid-level statistics, resulting in distinct failure and recovery patterns.

    \item A qualitative, strategy-level spatiotemporal analysis distinguishing Low vs. High Temporal Actions, together with recommendations for shifting model reliance from incidental context toward behaviourally critical objects and interactions.

\end{enumerate}

%% file: sec/2_Related_works.tex
\section{Related Works}
\label{sec:related_works}

\subsection{Human and Computer Vision}

Human action recognition is supported by two parallel neural processing streams, commonly referred to as the ventral and dorsal pathways \cite{Neural_mechanisms_for_the_recognition_of_biological_movements, GOODALE199220, SHMUELOF2005457}. These pathways originate in the primary visual cortex (V1) and project to higher-level regions, including inferotemporal (IT) and prefrontal cortex (PFC) \cite{A_feedforward_architecture_accounts_for_rapid_categorization}. Early stages of visual processing extract low-level features such as edges, colour, and motion \cite{HubelD.H.1959Rfos, Roe2004, Roe1995}, which are subsequently integrated into increasingly complex spatial and spatiotemporal representations. According to the reverse hierarchy theory \cite{AHISSAR2004457}, rapid bottom-up categorisation is followed by top-down refinement, enabling precise and flexible action recognition under variable visibility and partial information \cite{089892903321662976, doi:10.1073/pnas.0507062103, HOCHSTEIN2002791}. Motivated by these biological principles, recent AI models that incorporate motion-aware or attention-based mechanisms have achieved state-of-the-art performance in video action classification \cite{mofo}.

In contrast, AI-based action recognition predominantly relies on deep learning models trained on large-scale datasets. Early approaches employed convolutional neural networks such as AlexNet \cite{alexnet}, VGG \cite{vgg}, and ResNet \cite{resnet}, which were originally designed for static image analysis and therefore operate primarily on spatial information. Although their layered architectures were loosely inspired by human vision, their effectiveness for action recognition was limited, in part because temporal dynamics had to be approximated by stacking frames or using simple temporal pooling. More recent advances, including Transformers \cite{transformer}, Vision Transformers (ViTs) \cite{vit}, and contrastive learning frameworks such as CLIP \cite{clip}, have substantially improved visual representation learning. These developments have been extended to video-specific architectures that more explicitly model motion, such as Side4Video \cite{side4video} and MOFO \cite{mofo}, which leverage motion-based self-supervised learning, as well as FILS \cite{ahmadian2024filsselfsupervisedvideofeature}, which incorporates linguistic semantics into video representations. 

With the integration of multimodal data, recent large language models (LLMs) and video--language models (VLMs), including VideoLLaMA3 \cite{videollama3} and VideoChatGPT \cite{VideoChatGPT}, enable richer and more holistic scene understanding by combining spatial appearance, temporal evolution, and semantic context. However, despite these advances, relatively few works have systematically probed how such models trade off spatial detail and temporal information under explicit reductions or scrambling. Comparative analyses of human and computer vision offer valuable insights into both improving AI models and understanding human perception. While humans demonstrate strong generalisation and contextual reasoning abilities, particularly under degraded or incomplete spatial and temporal information, AI systems typically achieve high specific task accuracy but often lack adaptability. Bridging these gaps through biologically inspired architectures and controlled studies of spatial and spatiotemporal robustness represents a promising direction toward more robust artificial vision systems.

\subsection{Biologically inspired Vision Models}

Early neurocomputational models were strongly influenced by the hierarchical organisation of the human visual pathways. A seminal example is the HMAX model \cite{hmax, A_Biologically_Inspired_System_for_Action_Recognition, A_feedforward_architecture_accounts_for_rapid_categorization}, which processes retinal input through a neurophysiologically grounded hierarchy of alternating simple and complex cell layers, employing linear and nonlinear MAX-pooling operations. Giese et al. \cite{Neural_mechanisms_for_the_recognition_of_biological_movements} extended this framework to action recognition, such as walking and running, by introducing a two-pathway feedforward architecture inspired by the human ventral and dorsal streams, explicitly modelling form and motion pathways. More recent work has sought to enhance HMAX-style models while remaining within biological constraints. These efforts include incorporating layers corresponding to a more complete anatomy of the visual cortex \cite{10.5555/1195185}, modifying the sequence of information flow and weights according to context \cite{Bio_inspired_unsupervised_learning_of_visual_features_leads, Learning_simple_and_complex_cells_like_receptive_fields_from_natural}, or exploiting the temporal coordination of neural activity \cite{Mély2017}. 

These models emphasise the importance of hierarchical, parallel processing of spatial structure and temporal dynamics, and they provide valuable insights into biological mechanisms of visual recognition. However, they generally fall short of the performance achieved by modern deep learning--based vision systems described earlier, and only a small subset of this literature quantitatively examines how such models respond when spatial information is reduced or temporal structure is disrupted. Our work complements these biologically inspired approaches by using a modern video model and focusing on controlled manipulations of spatial and spatiotemporal information in natural ego-centric actions.

\subsection{Joint AI and Human Studies}

Research explicitly integrating human-inspired principles into computer vision has examined the role of informative image regions in improving recognition, particularly under partial occlusion or reduced spatial detail \cite{On_the_Minimal_Recognizable_Image_Patch}, as well as methods for aligning neural network receptive fields more closely with human visual processing \cite{Harmonizing_the_object_recognition_strategies_of_deep_neural_networks_with_humans}. Several studies have argued for aligning AI model behaviour with human perception, such that image manipulations imperceptible to humans should similarly have minimal impact on model predictions \cite{If_a_Human_Can_See_It_So_Should_Your_System_Reliability_Requirements_for_Machine_Vision_Component}. 

Comparative analyses have consistently demonstrated discrepancies between human observers and AI models in terms of visual attention and recognition performance on image classification tasks \cite{Laconic_Image_Classification_Human_vs_Machine_Performance, Minimal_Images_in_Deep_Neural_Networks_Fragile_object_recognition_in_Natural_Images, Oculo-retinal_dynamics_can_explain_the_perception_of_minimal_recognizable_configuration, What_Takes_the_Brain_so_Long_Object_Recognition_at_the_Level_of_Minimal_Images_Develops_for_up, Do_Humans_and_Convolutional_Neural_Networks_Attend_to_Similar_Areas_during_Scene_Classification, Seeing_Eye-to-Eye_A_Comparison_of_Object_Recognition_Performance}. Notably, Ben-Yosef et al. \cite{A_model_for_full_local_image_interpretation} introduced the concept of Minimal Recognisable Configurations (MIRCs) and sub-MIRCs for static images, demonstrating that human recognition accuracy drops sharply once critical spatial features are removed, whereas AI models exhibit a more gradual degradation.

Extending this framework to video, \cite{Minimal_videos_Trade-off_between_spatial_and_temporal_information_in_human_and_machine_vision} identified the minimal spatial and temporal configurations required for human action recognition in third-person videos from the UCF101 dataset \cite{ucf101}. Their findings showed a substantially steeper recognition decline from MIRCs to spatial sub-MIRCs for humans than for AI models, while no significant human--model differences were observed for temporally downsampled sub-MIRCs. This suggests that humans may be more sensitive than current models to the removal of critical spatial structure, but not necessarily to moderate temporal reduction in the third-person setting considered.
\begin{figure*}[htbp]
  \centering
  \includegraphics[width=\textwidth]{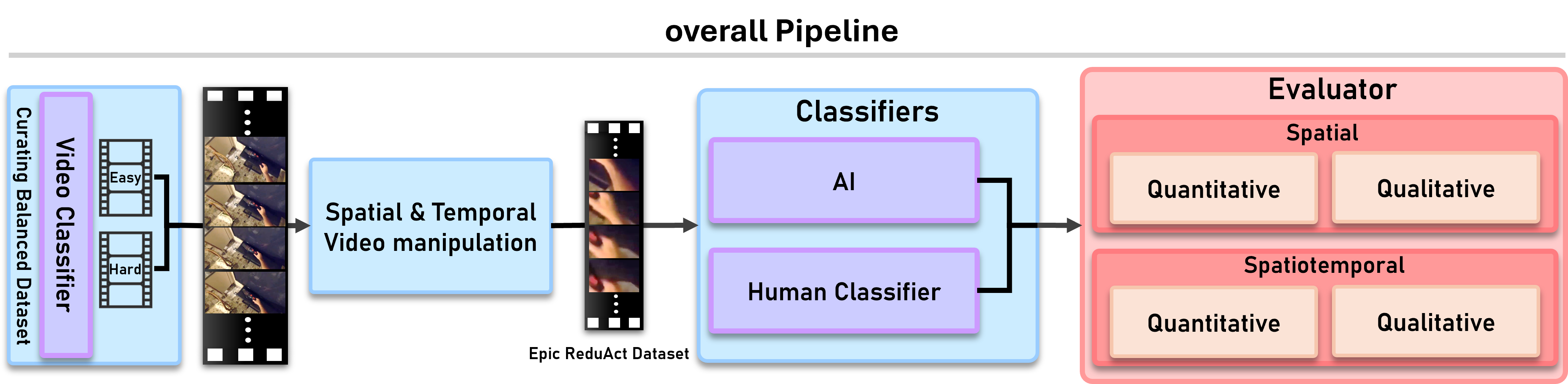}
  \caption{Research pipeline highlighting different processes, including data preparation and classification by human and AI classifiers and the final analysis stage.}
  \label{fig:research_pipeline}
\end{figure*}

Building on this comparative paradigm, our work focuses on ego-centric video, a domain that has received limited attention in this context, and examines a broader and more diverse set of ego-centric activities. We systematically manipulate both spatial information (through quadrant and sub-quadrant cropping) and spatiotemporal information (through temporal scrambling of MIRCs) to characterise the minimal configurations that remain recognisable to humans and to a state-of-the-art video model. In doing so, we provide a detailed comparison of how humans and AI systems trade off spatial versus temporal cues when recognising actions from natural first-person videos.

\begin{table*}[t]
  \centering
  \footnotesize
  \setlength{\tabcolsep}{4pt}
  \begin{tabular}{lccccc|p{4.9cm}}
    \toprule
    Category & Videos & Samples & MIRCs & Spatial sub-MIRCs & Spatiotemporal quadrants & Verb Classes \\
    \midrule
    Easy & 18 & 4{,}503 & 273 & 1{,}092 & 273 (200) &
    close, cut, hang, open, pour, put, remove, take, turn-off, turn-on, wash \\
    \midrule
    Hard & 18 & 3{,}173 & 402 & 804 & 201 (145) &
    close, hang, insert, open, peel, pour, put, remove, serve, take, turn-off, wash \\
    \bottomrule
  \end{tabular}
  \caption{Summary of the Epic-ReduAct dataset. Values in parentheses indicate spatiotemporal unrecognisable quadrants.}
  \label{tab:dataset}
\end{table*}

%% file: sec/3_Methodologies.tex
\section{Methodology}
\label{sec:Methodologies}

Our research pipeline, shown in \Cref{fig:research_pipeline}, summarises the methodology used to compare human and AI performance in ego-centric video action recognition. We first apply a classifier to preselect videos into Easy and Hard subsets, thereby constructing a representative and balanced dataset spanning multiple difficulty levels. To facilitate a controlled comparison of recognition behaviour, we then systematically reduce spatial information or scramble video frames to disrupt temporal structure. Both human participants and an AI model are subsequently evaluated on these spatially reduced or temporally scrambled clips, enabling direct comparison of recognition performance and quantification of disparities between human and AI action recognition.

\subsection{Problem Definition and Epic-ReduAct Preparation}

To support our investigation, we constructed a balanced subset of the EPIC-KITCHENS-100 dataset \cite{EPICKITCHENS}, comprising \emph{Easy} and \emph{Hard} videos that represent distinct levels of difficulty for action recognition in AI models. Each subset consists of 18 videos with an average duration of 2.35 seconds (SD = 1.11 seconds), enabling controlled comparisons of human and AI recognition performance across varying difficulty levels.

The Easy and Hard subsets of Epic-ReduAct were derived using a state-of-the-art action recognition model \cite{mofo} to estimate class prediction probabilities. Videos with top-1 confidence scores above 60\% were selected as candidates for the Easy subset. In contrast, videos in which the ground-truth label did not appear within the top-5 predictions were designated as Hard candidates. A subsequent manual curation step removed ambiguous or visually redundant samples to ensure clear separation between difficulty levels. The resulting Epic-ReduAct dataset comprises actions across 11 verb classes in the Easy subset and 12 in the Hard subset.

\begin{figure}[htbp]
  \centering
  \includegraphics[width=1\linewidth]{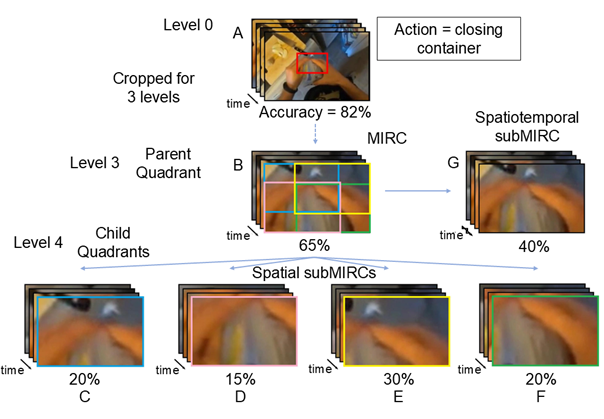}
  \caption{Illustration of the reduction process. The ground-truth action label for the video is \emph{close}. Panel~A undergoes three successive levels of spatial reduction, resulting in Quadrant~B, which is correctly recognised by 65\% of participants. Quadrant~B serves as the parent quadrant for four spatially reduced child quadrants, Upper-Left, Bottom-Left, Upper-Right, and Bottom-Right (C--F), each recognised by fewer than 50\% of participants, thereby classifying them as sub-MIRCs and Quadrant~B as the corresponding MIRC. In the spatiotemporal branch, the MIRC video is temporally scrambled, producing an additional child video (G), a spatiotemporal sub-MIRC, with a recognition rate of 40\%.}
  \label{fig:reducing_video}
\end{figure}

Next, we conducted online experiments in which the spatial information of the 18 Easy and 18 Hard videos (36 in total) was systematically reduced across eight hierarchical levels to identify Minimal Recognisable Configurations (MIRCs). The reduction procedure is illustrated in \Cref{fig:reducing_video} using an example video with the ground-truth label \emph{close}. At Level~0, each video was spatially cropped to a region that best preserved the full extent of the action. At Level~1, each video was cropped into corner-centred quadrants, yielding four child sub-videos per parent: Upper-Left, Bottom-Left, Upper-Right, and Bottom-Right. Levels~2 through~7 were generated by recursively applying the same corner-cropping procedure to each parent video at the preceding level (see \Cref{fig:reducing_video}).

Video labels for MIRCs and sub-MIRCs \cite{A_model_for_full_local_image_interpretation} were assigned based on human recognition performance. Following the spatial division of a parent video into four quadrants, any quadrant recognised by at least 50\% of human participants was selected for further recursive reduction. If none of the child quadrants exceeded this recognition threshold, the parent video was designated as a \textbf{MIRC}, and its unrecognisable child quadrants were labelled as \textbf{sub-MIRCs}.

Due to the exponential growth in the number of child quadrants at deeper reduction levels, we applied a efficient quadrant selection via overlap to restrict testing to the most informative candidates. This pruning substantially reduced experimental complexity while preserving the ability to reliably identify MIRCs. The procedure was applied as follows:
\begin{enumerate}
    \item After evaluating all quadrants at a given reduction level (see \Cref{method_human}), child quadrants were generated only from parent quadrants that were recognised by at least 50\% of participants.
    
    \item Child quadrants whose spatial extent was fully contained within a parent quadrant that had been unrecognised at any previous level were assumed to be unrecognisable and were therefore excluded from further testing.
    
    \item If child quadrants from a single level overlapped each other by at least 95\%, we assumed that they had equal accuracy and tested only one quadrant from each such cluster.
    
    \item Within each level's remaining quadrants, we selected child quadrants that were (i) least likely to be recognised (sharing $\ge 65\%$ area with any unrecognised quadrant from previous levels) and (ii) most informative (showing the greatest cumulative overlap in surface area with other child quadrants at the same level). We continued this process until we reached the maximum number of quadrants per video that could be evaluated efficiently at the specified level.
\end{enumerate}

Following the identification of spatial MIRCs through hierarchical cropping, we extended the reduction paradigm to the temporal domain to examine the role of temporal structure in ego-centric action recognition. Specifically, we tested all MIRCs identified in the spatial reduction phase by selectively disrupting their temporal coherence while preserving spatial content. This allowed us to distinguish purely spatial MIRCs from \emph{spatiotemporal} MIRCs, i.e., configurations for which both spatial and temporal information are critical for recognition.

\begin{figure*}[htbp]
  \centering
  \includegraphics[width=\textwidth]{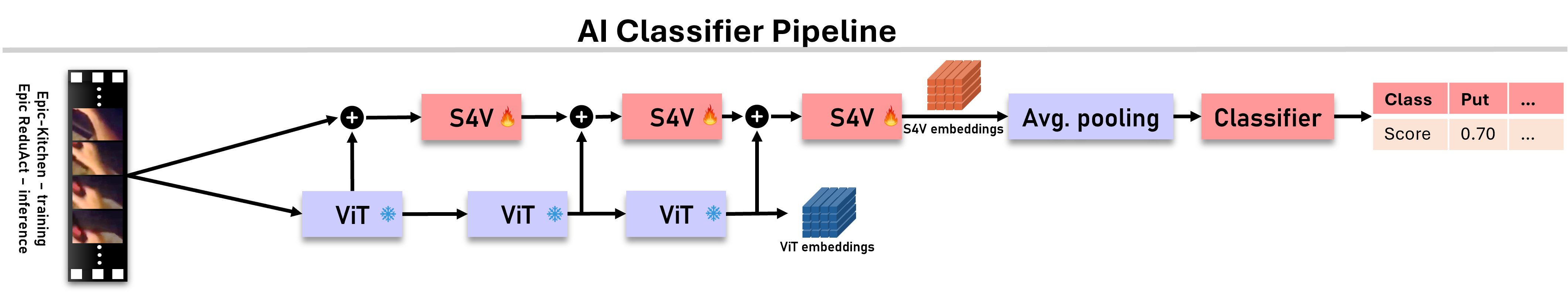}
  \caption{The AI classifier (see \Cref{fig:research_pipeline}) is described in greater detail, consisting of a video feature encoder augmented by a lightweight spatiotemporal side network integrated with a frozen, pre-trained vision backbone.}
  \label{fig:ai_research_pipeline}
\end{figure*}

To this end, each spatial MIRC video was temporally scrambled using a block-wise procedure designed to avoid excessive or unnatural discontinuities \cite{VUONG20041717_Rotation_direction_affects_object}. Each video was divided into five contiguous temporal blocks of approximately equal duration. The order of these blocks was then randomised under three constraints: (i) the first block could not remain in its original position, (ii) the last block was reassigned to a middle position, and (iii) no two blocks that were adjacent in the original video remained adjacent after scrambling. This manipulation preserved local motion statistics within blocks while disrupting the global temporal structure. 


Recognition of temporally scrambled videos was evaluated using the same experimental protocol and decision criteria as in the spatial reduction phase. If scrambling caused recognition performance to fall below the 50\% human recognition threshold, the configuration was considered unrecognisable. For simplicity, and to facilitate comparison with their spatial counterparts, we refer to these cases as spatiotemporal sub-MIRC, regardless of their recognisability status. Their corresponding unscrambled versions are classified as spatiotemporal MIRC (see \Cref{fig:reducing_video}). Across all spatiotemporal MIRC configurations, 345 became unrecognisable, corresponding to 72.78\% of the tested cases.

A single cohort of participants was recruited for this phase. Because the number of MIRCs varied across videos, we constructed multiple stimulus sets that were matched as closely as possible with respect to difficulty level (Easy/Hard), action category, and number of videos. Recognition testing was conducted in the same manner as in Phase~1, enabling direct comparison of spatial and spatiotemporal reduction effects.

All reductions, temporal scrambling, and pruning steps resulted in a total of 8{,}151 quadrant videos, including the Level~0 clips. The composition of the final Epic-ReduAct dataset is summarised in \Cref{tab:dataset}.

\subsection{Classifiers}

Using the spatially reduced and temporally scrambled videos, we evaluated action recognition using two classifiers: human observers, whose responses were also used to identify MIRCs and sub-MIRCs, and an AI model \cite{side4video}. Before describing these classifiers in detail, we first clarify how recognition accuracy is defined for each. For the AI model, accuracy corresponds to the confidence score assigned to the predicted verb relative to all other verb classes, expressed as a value between 0 and 1 (i.e., the softmax output). For human participants, accuracy is defined as the proportion of observers who correctly labelled the action shown in the video.

\input{sec/3_2_1_computational_classifiers}
\input{sec/3_2_2_human_classifiers}

%% file: sec/3_2_1_computational_classifiers.tex
\subsubsection{AI Classifiers}
\label{method_ai}

The AI classifier used in this study is based on the Side4Video (S4V) framework \cite{side4video}. This choice ensures that the balanced dataset of Easy and Hard videos is not biased by the MOFO-based selection process \cite{mofo}, which relies on a fundamentally different, unsupervised training paradigm. Although MOFO and S4V achieve comparable performance, S4V offers substantially faster training, making it a more practical choice for integration into our pipeline.

As illustrated in the upper portion of \Cref{fig:ai_research_pipeline}, S4V employs the OpenCLIP vision module \cite{clip} as a frozen, pre-trained video feature encoder, with a lightweight spatiotemporal side network attached. The trainable components operate in parallel with the frozen layers rather than being inserted sequentially, enabling efficient fine-tuning without back-propagation through the full model. This design leverages multi-level spatial features from the image encoder while significantly reducing memory usage (by up to 75\% compared to prior adapter-based approaches) and allows large models such as ViT-E (4.4B parameters) to be effectively adapted for video understanding.

For our experiments, the model was trained on the EPIC-KITCHENS-100 training split comprising 67{,}179 videos \cite{EPICKITCHENS}, excluding the 36 videos reserved for human--AI comparison. Each video clip was uniformly subsampled to 8 frames at a spatial resolution of $224 \times 224$. The model uses a ViT-B/16 backbone \cite{vit} and was trained using the AdamW optimiser \cite{adam}. The trained classifier was then applied to the spatially reduced and temporally scrambled videos in Epic-ReduAct to evaluate recognition performance against the ground-truth verb labels of the original videos, with results reported in \Cref{sec:experiments}.

During inference, the evaluation focuses on recognising ongoing action rather than identifying specific objects. Accordingly, verb recognition accuracy is computed by aggregating confidence scores across all correctly predicted verbs for each action label and comparing these aggregated scores with human recognition rates on the same spatially reduced and spatiotemporally scrambled clips.

%% file: sec/3_2_2_human_classifiers.tex
\subsubsection{Human Classifiers}
\label{method_human}

\begin{figure*}[htbp]
  \centering
  \includegraphics[width=\textwidth]{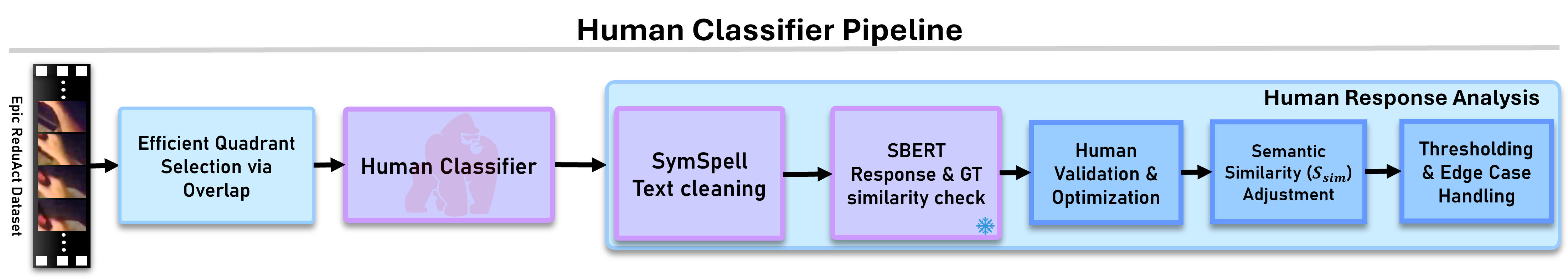}
  \caption{The human classifier (see \Cref{fig:research_pipeline}) is presented in more detail, comprising two stages: classification and response cleaning.}
  \label{fig:human_research_pipeline}
\end{figure*}

The human classifier in this work is based on the experimental setup introduced in \cite{Rybansky2026.02.15.705896}. Human recognition experiments were conducted using the Gorilla online platform \cite{Anwyl-Irvine2020} and were distributed via Prolific (www.prolific.com). In total, data were collected from 4,360 participants (1,964 female, 2,329 male, and 67 identifying as other or not stated; mean age = 33.2 years, (SD = 11.3 years). Depending on the size of the assigned stimulus set, median completion times ranged from 9 to 21 minutes, with participant compensation ranging from £0.70 to £2.75, ensuring fair reimbursement. Ethical approval was obtained from the University Research Ethics Committee (Ref: 38465/2023). 

The process of humans classifying a video is shown in \Cref{fig:human_research_pipeline}. Each experiment began with five practice trials. Participants were then presented with the balanced Easy/Hard dataset of 36 videos, and recognition accuracy was measured across trials. To control task difficulty and reduce learning effects, each participant viewed only one child quadrant per video, with trials presented in random order and interspersed with two catch trials. Practice and catch trials were drawn from additional Easy videos in the EPIC-KITCHENS-100 dataset \cite{EPICKITCHENS}. Seventeen participants were excluded and replaced due to poor performance on the catch trials.

Each trial began with a central fixation cross displayed for 500 ms, followed by a video presented on a white background until a response was made. After 4000 ms, participants were prompted to type their response describing the observed action. Free-text responses were used to avoid bias introduced by predefined labels. Participants were instructed to identify a single action and the object being acted upon, using up to three words.

Human response analysis was based on 
an optimised Semantic Similarity measure, $S_{\mathrm{sim}}$, relative to a human-GT label determined by \cite{Rybansky2026.02.15.705896}. Responses were first cleaned by removing punctuation, articles and generic subjects (e.g. “man”, “person”), correcting misspellings algorithmically using SymSpell \cite{Garbe2012}, and manually rewording responses with incorrect word counts or slang. 

The cosine similarity, $CS$, between each cleaned response and corresponding human-GT label was computed using the sentence-BERT (SBERT) model (\textit{all-mpnet-base-v2}) \citep{abs-1908-10084}, and, in addition,  the similarity between isolated verb terms (action), $CS_A$, and object terms, $CS_O$, was computed using Word2Vec embeddings \cite{MikolovSCCD13}. The final Semantic Similarity $S_{\mathrm{sim}}$ score was computed as:
$$
    S_{\mathrm{sim}} = CS - \bigl(CS_O \cdot p\bigr)^2 + \bigl(CS_A \cdot b\bigr)^2,
$$
where $p$ and $b$ denote penalty and bonus constants, respectively, which were optimised in advance on a separate validation set. A response was classified as correct if its $S_{\mathrm{sim}}$ score exceeded a predefined threshold calibrated to match manual labelling, and human recognition accuracy for each video was defined as the proportion of participants whose responses were classified as correct \cite{Rybansky2026.02.15.705896}.

%% file: sec/4_experiments.tex
\section{Experiments and Results}
\label{sec:experiments}

In this section, we present a comprehensive set of experiments investigating the impact of feature manipulation in both spatial and spatiotemporal dimensions on human and AI action recognition. We use two quantitative metrics, the \emph{Recognition Gap} and the \emph{Average Reduction Rate}, to characterise how performance changes as information is progressively removed. We further analyse which visual features persist or disappear across successive reduction levels, including temporal manipulations, to identify the features that induce transitions in model predictions. Together, these analyses enable a systematic study of how spatial and temporal cues affect both human observers and the AI classifier, and highlight features critical for recognition through complementary qualitative analyses.

\subsection{Quantitative Evaluation Metrics}

We employ two complementary metrics to analyse the effects of spatial and spatiotemporal manipulations on action recognition performance: \emph{Recognition Gap} and \emph{Average Reduction Rate}. Both metrics are defined to enable direct comparison between human and AI recognition behaviour.

\subsubsection{Recognition Gap Metric.}
The Recognition Gap quantifies how recognition performance changes between two consecutive levels of spatial reduction \cite{A_model_for_full_local_image_interpretation, humanvsmachine, Rybansky2026.02.15.705896} or spatiotemporal scrambling \cite{Rybansky2026.02.15.705896}, enabling a direct comparison between human observers and the AI classifier under matched baseline conditions. For human observers, the recognition gap is defined as the difference in average classification accuracy between parent MIRCs and their corresponding sub-MIRCs. For the AI model, we define the recognition gap relative to human performance using a class-wise matched operating point. Specifically, for each action class, we first compute the average human recognition accuracy on MIRC videos, denoted by $X \in [0,1]$. This value serves as a reference level for the task difficulty in that class. To compare the AI model at an equivalent operating point, we determine a class-specific confidence threshold $tl$ such that the proportion of MIRC samples with predicted confidence exceeding $tl$ equals $X$. In other words, $tl$ is chosen to ensure that the AI model achieves the same effective recognition rate as humans at the MIRC level. This procedure avoids reliance on absolute confidence calibration and enables a fair, class-wise comparison between human and AI behaviour. The threshold is illustrated as the green dotted line in \Cref{fig:recog_gap_per_class}.  We then calculate the average confidence gap between MIRCs equal or above the threshold and their corresponding sub-MIRCs. As an illustrative example, in \Cref{fig:recog_gap_per_class-a}, the average human accuracy for MIRCs is $59\%$. The corresponding AI threshold is defined as the confidence value above which $59\%$ of the model’s MIRC predictions lie (e.g., a confidence of $0.15$). The average gap for MIRCs above this threshold is $-1.05\%$, indicating a slight improvement under spatial reduction for that class. This small change in the recognition gap arises because the metric first accounts for the AI model’s performance relative to human accuracy. Lower values indicate that the AI does not substantially outperform humans. Nevertheless, this metric allows us to identify improvements under reduction conditions, which in turn motivates further qualitative analysis.



By aligning AI evaluation to the human operating point on the parent videos, this metric isolates the effect of information reduction itself, allowing us to measure whether reduced spatial or temporally scrambled input disproportionately harms or benefits the AI model relative to human observers.

\subsubsection{Average Reduction Rate Metric}
The Average Reduction Rate quantifies the reduction in recognition performance between two consecutive levels of spatial or spatiotemporal manipulation. For each parent–child pair, we compute the signed change in recognition accuracy $\Delta a$ as
$$
\Delta a = a_{\text{parent}} - a_{\text{child}},
$$
where $a_{\text{parent}}$ and $a_{\text{child}}$ denote recognition accuracy in the range $[0,1]$ for the parent and child, respectively.

To characterise the severity of recognition failures, we report the average reduction rate (ARR) over degrading transitions only, defined as the mean of positive accuracy drops, i.e. cases where recognition performance decreases under further reduction:
$$
\mathrm{ARR} = \mathbb{E}\bigl[\Delta a \;\big|\; \Delta a > 0\bigr].
$$
This focuses the analysis on the magnitude and localisation of performance collapse as information is progressively removed, rather than conflating degradation with recovery.

\subsection{Qualitative Evaluation Metrics}
\label{subsec:feature_extraction}

To qualitatively analyse the key features supporting recognition under spatial reduction, we extract two complementary feature sets: high-level object-based features and mid-level visual features. These features were specifically chosen to capture systematic changes in semantically and perceptually salient visual properties that are likely to affect recognition performance under progressive spatial reduction.

\begin{figure}[htbp]
  \centering
  \begin{subfigure}{0.9\linewidth}
    \centering
    \includegraphics[width=\linewidth]{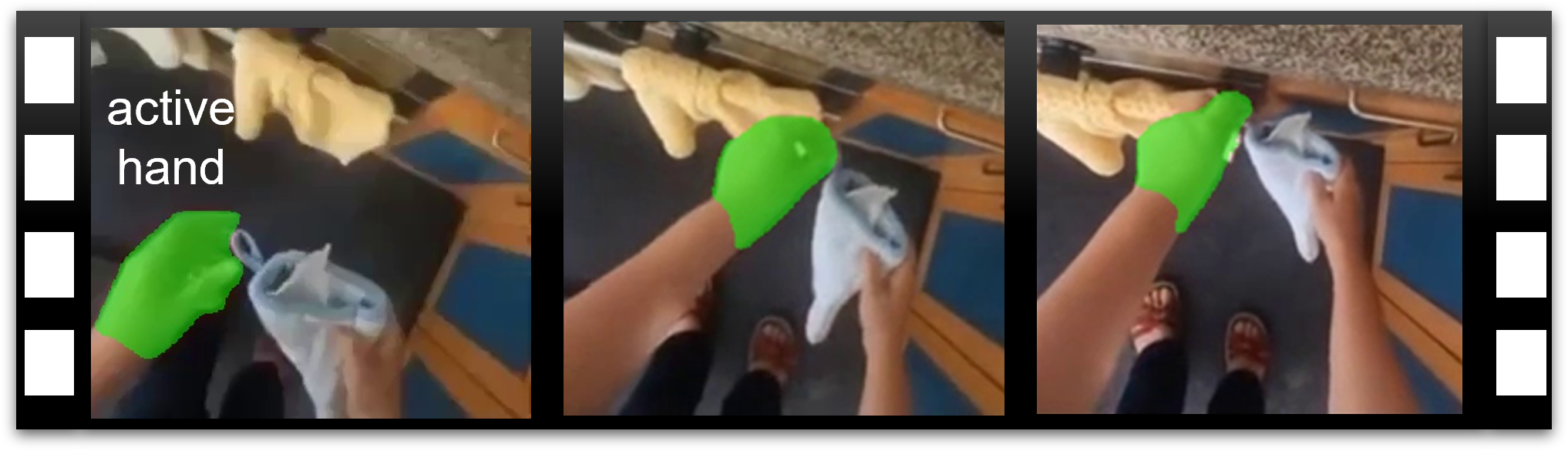}
    \caption{Segmented Active Hand}
    \label{fig:high_level_features-a}
  \end{subfigure}

  \vspace{0.8em}

  \begin{subfigure}{0.9\linewidth}
    \centering
    \includegraphics[width=\linewidth]{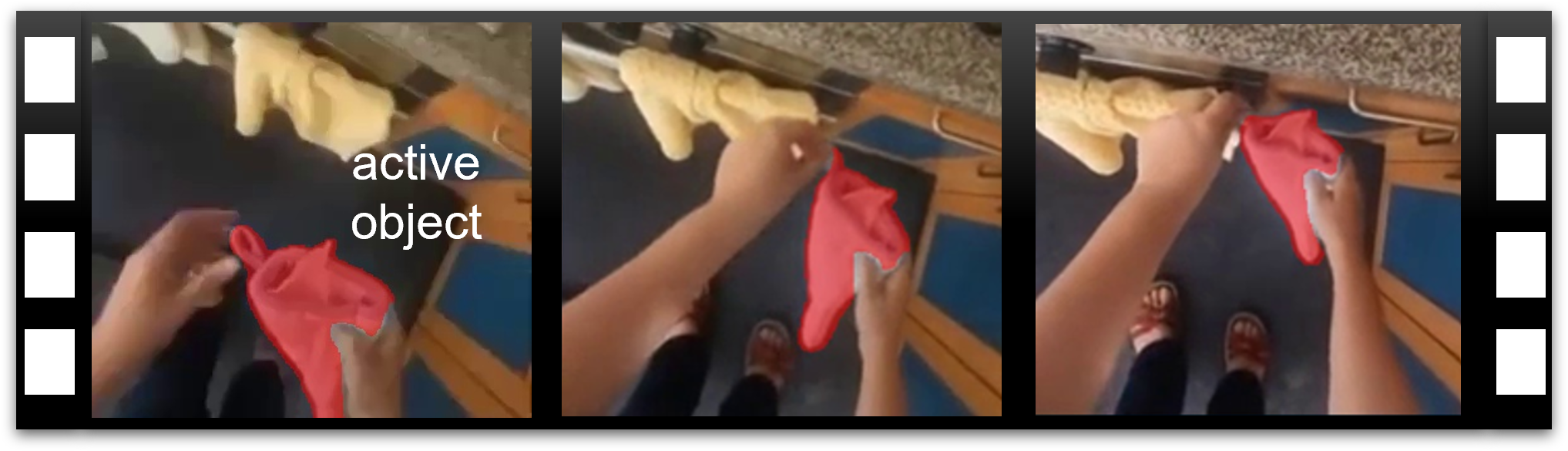}
    \caption{Segmented Active Object}
    \label{fig:high_level_features-b}
  \end{subfigure}

  \vspace{0.8em}

  \begin{subfigure}{0.9\linewidth}
    \centering
    \includegraphics[width=\linewidth]{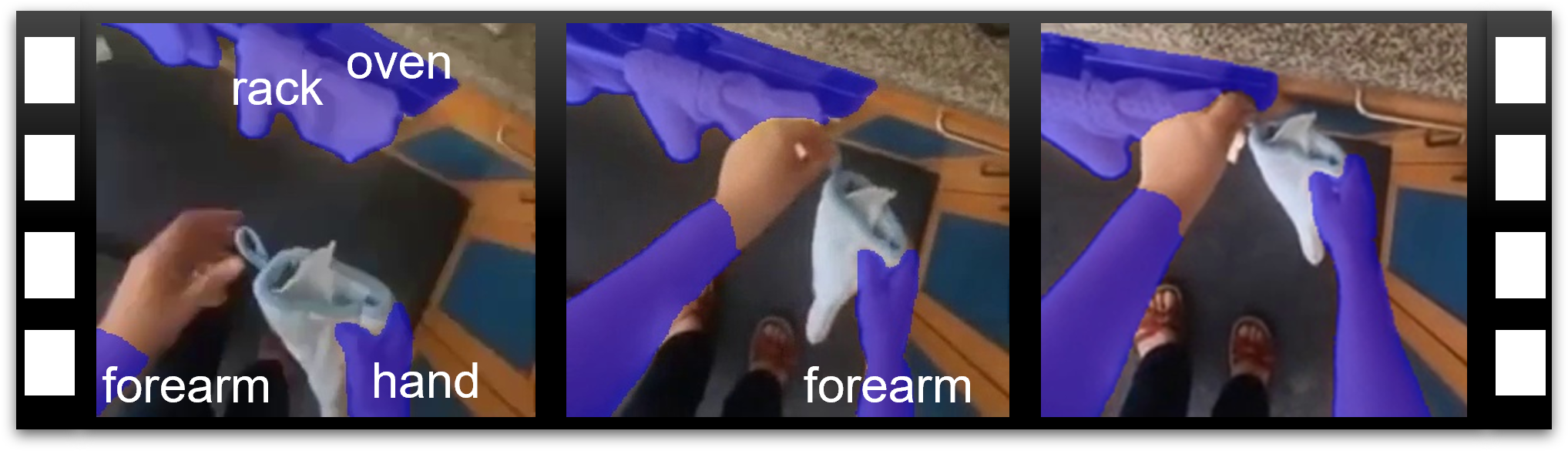}
    \caption{Segmented Contextual Objects}
    \label{fig:high_level_features-c}
  \end{subfigure}
  \caption{Example segmentations illustrating the Active Hand (\Cref{fig:high_level_features-a}), Active Object (\Cref{fig:high_level_features-b}), and Contextual Objects (\Cref{fig:high_level_features-c}) in a video depicting the action labelled as “\emph{hang gloves}”.}
  \label{fig:high_level_features}
\end{figure}

\subsubsection{High-level object features}
These features capture the proportion of semantically meaningful elements remaining in each quadrant relative to the original video, conceptually similar to occlusion-level metrics used in amodal segmentation studies \cite{8954364_amodal_KINS_dataset, zhu2017semantic_amodal_segmentation}. As illustrated in \Cref{fig:high_level_features}, segmented regions were grouped into four semantic categories: \textit{Active Hand} (one or both hands directly involved in performing the action; \Cref{fig:high_level_features-a}), \textit{Active Object} (up to two objects manipulated by the Active Hand and central to the action; \Cref{fig:high_level_features-b}), \textit{Contextual Objects} (the combined area of all remaining segmented objects with potential contextual relevance, including the passive hand and forearms; \Cref{fig:high_level_features-c}), and \textit{Background} (all image regions not covered by any segmented object). 

We used the Segment Anything Model 2 \cite{ravi2024sam2} (SAM v2.1\_hiera\_large) to produce initial segmentations. The model generated pixel-wise binary masks for each semantic category in each frame, assigning 1 to pixels belonging to that object and 0 to all others. Frame-wise masks were then temporally aggregated, and suboptimal masks were further refined manually using a custom Python-based annotation tool. 

\begin{figure}[htbp]
  \centering
  \begin{subfigure}{\linewidth}
    \centering
    \includegraphics[width=\linewidth]{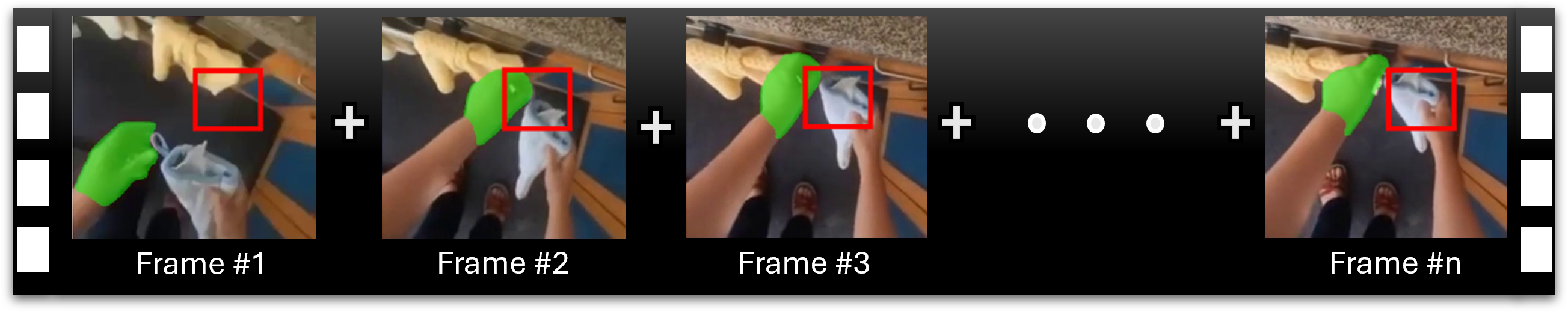}
    \caption{Object Surface Area Features}
    \label{fig:low_level_features-a}
  \end{subfigure}

  \vspace{0.8em}

    \centering
  \begin{subfigure}{\linewidth}
    \centering
    \includegraphics[width=\linewidth]{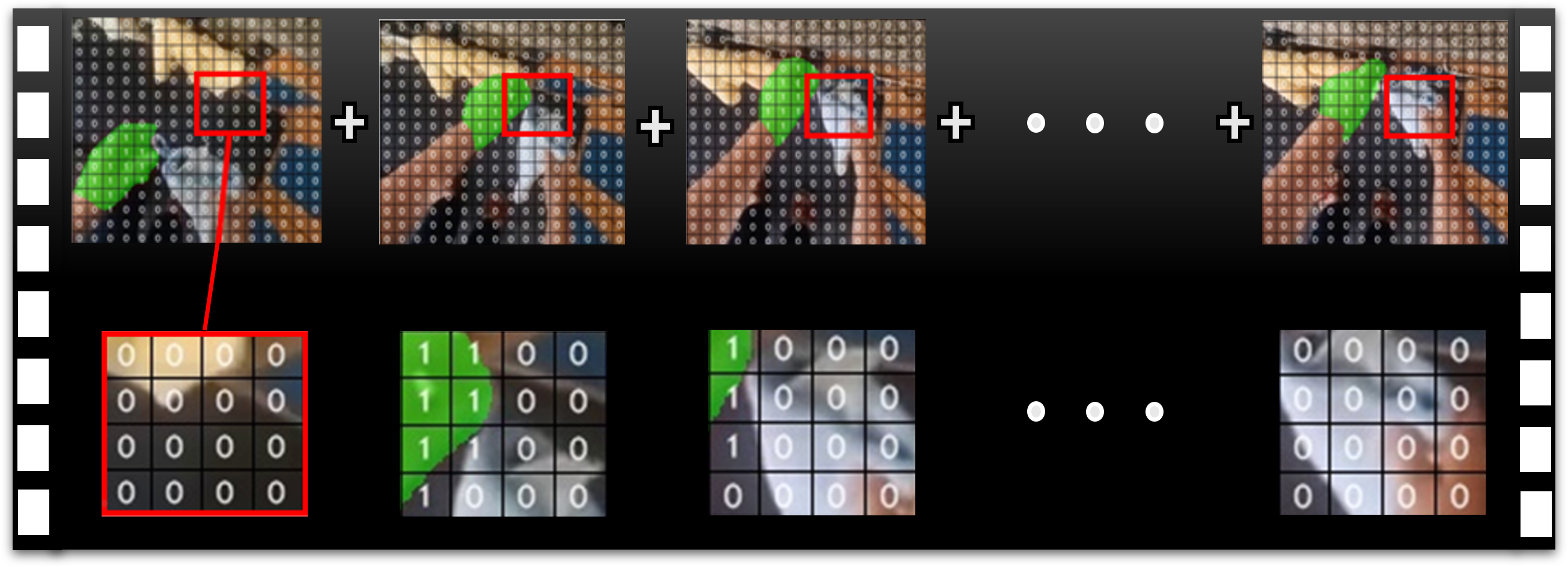}
    \caption{Object Surface Area Features with details}
    \label{fig:low_level_features-aa}
  \end{subfigure}

  \vspace{0.8em}

  \begin{subfigure}{\linewidth}
    \centering
    \includegraphics[width=\linewidth]{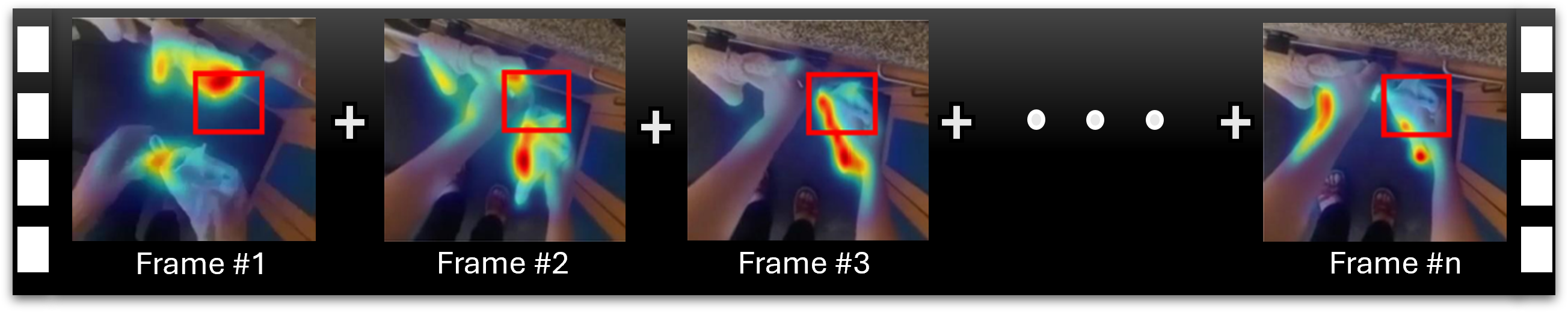}
    \caption{Mid-level Feature Maps}
    \label{fig:low_level_features-b}
  \end{subfigure}

    \vspace{0.8em}

  \begin{subfigure}{\linewidth}
    \centering
    \includegraphics[width=\linewidth]{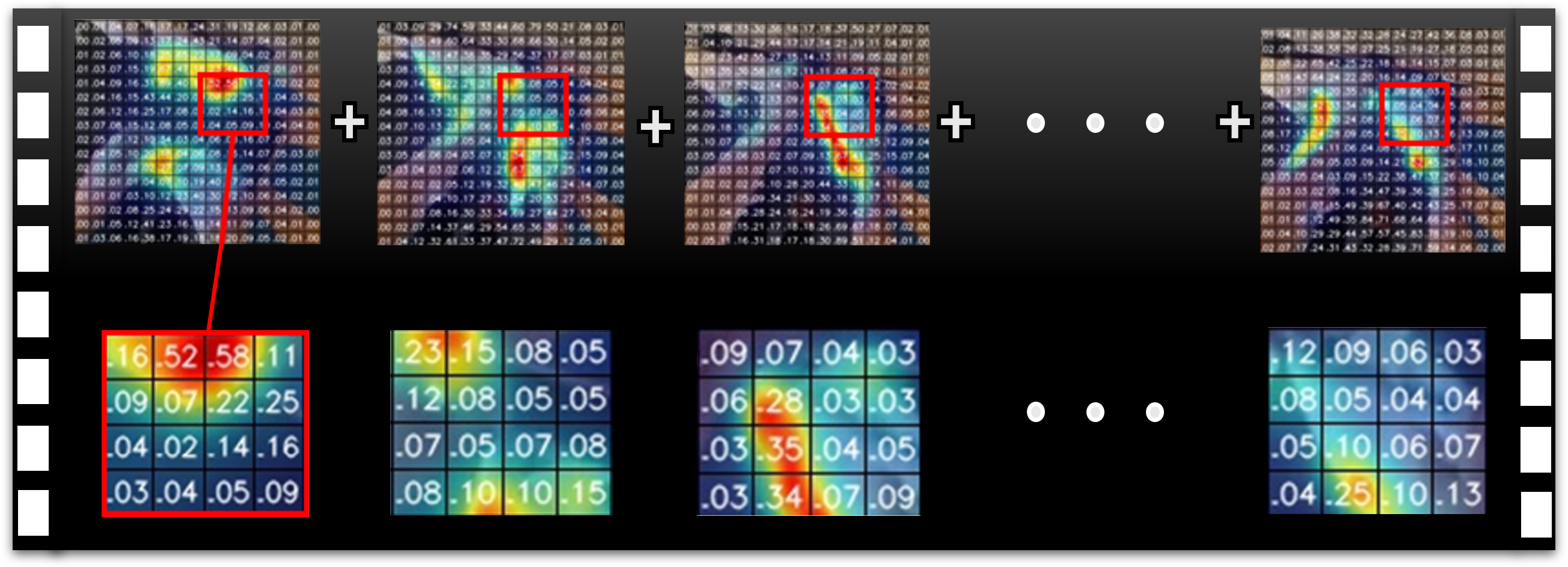}
    \caption{Mid-level Feature Maps with details}
    \label{fig:low_level_features-bb}
  \end{subfigure}
  
  \caption{Computation of Object Surface Area Features (\Cref{fig:low_level_features-a}) and Mid-level Feature Maps (\Cref{fig:low_level_features-b})}
  \label{fig:low_level_features}
\end{figure}

\begin{figure*}[htbp]
  \centering
  \begin{subfigure}{0.32\linewidth}
    \includegraphics[width=\linewidth]{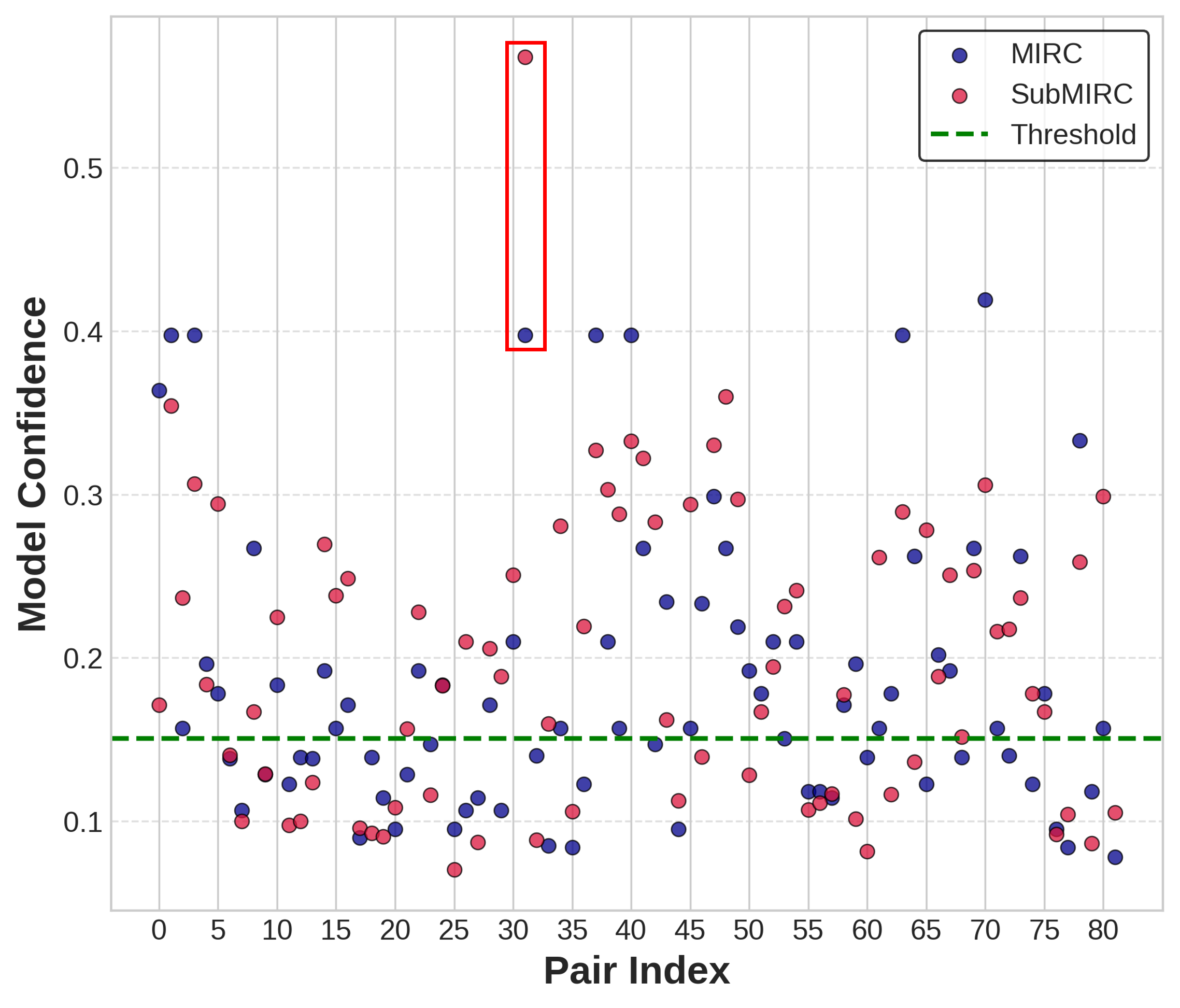}
    \caption{S Class \emph{Put}}
    \label{fig:recog_gap_per_class-a}
  \end{subfigure}
  \hfill
  \begin{subfigure}{0.32\linewidth}
    \includegraphics[width=\linewidth]{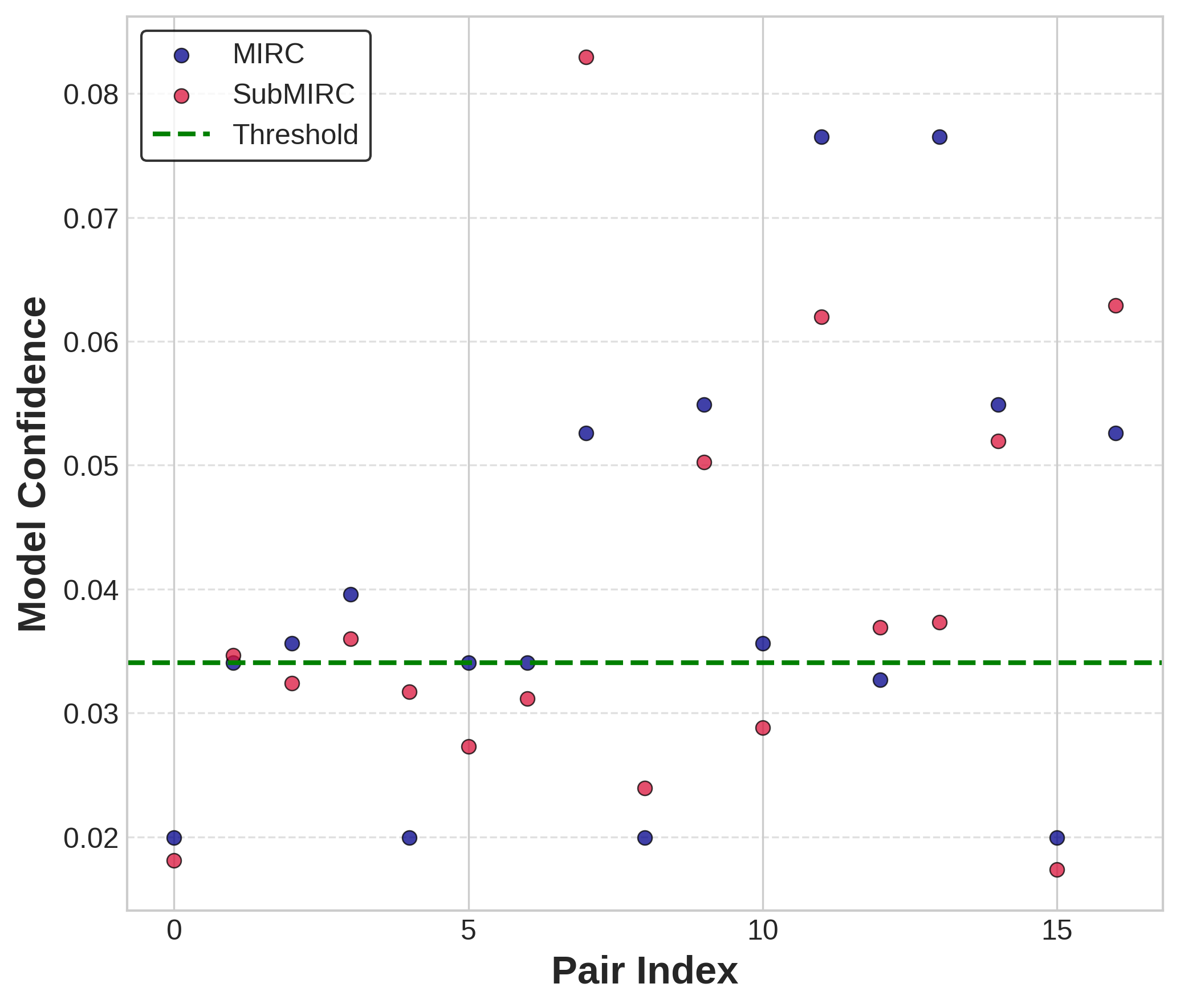}
    \caption{S Class \emph{Insert}}
    \label{fig:recog_gap_per_class-b}
  \end{subfigure}
  \begin{subfigure}{0.32\linewidth}
    \includegraphics[width=\linewidth]{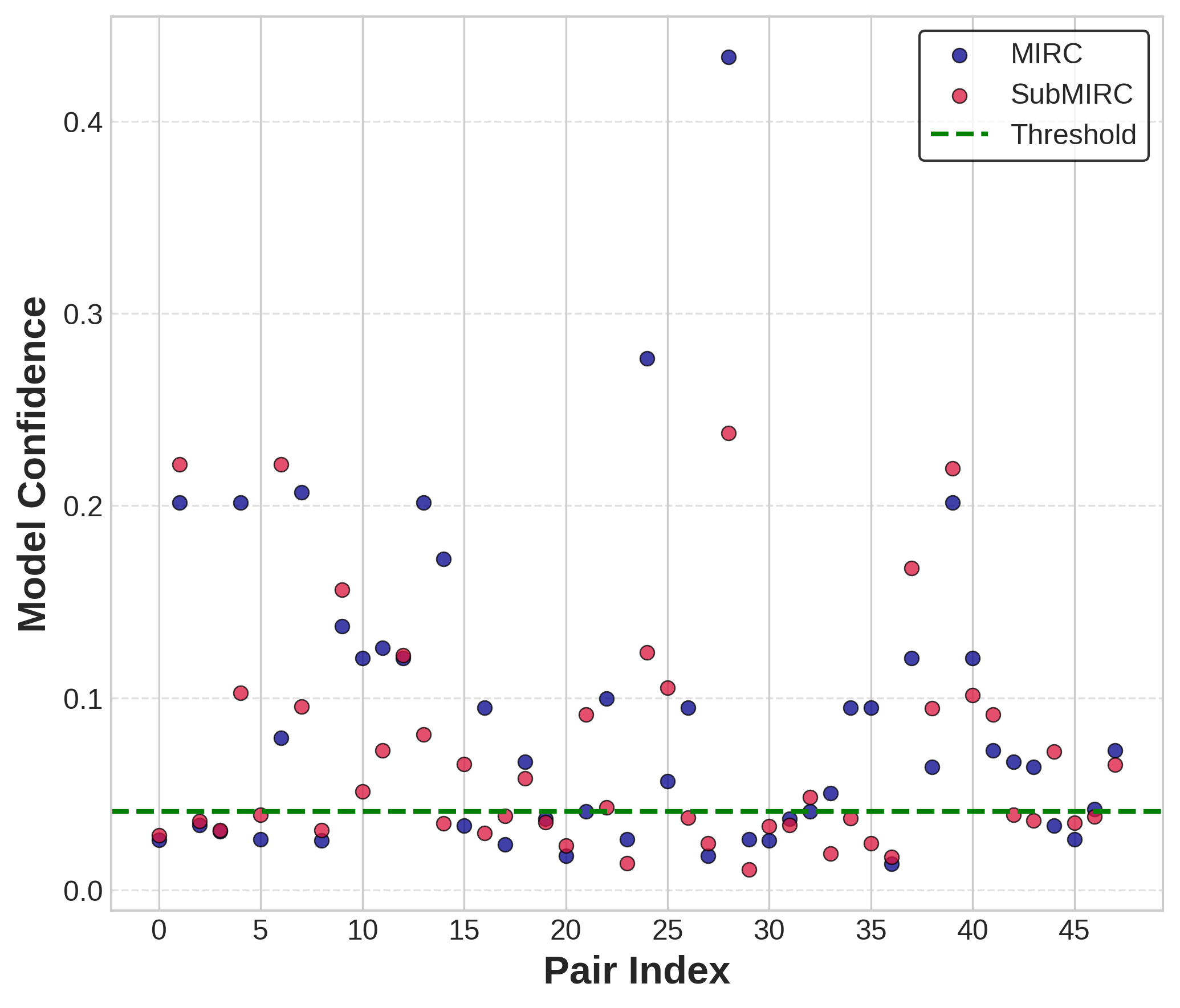}
    \caption{S Class \emph{Wash}}
    \label{fig:recog_gap_per_class-c}
  \end{subfigure}

  \vspace{1em}

  \centering
  \begin{subfigure}{0.32\linewidth}
    \includegraphics[width=\linewidth]{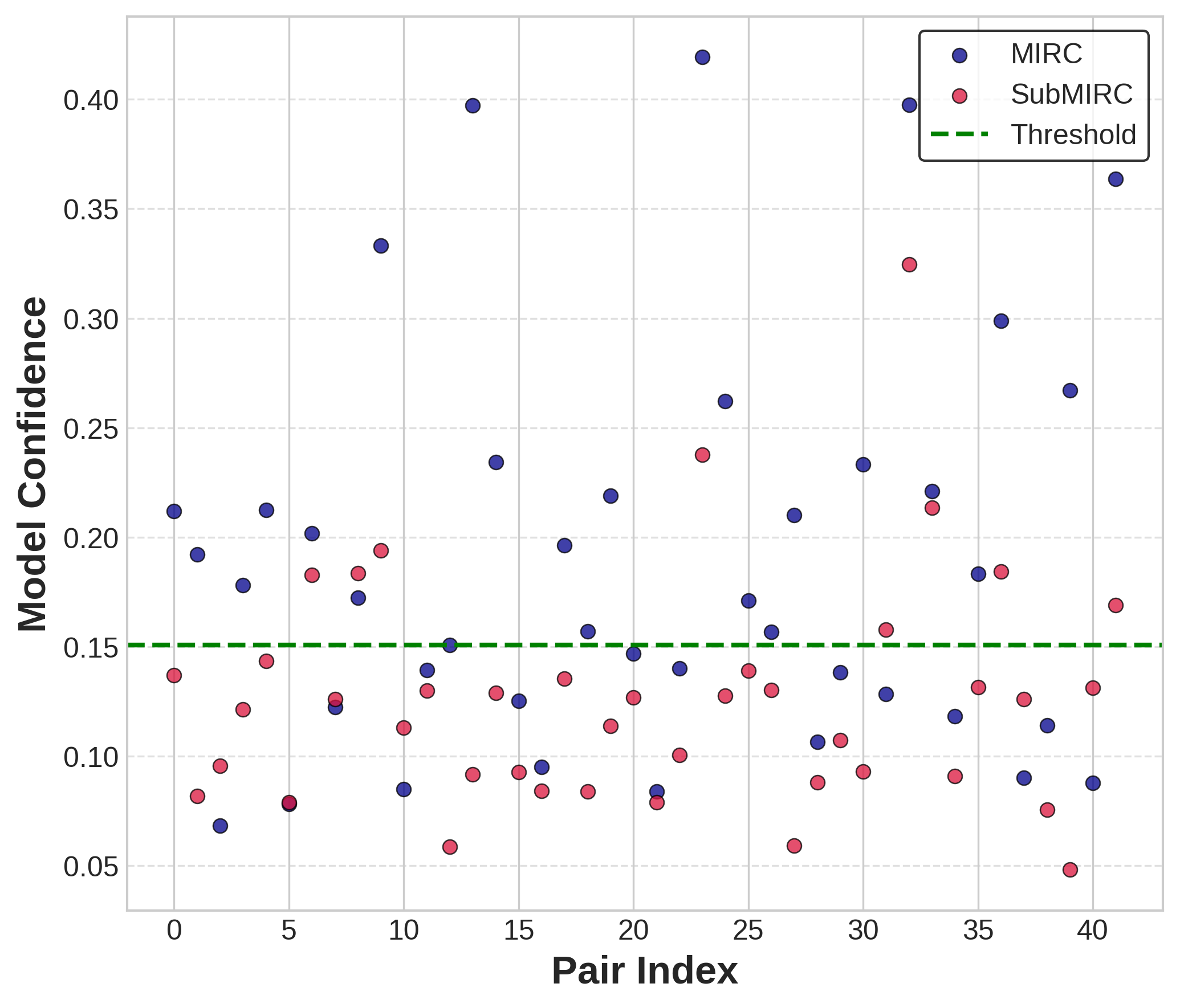}
    \caption{ST Class \emph{Put}}
    \label{fig:recog_gap_per_class-d}
  \end{subfigure}
  \hfill
  \begin{subfigure}{0.32\linewidth}
    \includegraphics[width=\linewidth]{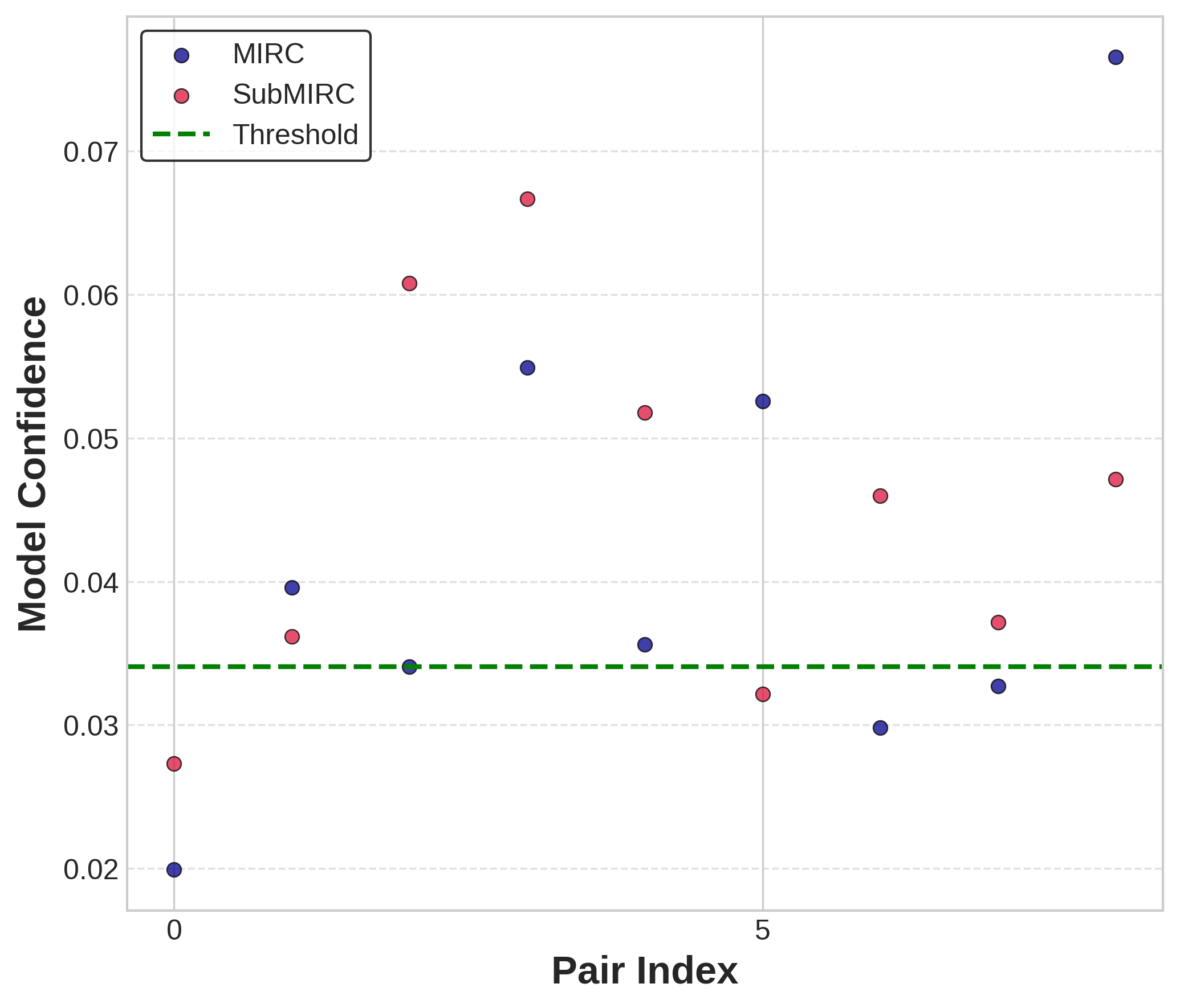}
    \caption{ST Class \emph{Insert}}
    \label{fig:recog_gap_per_class-e}
  \end{subfigure}
  \begin{subfigure}{0.32\linewidth}
    \includegraphics[width=\linewidth]{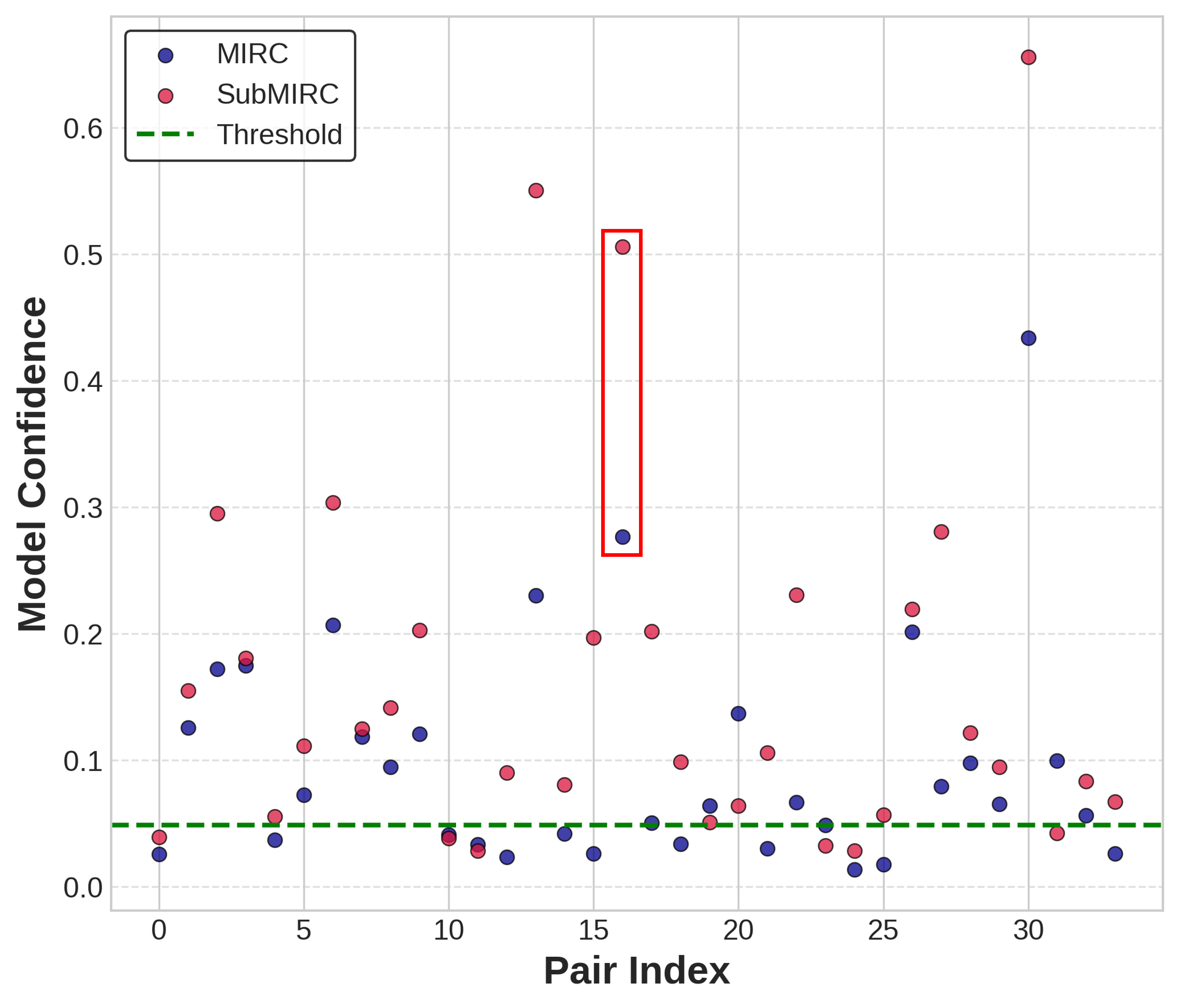}
    \caption{ST Class \emph{Wash}}
    \label{fig:recog_gap_per_class-f}
  \end{subfigure}
  \caption{Class-wise recognition gap for the AI model relative to a human-calibrated threshold. The recognition gap is computed for each MIRC–sub-MIRC pair using class-specific acceptance thresholds derived from human recognition performance. The top row shows results for spatial reductions, while the bottom row correspond to spatiotemporal scramblings, denoted with S and ST, respectively. In the spatial setting, (\Cref{fig:recog_gap_per_class-a,fig:recog_gap_per_class-b,fig:recog_gap_per_class-c}) present three representative classes and sample 31 of the \emph{put} class is highlighted by the red box and analysed in detail in Section~\ref{sec:experiments} (see \Cref{fig:reduced_video}). For the spatiotemporal setting, (\Cref{fig:recog_gap_per_class-d,fig:recog_gap_per_class-e,fig:recog_gap_per_class-f}) present the same classes as in the spatial case and sample 16 of the \emph{wash} class is highlighted by the red box for further analysis in \Cref{fig:spatio_temporal_reduced_video}.}
  \label{fig:recog_gap_per_class}
\end{figure*}

\begin{table*}[t]
  \centering
  \footnotesize
  \setlength{\tabcolsep}{1.3pt}
  \begin{tabular}{c|c|cccccccccccccc}
    \toprule
    \multicolumn{16}{c}{\textbf{Classes}} \\
    \midrule
    \textbf{Classifier} & \textbf{Metric} 
    & close & cut & hang & insert & open & peel & pour & put 
    & remove & serve & take & turn-off & turn-on & wash \\
    \midrule

    \multirow{2}{*}{Human} 
    & RG  
    & +42.92 & +36.59 & +40.52 & +40.00 & +38.07 & +38.20 & +32.21 & +35.49 
    & +31.36 & +42.88 & +39.03 & +38.75 & +31.11 & +31.56 \\
    & Std. 
    & 15.25 & 14.46 & 16.54 & 15.61 & 14.35 & 16.70 & 14.78 & 15.92 
    & 13.13 & 18.24 & 14.75 & 19.53 & 15.96 & 17.33 \\

    \midrule

    \multirow{2}{*}{AI} 
    & RG  
    & -2.30 & +1.21 & +0.34 & +0.36 & +5.40 & +0.04 & +0.06 & -1.05 
    & +5.71 & 0.00 & +15.11 & +5.36 & +19.43 & +3.12 \\
    & Std. 
    & 10.15 & 2.92 & 1.17 & 1.58 & 10.09 & 0.04 & 0.76 & 8.46 
    & 4.67 & 0.01 & 12.89 & 6.96 & 11.78 & 6.98 \\

    \bottomrule
  \end{tabular}
  \caption{\textbf{Spatial recognition gaps} (RG) and standard deviations (Std.) for human and AI classifiers, reported per action class. Values are percentages. Positive RG values indicate a reduction in recognition performance under scrambling, whereas negative values indicate an improvement.}
  \label{tab:AverageRecognitionGaps}
\end{table*}

\subsubsection{Mid-level visual features.}
 Different configurations of mid-level features compound into high-level features and  drive human attention in the initial stages of processing to potentially informative objects or regions. The features used here were obtained using the Graph-Based Visual Saliency (GBVS) algorithm, \cite{NIPS2006_4db0f8b0_Graph_Based_Visual_Saliency}, which computes interpretable feature channels (DKL Colour, Intensity, Orientation, Colour, Flicker, Contrast, and Motion) aligned with the psychophysics literature and able to accurately predict human fixations when combined into a saliency map. Here we focused on the 7 mid-level conspicuity feature maps. These maps provide normalised feature activation scores in the range $[0,1]$ for each pixel, reflecting its activation relative to the surrounding region.

\begin{table*}[htbp]
    \centering
    \begin{tabular}{c|l}
    \toprule
        \textbf{Feature} & \textbf{Definition} \\
        \toprule
        DKLcolour & Chromatic contrast in the Derrington‑Krauskopf‑Lennie colour space \\
        Intensity & Luminance (brightness) variations \\
        Orientation & Local edge orientation energy \\
        Colour & Double‑opponent Red–Green / Blue–Yellow contrasts \\
        Flicker & Frame-to-frame luminance change (temporal contrast) \\
        Contrast & Local brightness variability \\
        Motion & Apparent motion energy between consecutive frames derived from optic flow \\
        \bottomrule
    \end{tabular}
    \caption{GBVS feature channels and their definitions.}
    \label{tab:features_definition}
    
\end{table*}

We next analysed the visual drivers of recognition failure and recovery shown in \Cref{fig:feature_drop} by tracking the visual components encoded in mid- and high-level feature maps across spatial reductions. We did this by quantifying the proportion of each object’s surface area and each GBVS feature activation score preserved in the quadrant after cropping, relative to the full video. As illustrated in \Cref{fig:low_level_features}, we first summed pixel-wise values for every single frame inside the cropped quadrant, $M_{q,p}$, and the full video, $M_{f,p}$, then summed those across all frames of the quadrant, $S_{q}$, and the full video, $S_{f}$:

$$
S_q = \sum_{q=1}^{Q} \sum_{p=1}^{P} M_{q,p}
$$

$$
S_f = \sum_{f=1}^{F} \sum_{p=1}^{P} M_{f,p}
$$

Here, $M_{q,p}$ and $M_{f,p}$ denote either binary surface-area values (for object-based features) or continuous activation values in the range $[0,1]$ (for mid-level features) at pixel $p$, within the cropped quadrant or full frame, respectively. For temporally scrambled quadrants, GBVS conspicuity maps used for $M_{q,p}$ were recomputed after frame shuffling, while $M_{f,p}$ remained based on the full unscrambled video. 

The final feature value for each quadrant was obtained as the ratio
$$
p = \frac{S_q}{S_f},
$$
representing the proportion of total object surface area or feature activation retained after spatial reduction or temporal scrambling.
\subsection{Spatial Results}
\label{subsec:spatial_results}

To analyse the effect of multi-level spatial reduction on both human and AI model performance, we evaluate results from human observers and the S4V model \cite{side4video} using the two aforementioned metrics. This analysis reveals nuanced differences in behaviour and robustness between the two predictor types.

\subsubsection{Quantitative Spatial Results}
\label{subsubsec:spatial_quant}

\begin{figure}[t]
  \centering
  \begin{subfigure}{\linewidth}
    \centering
    \includegraphics[height=1.05cm]{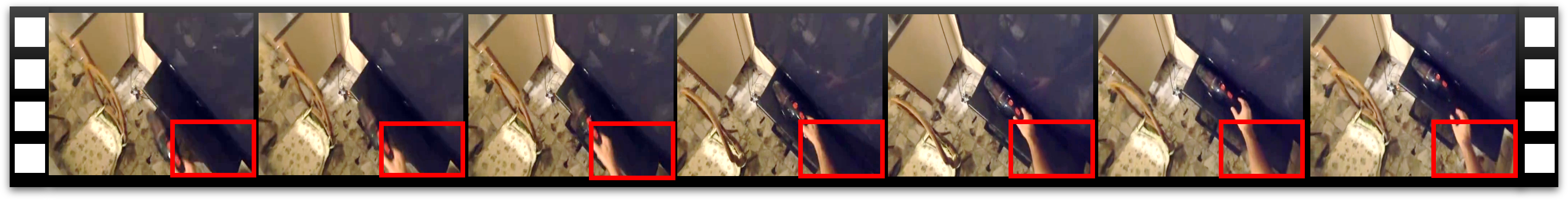}
    \caption{Original video - No reduction - GT label \emph{put}}
    \label{fig:reduced_video-a}
  \end{subfigure}

  \vspace{1em} 

  \begin{subfigure}{\linewidth}
    \centering
    \includegraphics[height=1.05cm]{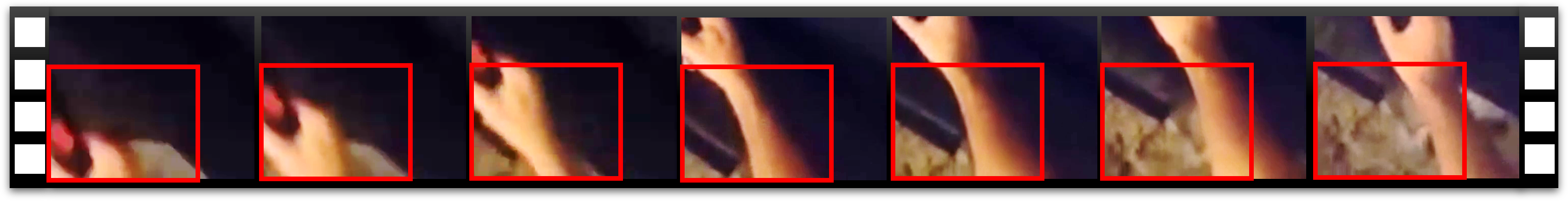}
    \caption{Level 2: The MIRC level}
    \label{fig:reduced_video-b}
  \end{subfigure}
  
  \vspace{1em} 

  \begin{subfigure}{\linewidth}
    \centering
    \includegraphics[height=1.05cm]{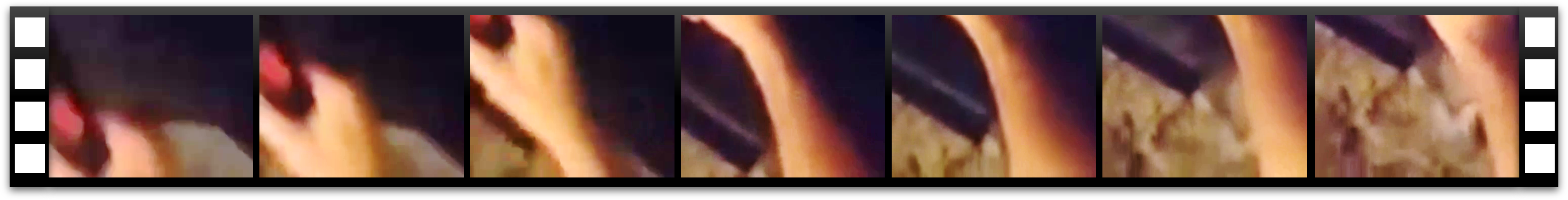}
    \caption{Level 3: The sub-MIRC version of \Cref{fig:reduced_video-b}}
    \label{fig:reduced_video-c}
  \end{subfigure}

  \caption{Sample illustration of spatial feature loss across the original frame, MIRC, and sub-MIRC levels for a video from the \emph{put} action class (corresponding to highlighted sample in \Cref{fig:recog_gap_per_class-a}). The red bounding boxes highlight the spatially reduced child video at the next reduction level. As spatial resolution decreases, background distractors are progressively cropped out, allowing the AI model to recognise the action correctly and thereby reducing the recognition gap between levels.}
  \label{fig:reduced_video}
\end{figure}

\begin{figure}[htbp]
  \centering
    \includegraphics[width=\linewidth]{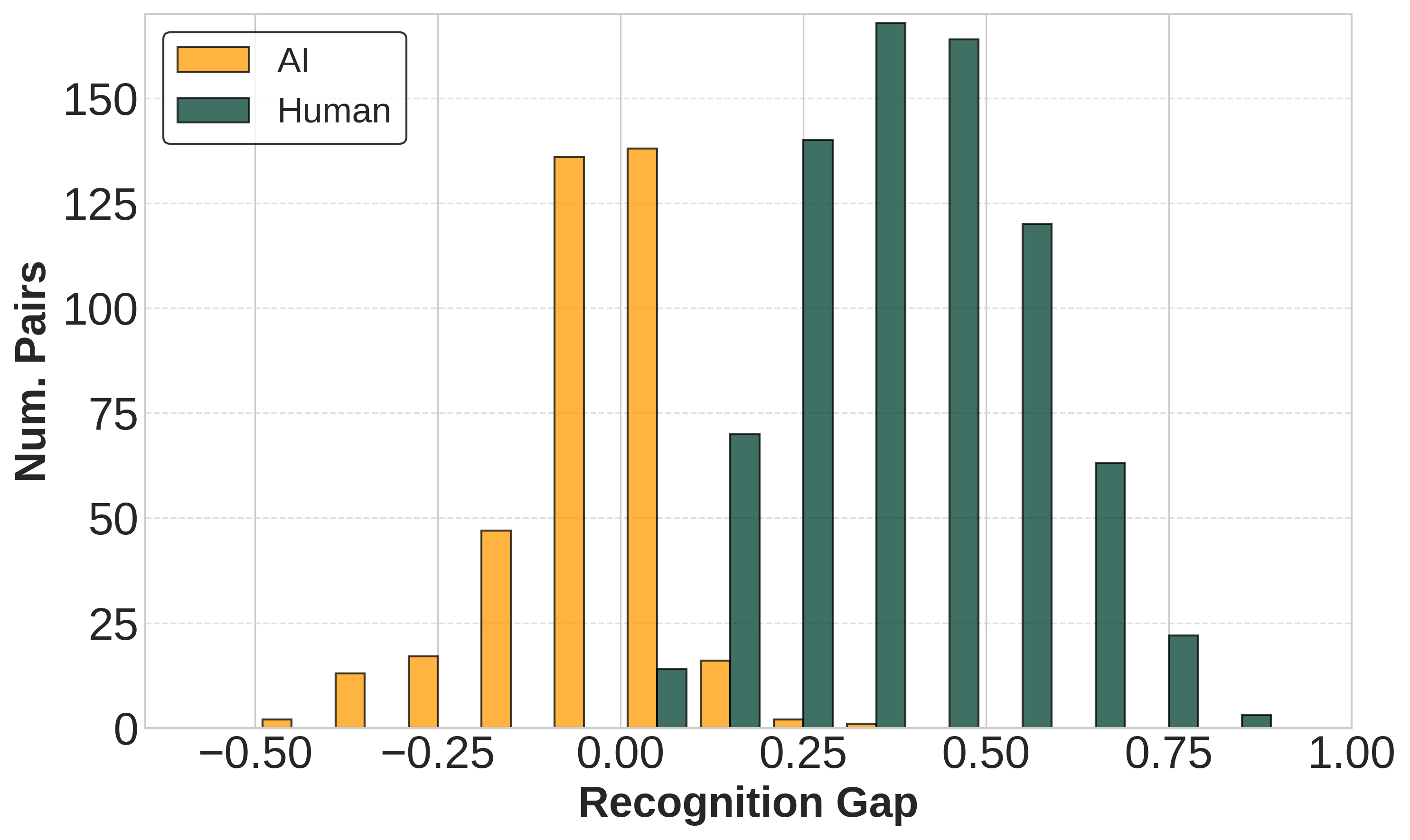}
  \caption{Frequency distribution of recognition gaps between AI model and humans.}
  \label{fig:recog_gap}
\end{figure}

\noindent\textbf{a) Recognition Gap.} This metric helps us understand the boundary of human recognition (MIRC and sub-MIRC) and compare how the AI model behaves at that same boundary. We evaluate the recognition gap across all action classes in Epic-ReduAct. Representative examples, including varying numbers of pairs, are shown in \Cref{fig:recog_gap_per_class-a}-\ref{fig:recog_gap_per_class-c} to facilitate a direct comparison between human performance and the AI model. These plots depict the relative positions of each MIRC/sub-MIRC pair with respect to the model’s confidence scores and the class-specific threshold line. A notable observation is that, in many cases, as shown in the highlighted sample in \Cref{fig:recog_gap_per_class-a}, sub-MIRC samples attain higher confidence scores than their corresponding MIRCs, which explains the negative recognition gap values shown later in \Cref{fig:recog_gap}. This indicates that recognition performance can, in some cases, improve as spatial resolution is reduced. To further illustrate this effect qualitatively, we present an example video which shows how, in certain scenarios, spatial reduction can improve model performance rather than degrade it.
\Cref{fig:reduced_video} shows an example video from the \emph{put} class at three levels of reduction: the original video, the MIRC level (Level~2), and the sub-MIRC level (Level~3). At Level 2, the video still contains numerous irrelevant visual details. By contrast, at Level~3, removing such features enables the model to focus more effectively on the ongoing action. This focus improves action recognition, increasing the model’s confidence from 39\% at the MIRC level to 56\% at the sub-MIRC level.
This example, similar to several other videos, corresponds to the red (sub-MIRC) and blue (MIRC) points highlighted in \Cref{fig:recog_gap_per_class-a} (pair 31). 


Recognition gaps across all classes for both humans and the AI model are summarised in \Cref{tab:AverageRecognitionGaps}. Positive values indicate a reduction in recognition performance, whereas negative values indicate an improvement. The small negative values in the results indicate that MIRC and sub-MIRC do not define clear recognition boundaries for the AI model, unlike those defined by human observers. Human recognition accuracy drops sharply across all classes. At the same time, the AI model generally shows improved or stable performance, as reflected by predominantly negative recognition gap values. For example, in the \emph{close} class, human performance decreases by 42.92\%, whereas the AI model improves by 2.30\%. The most minor reduction in human recognition is observed for the \emph{turn-on} class, which still shows a substantial drop of 31.11\%, underscoring the strong effect of MIRC and sub-MIRC constraints on human perception compared to AI with 19.43\% drop in the same class which is the largest among other classes.


\Cref{fig:recog_gap} shows the frequency distribution of recognition gaps for human observers and the AI model.
Our results exhibit distribution patterns that are consistent with prior image-based spatial reconstruction studies \cite{A_model_for_full_local_image_interpretation}. While AI models show modest improvements in recognition under spatial reduction, human accuracy consistently declines, with a markedly sharper drop than that of the AI models. Humans also experience substantial losses in recognition confidence, whereas spatial reduction can improve the AI model’s action recognition, as reflected by negative recognition gaps (\Cref{tab:AverageRecognitionGaps} and \Cref{fig:reduced_video}). Moreover, the frequency distributions are broader for humans, indicating greater variability in performance degradation, while changes in the AI model are more gradual and concentrated. This contrast is further reflected in the standard deviation values reported in \Cref{tab:AverageRecognitionGaps}. For humans, standard deviations range from 13.13\% to 19.53\%, indicating consistently high variability across action classes. In contrast, the AI model’s standard deviations range from 0.01\% to 12.89\%, with most values clustered below 8\%, suggesting more uniform behaviour under spatial reduction. Overall, these findings suggest that, despite recent advances, a significant gap between human and machine recognition capabilities remains.

\begin{figure}[htbp]
  \centering
  \begin{subfigure}{\linewidth}
    \includegraphics[width=\linewidth]{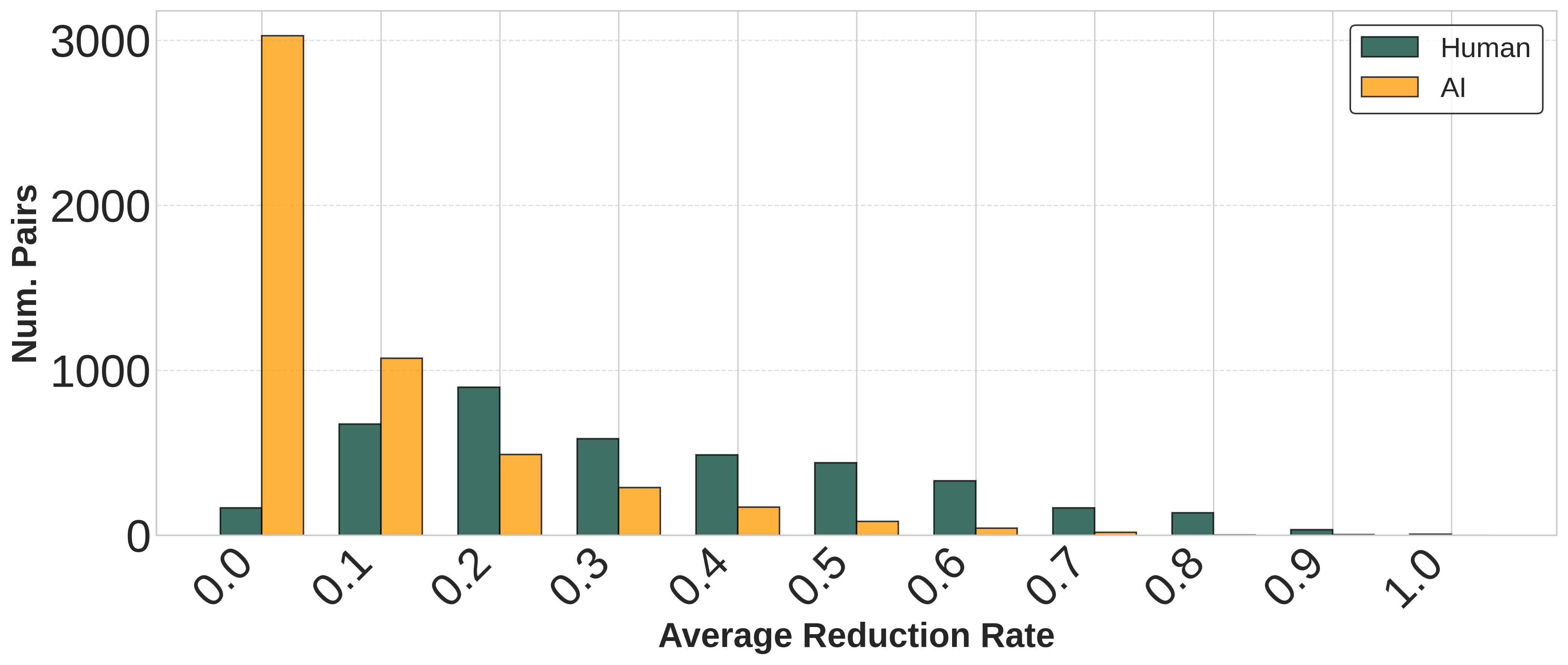}
    \caption{Frequency distribution of average reduction rates across pairs of video quadrants.}
    \label{fig:reduc_rate-a}
  \end{subfigure}

  \vspace{1em}
  \begin{subfigure}{\linewidth}
    \includegraphics[width=\linewidth]{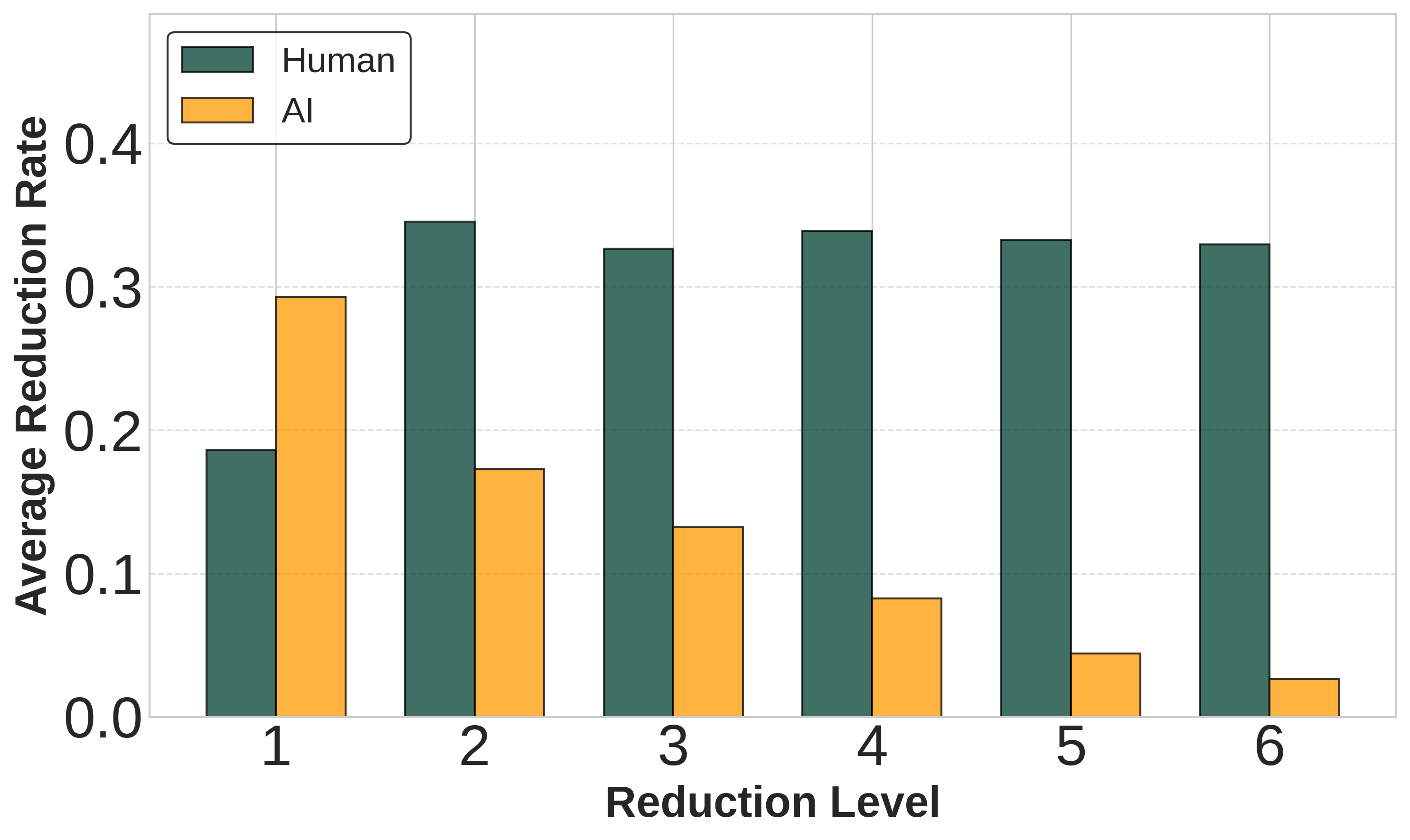}
    \caption{Average reduction rate (positive gaps) as a function of reduction level, restricted to MIRC–sub-MIRC pairs}
    \label{fig:reduc_rate-b}
  \end{subfigure}
  
  \vspace{1em}

  \begin{subfigure}{\linewidth}
    \includegraphics[width=\linewidth]{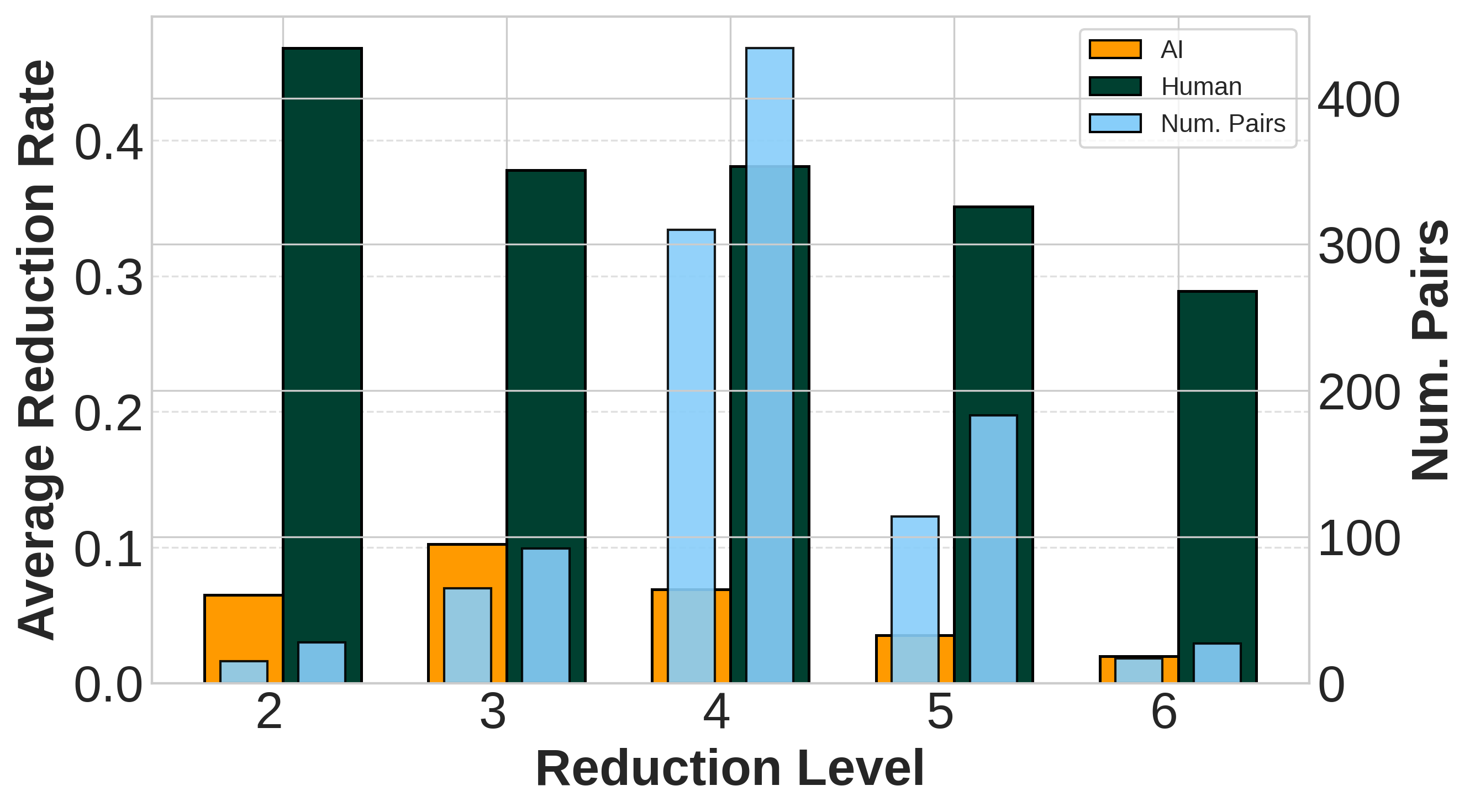}
    \caption{Average reduction rate as a function of reduction level}
    \label{fig:reduc_rate-c}
  \end{subfigure}

  \caption{Average reduction rate comparison between the AI model and humans. \Cref{fig:reduc_rate-a} shows the distribution of parent–child pairs across all reduction levels, binned at intervals of 0.1 in reduction rate. \Cref{fig:reduc_rate-b} shows the average reduction rate as a function of reduction level for all parent–child quadrant pairs. \Cref{fig:reduc_rate-c} shows the same comparison, restricted to MIRC/sub-MIRC pairs, with the number of pairs visualised for each reduction level.}
  \label{fig:reduc_rate}
\end{figure}

\noindent\textbf{b) Average Reduction Rate.} This metric enables comparison of model performance with results from psychophysical studies, highlighting levels at which recognition deteriorates significantly. \Cref{fig:reduc_rate} presents human and AI classification performance.  \Cref{fig:reduc_rate-a} shows the frequency distribution of the average reduction rate for parent and child quadrants. Human and AI reduction rates exhibit different overall patterns as humans show a more pronounced decline in recognition, as reflected by a wider spread of higher reduction rates and more catastrophic failures across levels. To explain the larger number of pairs for the AI model compared to humans, this metric only counts pairs that show a reduction in recognition from parent video to child quadrant. This indicates that the AI experiences reductions in more pairs, though most are small, as reflected by the high frequencies of pairs in the 0.0–0.2 range. In contrast, humans exhibit larger reductions overall, with a greater number of samples showing substantial performance drops, as seen in both \Cref{fig:reduc_rate-a} and \Cref{fig:reduc_rate-c}.

\Cref{fig:reduc_rate-b} shows the average reduction rate as a function of reduction level, highlighting the stages at which recognition declines most markedly. The figure indicates that the AI model experiences a gradual decrease in accuracy across levels, reflecting steady sensitivity to diminishing spatial information. In contrast, human recognition is robust to the first reduction level compared to the AI but deteriorates sharply from level 2 onwards as indicated by values exceeding 0.3. 

\Cref{fig:reduc_rate-c} reports the average reduction rate as a function of reduction level, restricted to MIRC–sub-MIRC pairs. In addition to performance trends, \Cref{fig:reduc_rate-c} indicates the number of MIRC/sub-MIRC pairs available at each reduction level. This analysis highlights Levels~2 to~5 as the most critical stages for information loss for both humans and the AI model. However, while human recognition exhibits substantial and abrupt degradation at these levels, the AI model shows a more gradual decline, indicated by the generally lower average reduction rates less than 0.1. Overall, the trends observed in \Cref{fig:reduc_rate-c} closely mirror those seen for all parent–child pairs in \Cref{fig:reduc_rate-b}.

\subsubsection{Qualitative Spatial Results}
\label{subsubsec:spatial_qual}

To complement the recognition-gap and reduction-rate results, we conducted an analysis of the features present in the quadrants to understand how and why action predictions change as the videos are progressively cropped. For each video, we construct a tree structure over its spatially cropped variants, as in \Cref{fig:reducing_video}, but restricted to spatial reductions. Each node in the tree corresponds to a spatial quadrant of its parent video. We then traverse each tree and record classification transitions by comparing the model’s predicted class (only the verb) at each parent–child pair, specifically identifying cases in which the verb prediction reverses between correct and incorrect responses. To ensure that these transitions reflect changes in action recognition rather than object detection errors, we analyse only pairs with a verb-annotation transition.
\begin{figure*}[htbp]
  \centering
  \begin{subfigure}{0.49\linewidth}
    \includegraphics[width=\linewidth]{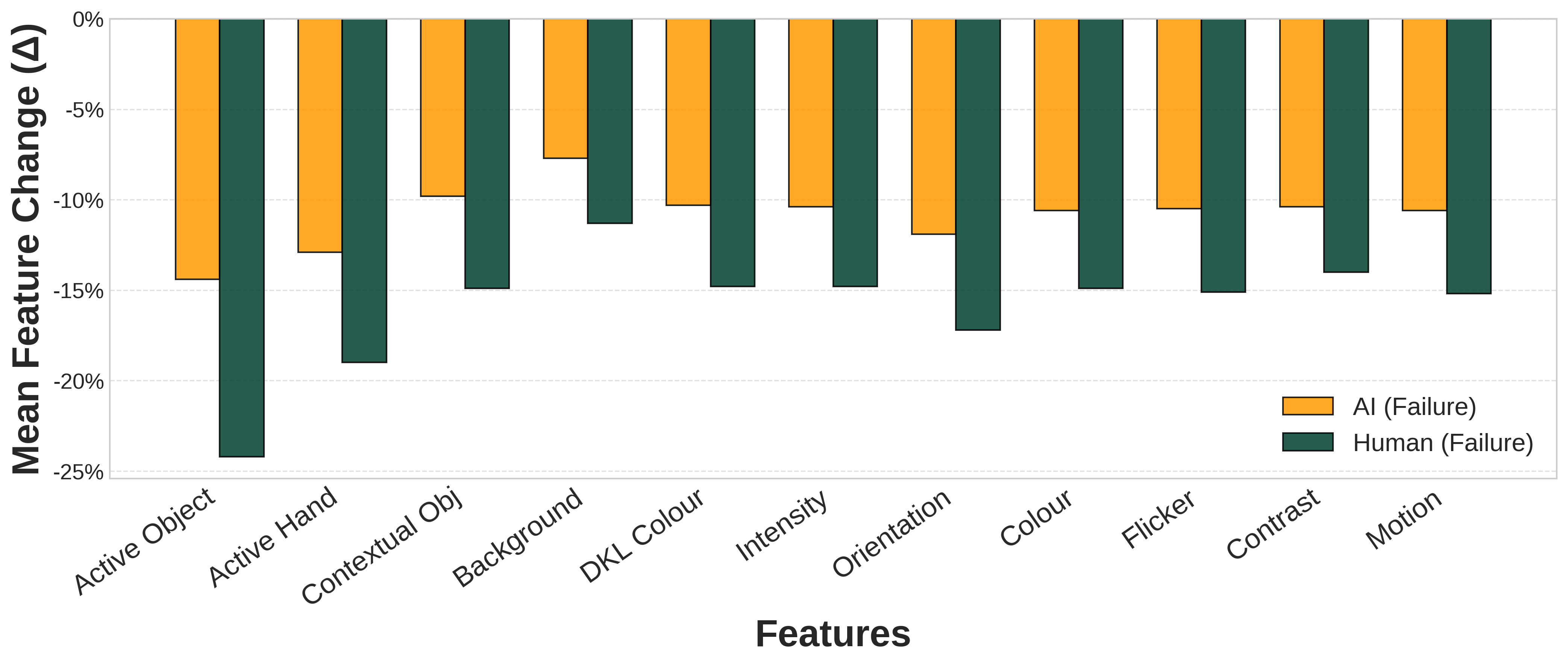}
    \caption{Failure (Feature Loss)}
    \label{fig:feature_drop-a}
  \end{subfigure}
  \hfill
  \begin{subfigure}{0.49\linewidth}
    \includegraphics[width=\linewidth]{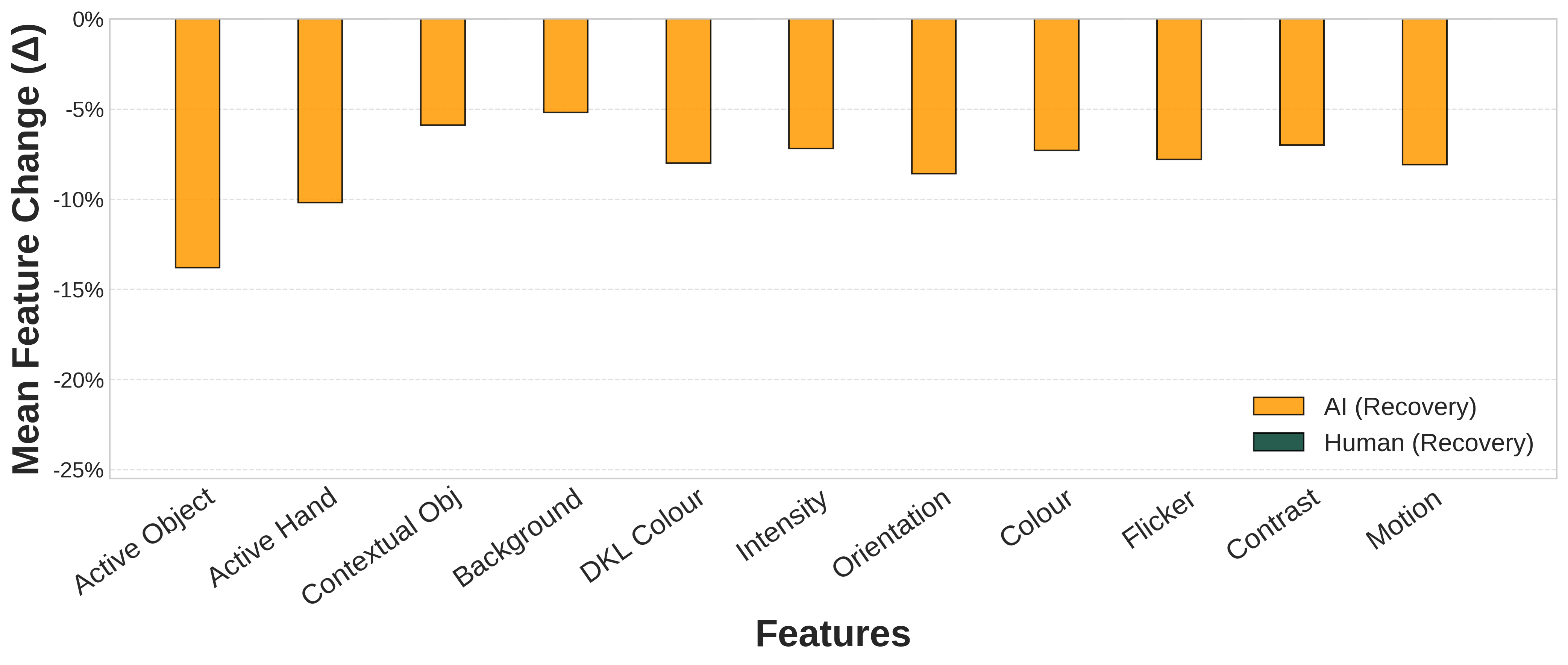}
    \caption{Recovery (Noise Reduction)}
    \label{fig:feature_drop-b}
  \end{subfigure}
  \caption{Feature drop analysis. Quantifying the mid- to high-level features that lead to recognition failure and recovery. 
  \Cref{fig:feature_drop-a} and \Cref{fig:feature_drop-b} show the mean feature change ($\Delta$) in terms of feature visibility for Failure (Correct $\rightarrow$ Incorrect) and Recovery (Incorrect $\rightarrow$ Correct) transitions in human observers and the AI model. Note that human observers did not show any recovery transitions.}

  \label{fig:feature_drop}
\end{figure*}

\begin{figure}[t]
  \centering
    \centering
    \includegraphics[width=\linewidth]{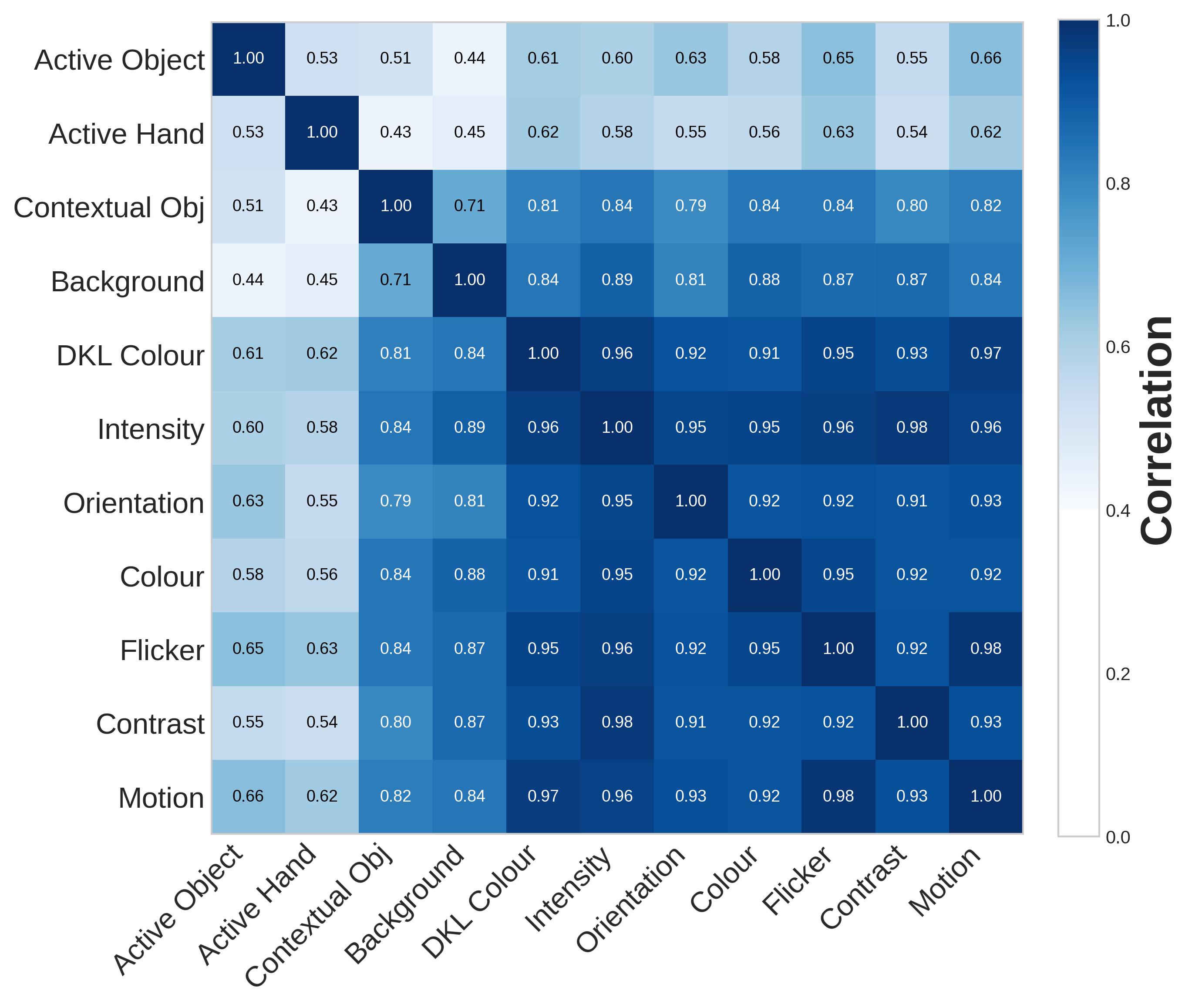}
    \caption{Correlation matrix for AI failure cases, illustrating the strength of feature correlations across transitions in failure cases. (Correct $\rightarrow$ Incorrect).}
    \label{fig:corr_matrix_failure}
\end{figure}


\Cref{fig:feature_drop} tracks how the presence of key spatial features changes across successive spatial reductions at points where action recognition changes. This feature drop analysis reveals two distinct mechanisms that govern recognition behaviour under spatial degradation: \textit{feature loss}, where the removal of informative visual cues leads to recognition failure (Correct prediction $\rightarrow$ Incorrect prediction, shown in \Cref{fig:feature_drop-a}), and \textit{noise reduction}, where suppressing distracting or irrelevant cues enables recognition recovery (Incorrect prediction $\rightarrow$ Correct prediction, shown in \Cref{fig:feature_drop-b}). Before analysing feature changes across transitions, we note a quantitative pattern not directly visible in the previous figures: Humans exhibit fewer prediction flips than the AI model, largely because the human experiment did not continue after failure cases.
This was done to limit the number of tested quadrants and optimise the search for MIRCs.
Specifically, the AI produced 866 transitions from correct to incorrect predictions and 257 from incorrect to correct, whereas humans exhibited 573 transitions from correct to incorrect.
Although previous work has shown that human visual recognition can improve with reduced clutter \cite{LEVI2008635}, in a pilot experiment with the humans participants, they rarely recovered once a quadrant was not recognised, therefore we only consider AI in the recovery analysis in \Cref{fig:feature_drop-b}.

\paragraph{Failure via Feature Loss (Correct $\rightarrow$ Incorrect)}

For the AI model, our analysis first examined the magnitude of feature reduction when the AI transitions from a correct to an incorrect prediction. A first glance at \Cref{fig:feature_drop-a}, the $\Delta$ values suggest that the loss of the \emph{Active Object} (mean $\Delta = -0.144$) and \emph{Active Hand} (mean $\Delta = -0.129$) are the primary drivers of failure, as these features experience the largest absolute reductions. However, this surface-level interpretation is misleading. A deeper statistical comparison reveals that the AI loses nearly identical amounts of the \emph{Active Object} even when it succeeds ($\Delta = -0.137$), with a negligible gap of only $0.007$ between success and failure. Thus, the loss of high-level features (e.g., the main actor and action) are not the distinguishing features behind the AI failure.

Instead, a correlation analysis (\Cref{fig:corr_matrix_failure}) reveals the true failure mechanism: \emph{systemic collapse}. In failure cases, analysis of feature changes across transitions reveals extremely high correlations between the \emph{Contextual Objects} feature and mid-level visual statistics, such as \emph{Flicker} ($r = 0.84$), \emph{Colour} ($r = 0.84$), and \emph{Motion} ($r = 0.82$). Crucially, there is a moderate-to-strong correlation between the \emph{Active Object} and \emph{Contextual Objects} ($r = 0.51$), which is substantially higher than in success cases ($r = 0.38$). This indicates that AI failure is rarely caused by the loss of a single semantic cue. Rather, failure emerges when spatial reduction induces a synchronised collapse of the entire scene structure, wherein environmental context, lighting, motion, and the primary object degrade simultaneously.

This behaviour stands in contrast to human observers. For humans, failure is strictly \emph{object-focused}. The loss of the \emph{Active Object} in human failure cases is substantial ($\Delta = -0.24$), nearly double the corresponding loss observed for the AI. 
This indicates that humans fail primarily when the semantic core of the action, the actor itself, is removed, and they remain comparatively resilient to the loss of background environment. This is consistent with human observers' tendency to use high-level features \cite{PMID:33745819, LOUCKS200984}. The AI, by contrast, exhibits brittle behaviour: it fails when the global statistical structure of the image collapses.

\begin{figure}[t]
  \centering
  \begin{subfigure}{\linewidth}
    \centering
    \includegraphics[width=\linewidth]{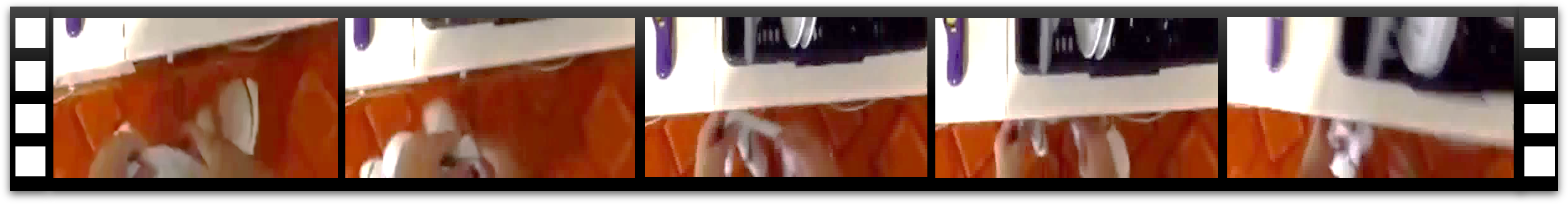}
    \caption{Original video (no reduction), verb correctly predicted as \emph{Put}.}
    \label{fig:easy_signal_loss-a}
  \end{subfigure}

  \vspace{0.8em}

  \begin{subfigure}{\linewidth}
    \centering
    \includegraphics[width=\linewidth]{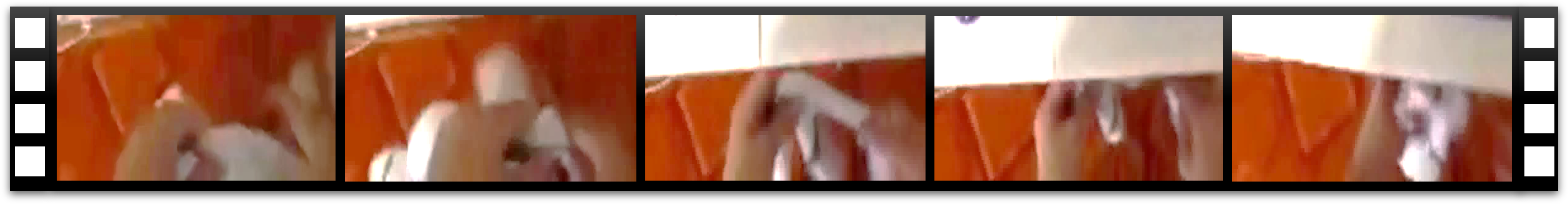}
    \caption{Level~1 (Lower-left crop), verb incorrectly predicted as \emph{Close}. \emph{Contextual Objects} visibility drop by 96\%.}
    \label{fig:easy_signal_loss-b}
  \end{subfigure}

  \vspace{0.8em}

  \begin{subfigure}{\linewidth}
    \centering
    \includegraphics[width=\linewidth]{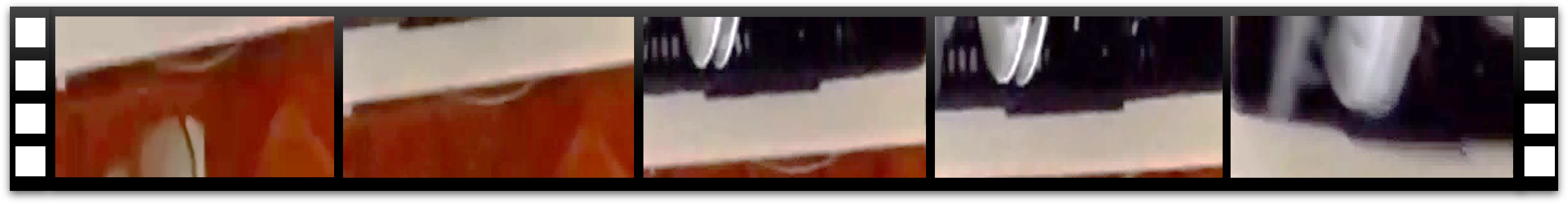}
    \caption{Level~1 (upper-right crop), verb correctly predicted as \emph{Put}. 79\% of \emph{Contextual Objects} visibility and more than 30\% activation of mid-level features.}
    \label{fig:easy_signal_loss-c}
  \end{subfigure}

  \caption{Example of a failure case under spatial reduction in a video in which the model produces an incorrect prediction after the first level of spatial reduction due to the complete occlusion of the \emph{Contextual Objects}, while correctly recognising the action when the \emph{Contextual Objects} and mid-level features remains visible.}
  \label{fig:easy_signal_loss}
\end{figure}

\Cref{fig:easy_signal_loss} illustrates this distinction between human and AI behaviour for the representative \emph{put} class.

As shown in \Cref{fig:easy_signal_loss-b} with a lower-left crop of the original video, the AI flips to an incorrect prediction despite the \emph{Active Object} and \emph{Active Hand} being largely preserved by $\Delta = -0.04$ and $\Delta = -0.12$, respectively. Instead, the failure is driven entirely by the destruction of the scene: the \emph{Contextual Objects} drop by $96\%$, while \emph{Motion} feature activation is reduced by $57\%$. Although the primary actor remains visible, allowing human observers to correctly identify the action, the AI fails because the surrounding environmental anchors that support its prediction are removed.

We next consider a scenario in which another spatial quadrant is correctly classified to further isolate the contribution of the \emph{Active Object} and \emph{Active Hand}. \Cref{fig:easy_signal_loss-c} illustrates this spatial quadrant from the same video at the same reduction level but obtained with an upper-corner crop.
In this crop, the visibility of the \emph{Active Object} and \emph{Active Hand} is severely diminished, dropping to only 3\% and 0.6\%, respectively. In contrast, the \emph{Contextual Objects} remain highly visible (79\%), and the feature activation of colour, flicker, intensity, and motion, retain substantial presence (50\%, 40\%, 42\%, and 31\%, respectively). Despite the near-complete loss of the primary action cues, the model maintains a correct prediction, highlighting that contextual and mid-level signals can occasionally compensate for missing object--hand information, albeit in a limited and case-specific manner.

Overall, these findings indicate that AI failures are predominantly driven by the degradation of environmental context such as contextual objects and also mid-level visual features, whereas human errors arise primarily from the loss of high-level semantic information related to the actor and main action.

\paragraph{Recovery (Incorrect $\rightarrow$ Correct)}

When analysing AI recovery cases, reductions in feature values again reveal a counterintuitive pattern. Successful crops do not necessarily maximise available information; instead, reduce it. In recovery scenarios, as shown in \Cref{fig:feature_drop-b}, the AI still suffer substantial loss of the \emph{Active Object} ($\Delta = -0.132$). However, the critical differentiator lies in the preservation of the \emph{Contextual Objects}, for which the reduction is minimal ($\Delta = -0.059$), compared to a substantially larger loss in failure cases ($\Delta = -0.097$). Although numerically small, this difference constitutes a performance \emph{tipping point} for the model.

\begin{figure}[t]
  \centering
    \centering
    \includegraphics[width=\linewidth]{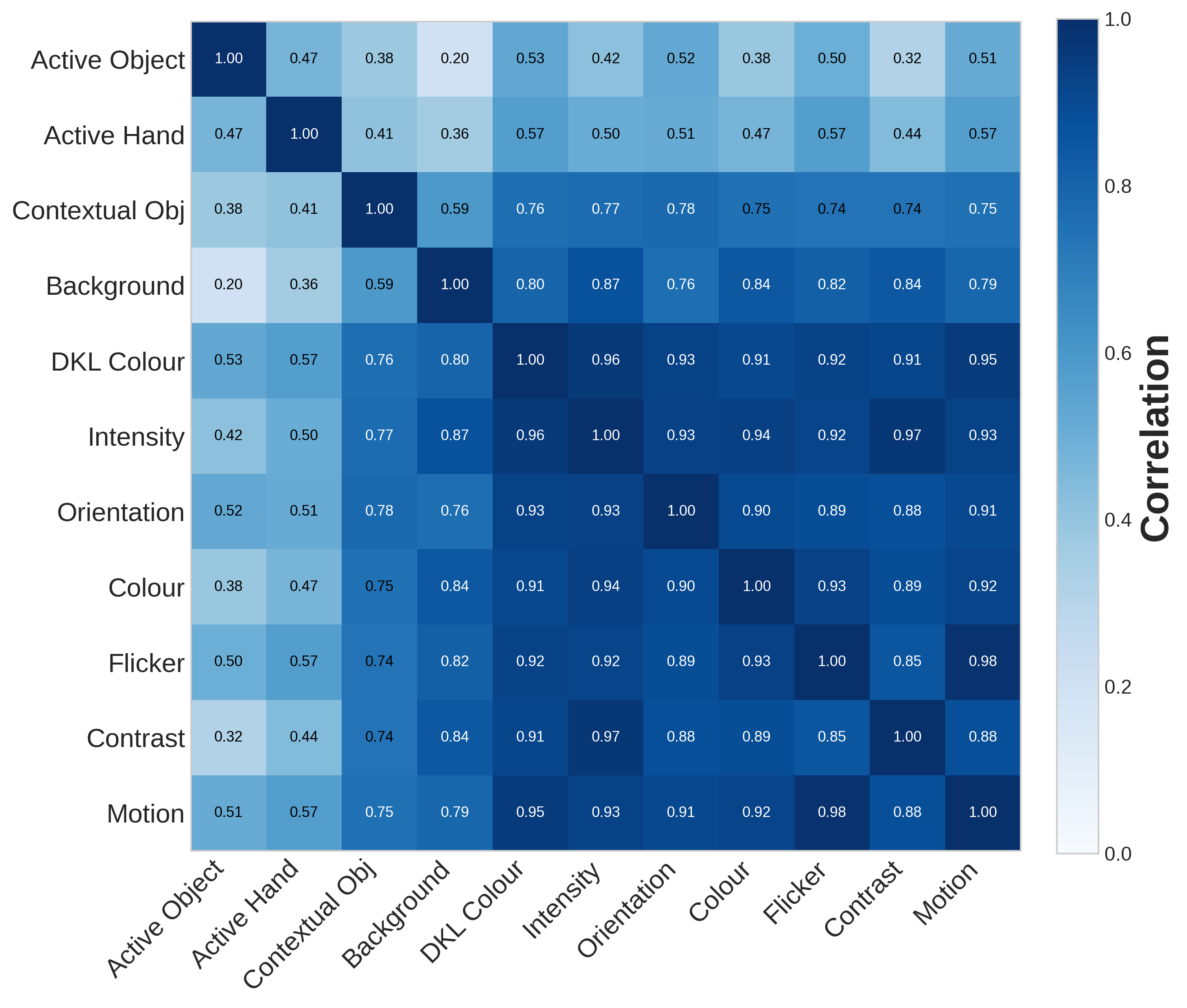}
    \caption{Correlation matrix for AI recovery cases, illustrating the strength of feature correlations across transitions in recovery cases (Incorrect $\rightarrow$ Correct).}
    \label{fig:corr_matrix_recovery}
\end{figure}

The correlation analysis characterises this recovery mechanism as \emph{holistic preservation}. As shown in the correlation matrix in \Cref{fig:corr_matrix_recovery}, successful transitions exhibit a tight cluster of correlations across the transition between \emph{Contextual Objects} and mid-level features such as \emph{Orientation} ($r = 0.78$) and \emph{Intensity} ($r = 0.77$). Crucially, the correlation between \emph{Contextual Objects} and the \emph{Active Object} weakens to $r = 0.38$. This relative independence supports a \emph{pruning} hypothesis: the AI recovers not by retaining all available information, but by selectively preserving the environmental structure (context and orientation) while remaining largely insensitive to the partial loss of the \emph{Active Object}.

While a direct comparison to a human ‘flip to correct’ is not possible, because we did not track the child quadrants of incorrect human classifications and thus did not explicitly measure recovery, the AI model nonetheless exhibits systematic feature amplification and noise suppression.
If our assumption about the relative stability of human recognition failure is true, observers may rely primarily on their initial interpretation of the scene and require the restoration of the \emph{Active Object} to correctly identify the action. By contrast, the AI can recover even when parts of the \emph{Active Object} are missing, provided that the \emph{Contextual Objects} are preserved. This indicates that the AI corrects its predictions by effectively "zooming in" on stable environmental cues rather than tracking the agent’s dynamics. 

\begin{figure}[htbp]
  \centering
  \begin{subfigure}{\linewidth}
    \centering
    \includegraphics[width=\linewidth]{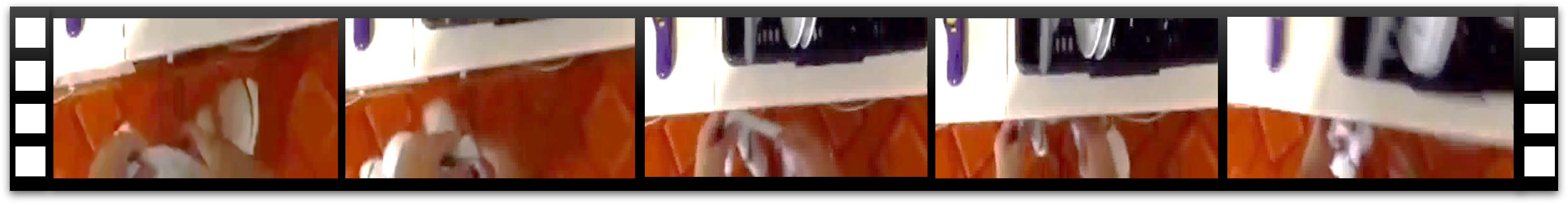}
    \caption{Original video (no reduction), verb correctly predicted as \emph{Put}.}
    \label{fig:hard_recovery-a}
  \end{subfigure}

  \vspace{0.8em}

  \begin{subfigure}{\linewidth}
    \centering
    \includegraphics[width=\linewidth]{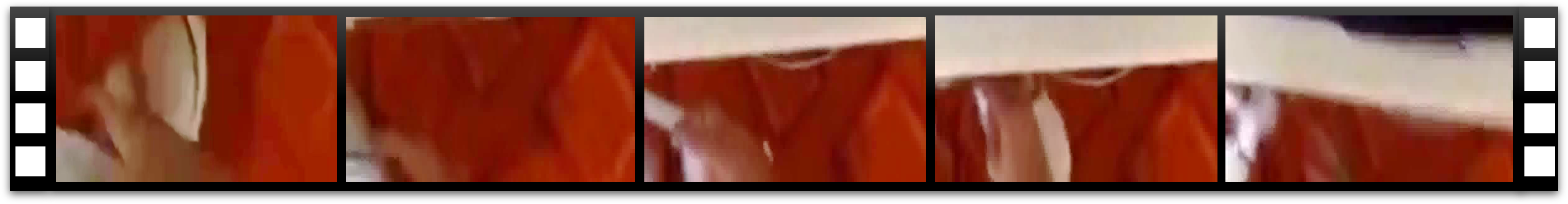}
    \caption{Level 1 (Lower-Right crop), incorrect verb prediction as \emph{Close}. \emph{Active Object} and \emph{Active Hand} presence is 40\% and 49\%.}
    \label{fig:hard_recovery-b}
  \end{subfigure}

  \vspace{0.8em}

  \begin{subfigure}{\linewidth}
    \centering
    \includegraphics[width=\linewidth]{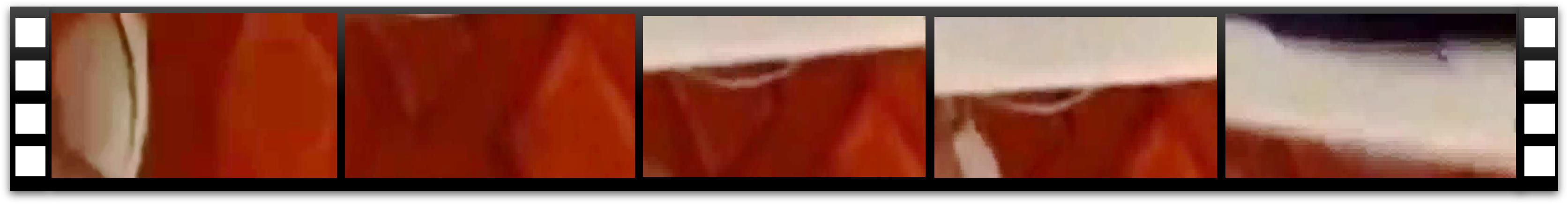}
    \caption{Level 2 (Upper-Right, Lower-Right Crop), verb correctly predicted as \emph{Put}. \emph{Active Object} and \emph{Active Hand} existence both are $\approx 4\%$ while \emph{Contextual Objects} are 34\%}
    \label{fig:hard_recovery-c}
  \end{subfigure}
  \caption{Performance recovery via spatial reduction for a video with substantial \emph{Active Hand} and \emph{Active Object} suppression at Level~2 ($\Delta = -96\%$ compared to the original video). Preserving \emph{Contextual Objects} and other mid-level features enables correct verb (action) recognition (\emph{put}) after the initial misclassification.}
  \label{fig:hard_recovery}
\end{figure}

Similar to \Cref{fig:easy_signal_loss}, \Cref{fig:hard_recovery} illustrates the differences between human and AI behaviour in a recovery scenario for the same representative \emph{put} class.
This transition serves as definitive evidence for \emph{context pruning}. 
In this case, the AI transition from an incorrect (\Cref{fig:hard_recovery-b}) to a correct (\Cref{fig:hard_recovery-c}) action prediction despite a substantial reduction in the \emph{Active Object} ($\Delta = -60\%$) and \emph{Active Hand} ($\Delta = -51\%$). The decisive factor is the visibility of the \emph{Contextual Object}, which remains at 34\% in \Cref{fig:hard_recovery-c}. By removing potentially distracting foreground information, effectively pruning the object, and anchoring its reasoning in stable background cues (e.g., the sink and counter), the AI can refine and correct its prediction.

To further support this claim, consider \Cref{fig:reduced_video} again which is a representative example of the \emph{put} class. In this example, the presence of the \emph{Active Object} declines substantially from MIRC to sub-MIRC, falling to less than 2\%. Although the \emph{Active Hand} remains visible at $\approx29\%$ in sub-MIRCs, this still represents a large decrease of $\approx50\%$ compared to MIRCs. In contrast, \emph{Contextual Objects}, as well as other mid-level features, largely preserve their presence, remaining at $\approx22\%$ and within the 10\%–15\% range, respectively, with only minor reductions compared to the \emph{Active Object} and \emph{Active Hand}. This shift corresponds to an increase in model confidence from 39\% in MIRCs to 56\% in sub-MIRCs, and suggests that the AI model relies heavily on local textures and fine-grained cues which may be amplified through spatial cropping.

Overall, recovery represents an act of \emph{surgical pruning}. The model does not require a clearer view of the action itself to succeed; rather, it only needs an unobstructed view of the environment. Success is driven by the independence of background and foreground information, enabling the AI to reliably reach the correct prediction using stable environmental cues.

\begin{figure}[t]
  \centering
  \begin{subfigure}{\linewidth}
    \centering
    \includegraphics[width=\linewidth]{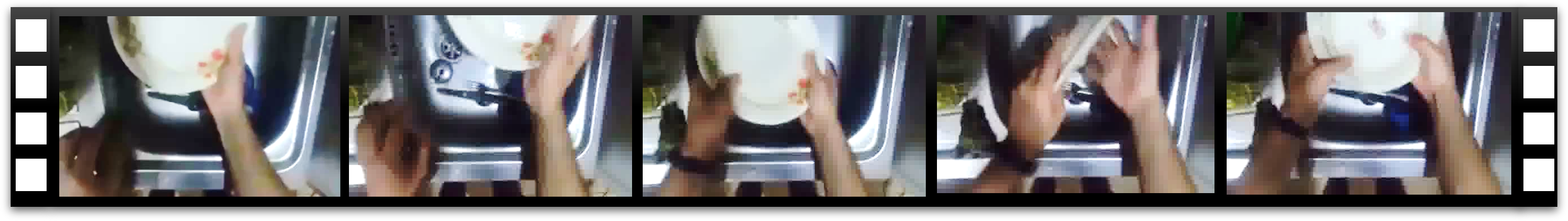}
    \caption{Original video - No reduction - GT label \emph{wash}}
    \label{fig:spatio_temporal_reduced_video-a}
  \end{subfigure}

  \vspace{1em} 

  \begin{subfigure}{\linewidth}
    \centering
    \includegraphics[width=\linewidth]{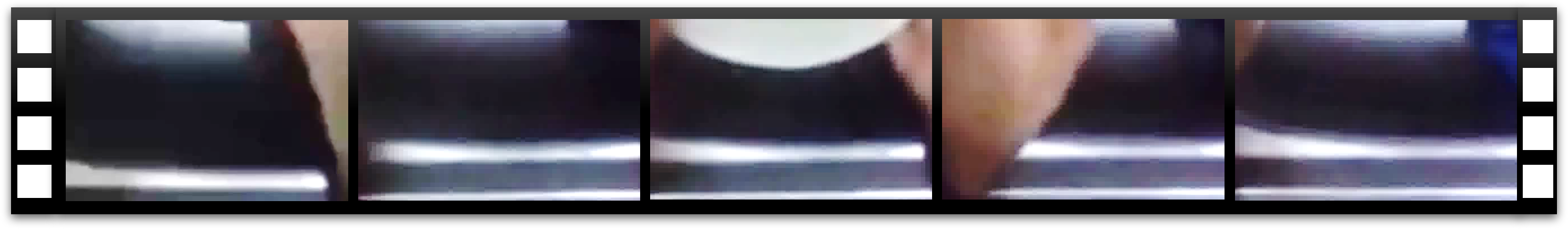}
    \caption{Level 3: The spatial MIRC level}
    \label{fig:spatio_temporal_reduced_video-b}
  \end{subfigure}
  
  \vspace{1em} 

  \begin{subfigure}{\linewidth}
    \centering
    \includegraphics[width=\linewidth]{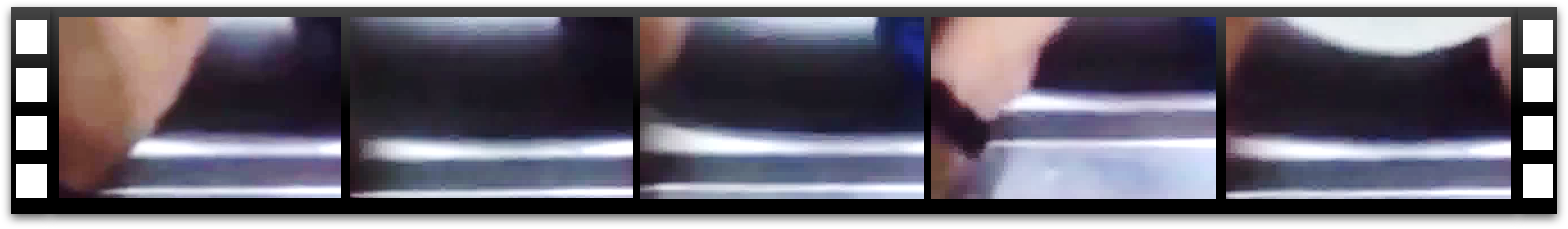}
    \caption{Level 4: The spatiotemporal reduced version of \Cref{fig:spatio_temporal_reduced_video-b}}
    \label{fig:spatio_temporal_reduced_video-c}
  \end{subfigure}

  \caption{An example of temporal scrambling applied to a spatial MIRC. Although the same frames are retained, their temporal ordering differs between \Cref{fig:hard_recovery-b} and \Cref{fig:hard_recovery-c}. As no spatial reduction is applied between the two levels, the observed differences arise solely from changes in temporal structure.}
  \label{fig:spatio_temporal_reduced_video}
\end{figure}

\subsection{Spatiotemporal Results}
\label{subsec:spatiotemporal_results}
This section examines how temporal context interacts with spatial reduction in action recognition for both human observers and the AI model. Overall, humans perform better than the AI model at recognising actions in temporally scrambled videos, correctly identifying 129 out of 474 samples, compared with 48 correct predictions by the AI model. However, this aggregate trend masks more nuanced findings, which we explore in detail in this section. To this end, we apply the same evaluation protocol used in the spatial analysis to the spatiotemporal setting, and report results for both human observers and the S4V model \cite{side4video}. Using the two previously introduced metrics, Recognition Gap and Average Reduction Rate, we analyse differences in behaviour and robustness between humans and the AI model with respect to temporal understanding.

\subsubsection{Quantitative Spatiotemporal Results}
\label{subsubsec:spatiotemporal_quant}

\noindent\textbf{a) Recognition Gap.} We conduct quantitative experiments on spatiotemporal manipulations restricted to MIRC and sub-MIRC videos, as the most substantial changes occur at these stages; all preceding reduction levels are purely spatial and are therefore excluded from this analysis. The recognition gap is computed across all action classes in Epic-ReduAct. A subset of representative classes with varying sample counts is shown in \Cref{fig:recog_gap_per_class-d}–\ref{fig:recog_gap_per_class-f} to facilitate comparison between human observers and the AI model. These figures depict the relative positioning of each MIRC/sub-MIRC pair with respect to the AI model’s confidence and the decision threshold. Consistent with the spatial-only results, the sub-MIRC condition exhibits higher confidence than its corresponding MIRC instance in many cases. This suggests that, similar to spatial reductions, temporal scrambling can sometimes preserve or improve action recognition performance. An example of this behaviour is illustrated by the red (sub-MIRC) and blue (MIRC) points in \Cref{fig:recog_gap_per_class-f}, highlighted by the red rectangle (pair 16), which experienced more than 20\% improvement in confidence. To further examine this case, we visualise the corresponding videos in \Cref{fig:spatio_temporal_reduced_video}. In \Cref{fig:spatio_temporal_reduced_video-b}, salient cues indicating the action \emph{washing} (e.g., the presence of a sink and hand–object interactions) remain visible despite the shuffled frame order. As a result, no major loss of discriminative information is observed, indicating that temporal scrambling can preserve and, in some cases, enhance action recognition for the AI model. Similar patterns are also evident in several other videos across different classes.

\begin{table*}[htbp]
  \centering
  \footnotesize
  \setlength{\tabcolsep}{6pt}
  \begin{tabular}{lcrrrr}
    \toprule
    \textbf{Manipulation Type} & \textbf{Classifier} & \textbf{Min} & \textbf{Max} & \textbf{Std.} & \textbf{Mean} \\
    \midrule
    \multirow{2}{*}{Spatial}
      & AI    & $-$0.425 & +0.315 & +0.112 & $-$0.037 \\
      \cmidrule(lr){2-6}
      & Human & $-$0.050 & +0.800 & +0.159 & +0.373 \\
    \midrule
    \multirow{2}{*}{Spatiotemporal}
      & AI    & $-$0.607 & +0.320 & +0.128 & $-$0.038 \\
      \cmidrule(lr){2-6}
      & Human & $-$0.250 & +0.750 & +0.179 & +0.247 \\
    \bottomrule
  \end{tabular}
  \caption{\textbf{Recognition gap statistics} (percentages) for human and AI classifiers under spatial and spatiotemporal reductions. Positive values indicate reduced recognition performance, while negative values indicate improvement.}
  \label{tab:recog_gap_charts_statistics}
\end{table*}

\begin{table*}[t]
  \centering
  \footnotesize
  \setlength{\tabcolsep}{1.3pt}
  \begin{tabular}{c|c|cccccccccccccc}
    \toprule
    \multicolumn{16}{c}{\textbf{Classes}} \\
    \midrule
    \textbf{Classifier} & \textbf{Metric} 
    & close & cut & hang & insert & open & peel & pour & put 
    & remove & serve & take & turn-off & turn-on & wash \\
    \midrule

    \multirow{2}{*}{Human} 
    & RG  
    & +32.35 & +15.00 & +35.63 & +24.44 & +23.56 & +31.25 & +19.05 & +24.17 
    & +20.00 & +16.88 & +26.69 & +35.82 & +15.00 & +14.41 \\
    & Std. 
    & 16.40 & 12.91 & 19.25 & 8.08 & 16.37 & 16.94 & 15.27 & 17.00 
    & 15.53 & 11.67 & 16.26 & 21.51 & 18.81 & 16.91 \\

    \midrule

    \multirow{2}{*}{AI} 
    & RG  
    & +3.58 & +0.03 & -1.52 & -0.02 & +14.44 & +0.08 & +0.81 & +10.10 
    & -2.51 & 0.00 & +15.51 & +5.66 & +18.76 & -7.52 \\
    & Std. 
    & 10.02 & 0.88 & 2.61 & 2.19 & 14.53 & 0.03 & 0.36 & 7.23 
    & 9.26 & 0.01 & 13.90 & 8.08 & 19.49 & 10.17 \\

    \bottomrule
  \end{tabular}
  \caption{\textbf{Spatiotemporal} Recognition Gaps (RG) and standard deviations (Std) for both classifiers, split by class labels for spatiotemporal MIRC-subMIRC videos. Values are percentages. Positive RG values indicate a reduction in recognition performance under scrambling, whereas negative values indicate an improvement.}
  \label{tab:spatiotemporal_AverageRecognitionGaps}
\end{table*}

\begin{figure}[htbp]
  \centering

    \includegraphics[width=\linewidth]{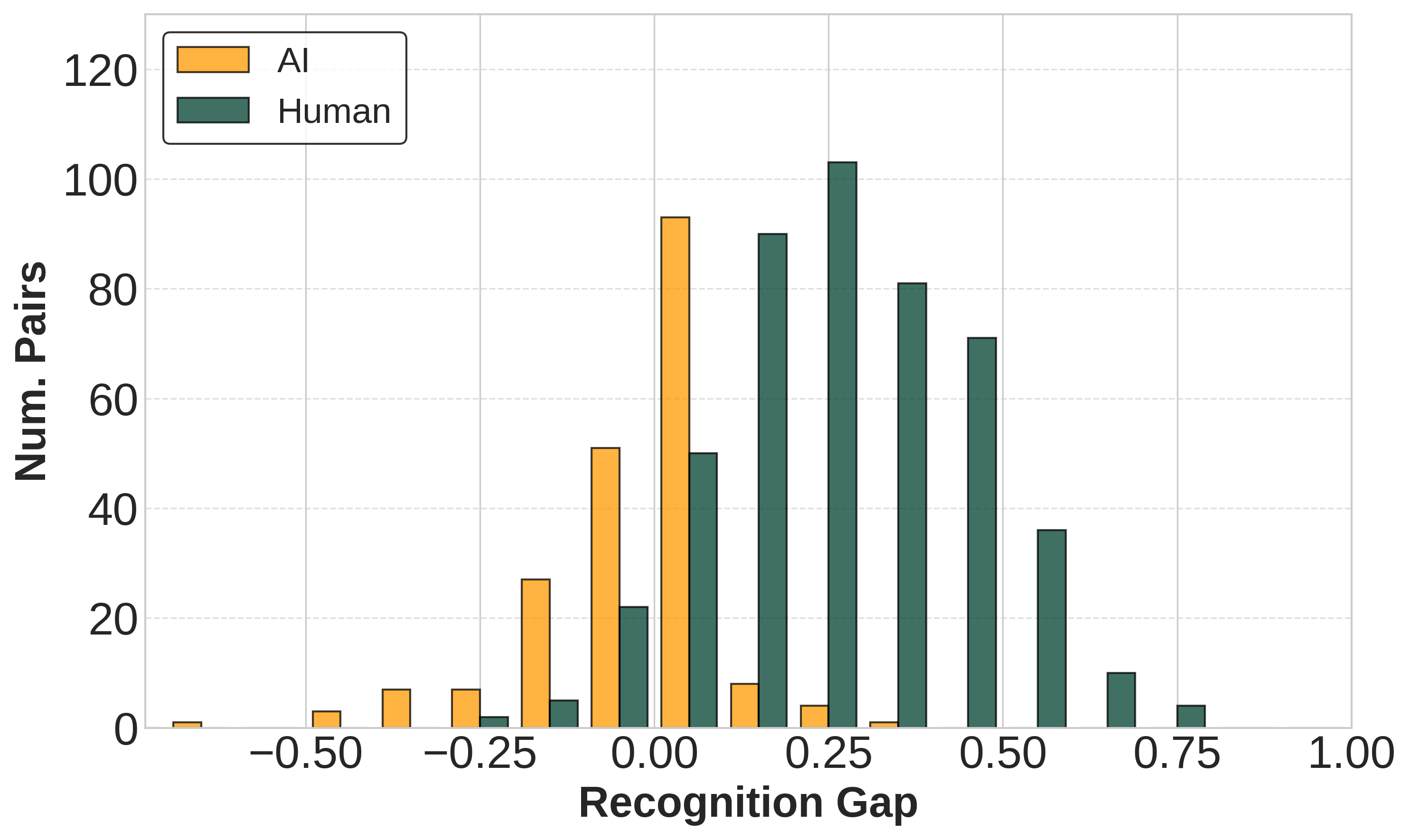}
  \caption{
  Frequency distribution of the recognition gap between humans and the AI model under spatiotemporal reductions.
  }
  \label{fig:recog_gap_spatiotemporal}
\end{figure}

\begin{figure}[htbp]
  \centering
    \includegraphics[width=\linewidth]{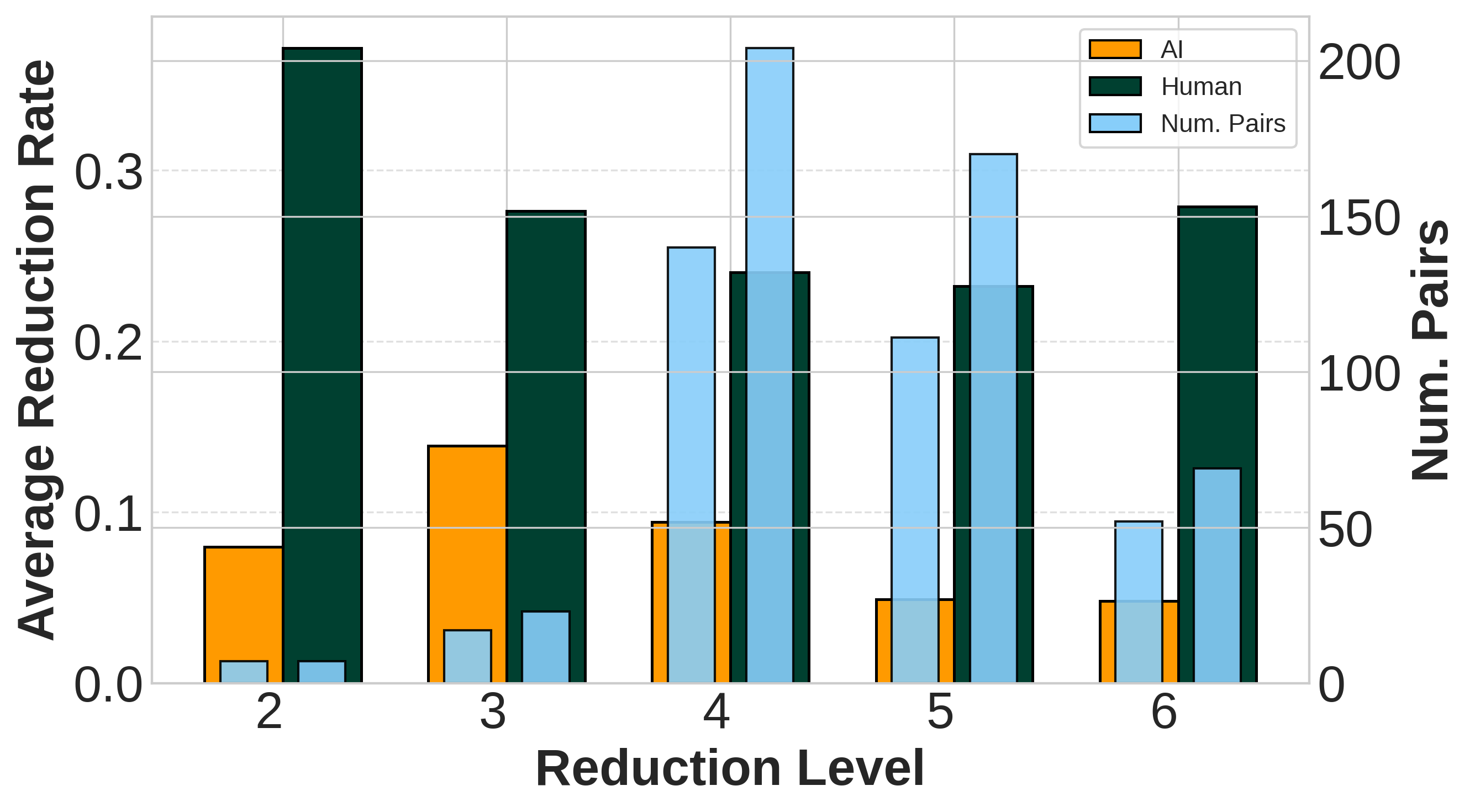}
  
  \caption{Average reduction rate as a function of reduction level for MIRCs and spatiotemporal sub-MIRC pairs, along with the number of pairs in each level of reduction.}
  \label{fig:spatiotemporal_reduc_rate}
\end{figure}

Notably, and less apparent in the per-class scatter plots, Spatiotemporal manipulation can also improve performance for human observers in some instances. This adverse recognition gap effect is reflected in the aggregated results shown in \Cref{fig:recog_gap_spatiotemporal}. 

Consistent with the spatial analysis, AI models exhibit modest performance improvements under spatiotemporal manipulation (\Cref{fig:recog_gap_per_class}); however, human observers also show limited gains following frame scrambling. Despite these improvements, most samples show a decline in recognition performance, with the drop substantially more pronounced for humans than for AI models, as indicated by the higher positive values.

The corresponding frequency distributions further emphasise this disparity: human performance shows a broader spread, reflecting more heterogeneous and occasionally abrupt changes in recognition, whereas AI models exhibit more gradual, concentrated variations. Taken together, these findings indicate that although recent advances in AI have narrowed certain aspects of the human–machine performance gap, substantial differences persist. Notably, based on \Cref{tab:recog_gap_charts_statistics}, which indicates statistics of \Cref{fig:recog_gap} and \Cref{fig:recog_gap_spatiotemporal}, the AI model's mean remains close to zero and slightly negative ($\approx -0.03$), reflecting relative robustness and occasional improvements. A similar pattern holds under spatiotemporal reductions. In contrast, human observers exhibit substantially larger mean recognition gaps across all settings. Specifically, under spatiotemporal manipulations, human recognition degradation is more gradual than under purely spatial reductions, with the mean reduction being approximately 13\% lower (37.3\% for spatial compared with 24.7\% for spatiotemporal) compared to the spatial condition. These contrasts between spatial reduction and temporal scrambling suggest that temporal scrambling is less disruptive to human recognition.

The range and variability statistics reveal further differences between human observers and the AI model. Human performance exhibits a substantially wider spread, with maximum recognition gaps reaching as high as $+0.80$ under spatial reductions and $+0.75$ under spatiotemporal reductions, indicating severe degradation in worst-case scenarios. In contrast, the AI model shows much narrower upper bounds, with maxima remaining below $+0.32$ across both settings. Similarly, the minimum values suggest that while both humans and the AI occasionally benefit from reductions, the AI model experiences larger improvements, particularly under spatiotemporal manipulations (minimum of $-0.61$), compared to humans ($-0.25$). This pattern is further reflected in the standard deviations, which are consistently higher for human observers, indicating greater variability and sensitivity to reduction conditions, whereas the AI model demonstrates more stable behaviour across both spatial and spatiotemporal settings.


Overall, the statistics in Table~\ref{tab:recog_gap_charts_statistics} reinforce two key findings: (i) spatial reductions are more damaging than spatiotemporal scrambling, especially for humans, and (ii) humans exhibit larger degradation and greater variability than AI models.

In a more detailed, class-level consideration, the recognition gaps across all classes for spatiotemporal videos for human and AI models are shown in \Cref{tab:spatiotemporal_AverageRecognitionGaps}. The results indicate that spatiotemporal MIRC and sub-MIRC constraints continue to strongly affect human observers, while the AI model shows more stable or moderately improved performance. Human recognition accuracy drops noticeably across several action classes, with reductions reaching up to 35.82\% for the \emph{turn-off} class or 15.00\% for the \emph{turn-on} class, reflecting the variability in human detection under spatiotemporal distortions. The smallest decline among humans is observed in the \emph{wash} class, with a modest 14.41\% drop, suggesting some robustness to specific actions. In contrast, the AI model demonstrates mostly minor changes in performance. Notably, the AI improves for the \emph{wash} class (-7.52\%) and maintains relatively stable recognition for actions such as \emph{cut}, \emph{serve}, \emph{peel}, \emph{pour}, and \emph{insert}. Even when performance declines, as in the \emph{take} and \emph{turn-on} classes (15.51\% and 14.44\%, respectively), the reductions are considerably smaller than those observed for human observers. 

Variability patterns further amplify this divergence, with the spatiotemporal condition exhibiting higher dispersion than the spatial setting. For humans, standard deviations range from 8.08\% to 21.51\%, with most classes exceeding 15\%, indicating substantial dispersion in how scrambling affects recognition across instances. Particularly high variability is observed for turn-off (21.51\%), hang (19.25\%), and turn-on (18.81\%). In contrast, although the AI model shows a broader spread than in the spatial condition, its variability remains more concentrated overall. Standard deviations range from 0.01\% to 19.49\%, but many classes remain tightly clustered (e.g., peel = 0.03\%, pour = 0.36\%, cut = 0.88\%), suggesting a more uniform response to temporal scrambling.

Overall, the AI model appears less sensitive to spatiotemporal MIRC-subMIRC manipulations than human observers, highlighting differences in how temporal cues are integrated across actions. Comparing the results for spatial (\Cref{tab:AverageRecognitionGaps}) and spatiotemporal (\Cref{tab:spatiotemporal_AverageRecognitionGaps}) MIRC-subMIRC videos reveals several key differences. For both the AI model and humans, recognition gaps are generally smaller in the spatiotemporal condition. This suggests that the lack of temporal continuity features is less destructive than the lack of spatial features.

\begin{table*}[tbp] 
\centering 
\resizebox{
\textwidth}{!}{ 
\begin{footnotesize} 
\begin{tabular}{cp{6.8cm}c|cc|cc} 
\toprule 
\multirow{2}{*}{Verb Category} & 
\multirow{2}{*}{Verbs} & 
\multirow{2}{*}{Num. Pairs} & 
\multicolumn{2}{c}{Human Improved ($\text{Conf. $\Delta$ } < 0$)} & \multicolumn{2}{c}{AI Improved ($\text{Conf. $\Delta$} < 0$)} \\ 
\cmidrule(lr){4-5} \cmidrule(lr){6-7} & & & Count & Percentage & Count & Percentage \\ 
\midrule HTA & opening, closing, turning, putting, taking, pouring, filling/serving, hanging, removing & 406 & 21 & 5.17\% & 106 & 26.11\% \\ 
\midrule LTA & washing, cutting, peeling & 68 & 8 & 11.76\% & 41 & 60.29\% \\ 
\bottomrule 
\end{tabular} 
\end{footnotesize} 
\label{tab:class_temporal_categories} } 
\caption{Action categories grouped into High Temporal Actions (HTA) and Low Temporal Actions (LTA). The table reports the total number of pairs and the proportion of samples in which recognition improved after temporal scrambling for both human observers and the AI model.} 
\label{tab:class_temporal_categories_stats} 
\end{table*}

\hfill 
\hfill 

\noindent\textbf{b) Average Reduction Rate.} \Cref{fig:spatiotemporal_reduc_rate} presents the average reduction rate as a function of reduction level for MIRC/sub-MIRC pairs only, for both the AI model and humans.

In the analysis and comparison, as in the spatial domain, the superimposed bar charts (light blue) indicate that most pairs are concentrated at Levels 4 and 5. Additionally, the Average Reduction Rate trends across spatiotemporal reduction levels are strongly aligned with the spatial results, indicating that reduction levels have the same impact on changes in action understanding. However, the Average Reduction Rate values show that the average reduction is more severe for spatial manipulation for humans. Human reduction rates in the spatial domain consistently exceed those in the spatiotemporal domain, peaking at over 0.45 at Level~2, \Cref{fig:reduc_rate-c}, compared to a peak of approximately 0.37 in the spatiotemporal set, \Cref{fig:spatiotemporal_reduc_rate}.

An interesting finding for the AI model is that, although more samples in spatiotemporal experience improvements or maintain performance, reductions were slightly higher in spatiotemporal pairs, as shown by most of the AI-related (orange) bars. For instance, at Level~3, the AI average reduction rate reaches $\approx 0.14$ in the spatiotemporal (\Cref{fig:spatiotemporal_reduc_rate}), noticeably higher than the $\approx 0.10$ observed in the spatial domain (\Cref{fig:reduc_rate-c}). A similar gap exists at Level~4, where spatiotemporal AI reduction is $\approx 0.10$ compared to $\approx 0.07$ for spatial.

\subsubsection{Qualitative Spatiotemporal Analysis}
\label{subsubsec:spatiotemporal_qual}

We replicated the qualitative analysis used for spatial reductions in the spatiotemporal setting. However, this approach proved uninformative, as our high-level features were identical between spatial MIRC and scrambled sub-MIRC pairs. In contrast, mid-level features yielded negligible differences of $-4.48 \times 10^{-4}$ to $1.08 \times 10^{-3}$. These negligible differences confirm that the paired videos are effectively indistinguishable in spatial content, isolating temporal structure as the primary factor driving any recognition differences.

\begin{figure}[htbp]
  \centering
  \begin{subfigure}{0.49\linewidth}
    \includegraphics[width=\linewidth]{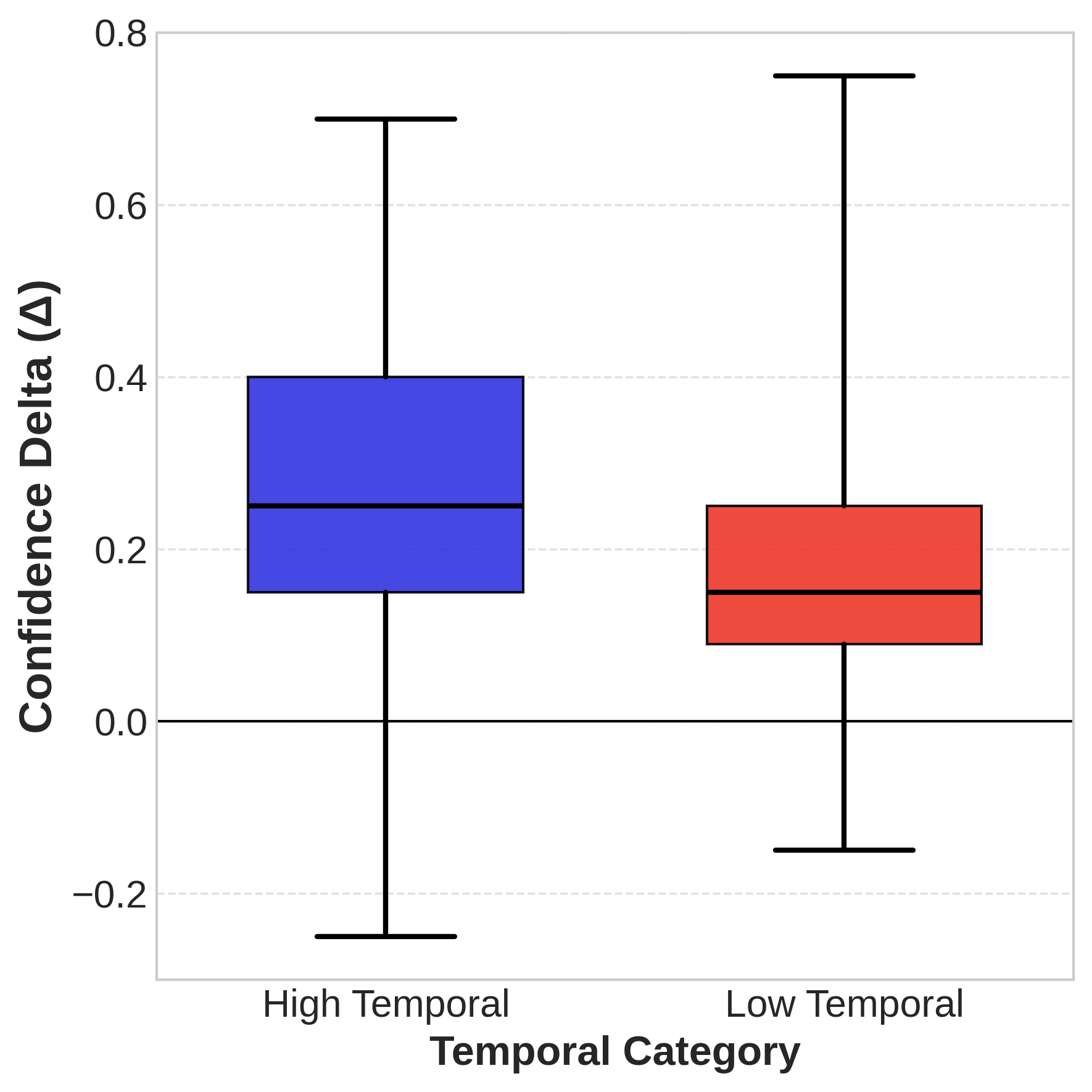}
    \caption{Humans}
    \label{fig:spatiotemporal_conf_delta_candle-a}
  \end{subfigure}
  \hfill
  \begin{subfigure}{0.49\linewidth}
    \includegraphics[width=\linewidth]{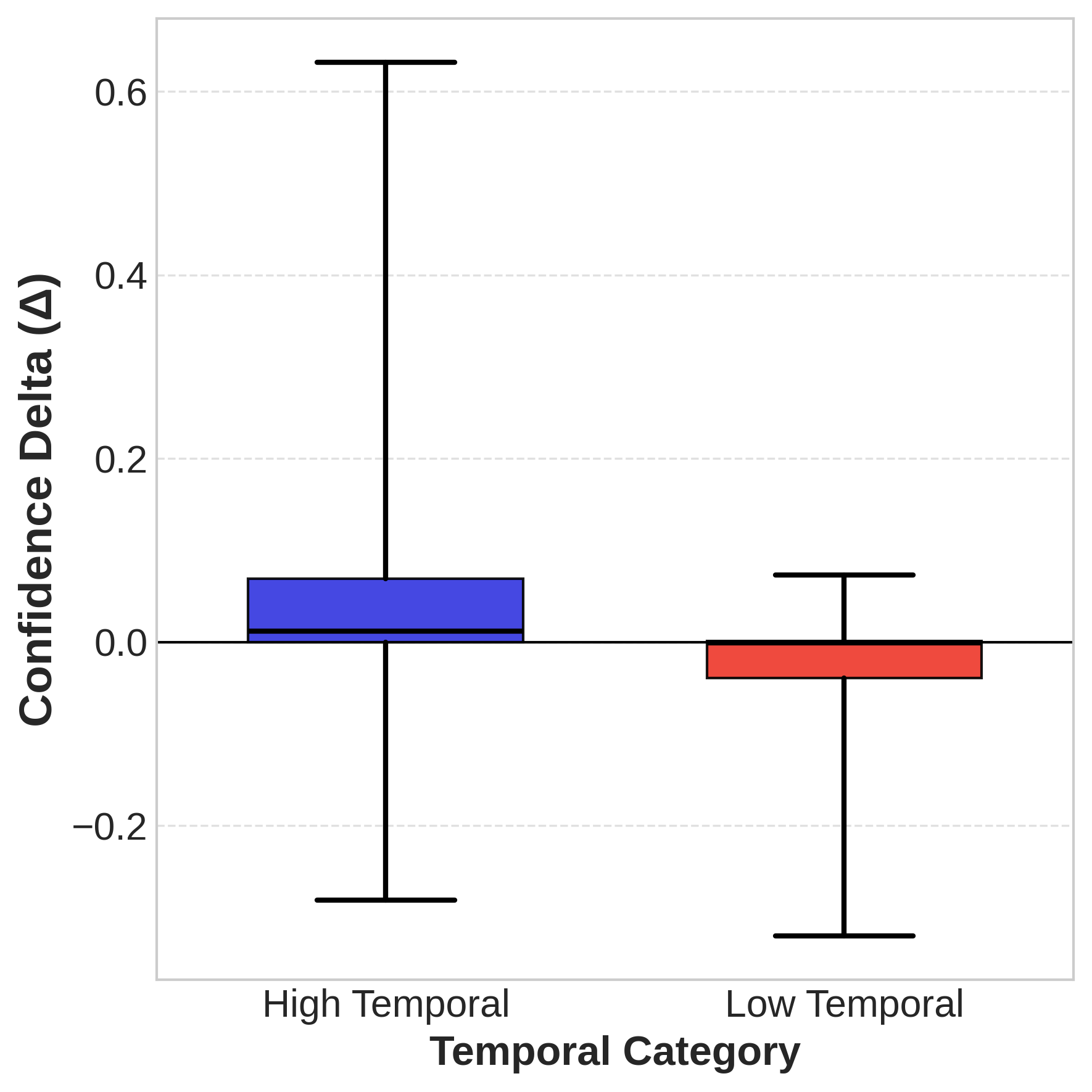}
    \caption{AI Model}
    \label{fig:spatiotemporal_conf_delta_candle-b}
  \end{subfigure}
  \caption{Mean difference of Humans and AI Model prediction confidence between each MIRC and its spatiotemporal sub-MIRC.}
  \label{fig:spatiotemporal_conf_delta_candle}
\end{figure}

We further examined class-level differences in classification outcomes following temporal scrambling. The results revealed a clear class-dependent structure. Based on this observation, we partitioned action classes into two verb categories: \emph{Low Temporal Actions} (LTA) and \emph{High Temporal Actions} (HTA), as summarised in Table~\ref{tab:class_temporal_categories_stats}. The analysis shows that HTA ($\approx 26\%$) yields substantially less improvement over scrambling for the AI model compared to LTA ($\approx 60\%$). A similar pattern is observed for human predictions; however, the magnitude of improvement is smaller. \Cref{fig:spatiotemporal_conf_delta_candle-a} shows that the confidence delta for MIRC vs sub-MIRC recognition is higher for HTA than LTA classes, indicating that temporal scrambling not only degrades human accuracy for HTA classes but also leads to a marked reduction in subjective certainty. In contrast, LTA classes exhibit partial robustness to temporal perturbations.

In contrast, the AI model exhibits markedly different behaviour. For the model, the recognition gap is approximately $5\%$ for HTA classes and $-3\%$ for LTA classes, suggesting that temporal scrambling has little impact on HTA recognition and may even improve performance on LTA classes. This pattern is mirrored in the confidence distributions shown in \Cref{fig:spatiotemporal_conf_delta_candle-b}, where confidence deltas are tightly concentrated around zero, with minimal separation between HTA and LTA classes. Notably, LTA classes exhibit a slight negative shift in median confidence, indicating increased model confidence despite temporal disruption. A $t$-test comparing the mean recognition gaps between LTA and HTA classes yields $p = 3 \times 10^{-4}$ for human observers and $p = 10^{-14}$ for the AI model. To ensure a more robust analysis and account for multiple clips originating from the same source video, we aggregate responses by unique video ID ($N = 36$) before comparing the groups, thereby allowing us to generalise the findings to new video IDs with the same action verbs. Under this stricter evaluation, the difference remains statistically significant for the AI model ($p = 0.001$), confirming that the inverted response to LTA scrambling is a robust model characteristic rather than an artefact of specific video instances. Conversely, the human response shows a trend toward greater temporal reliance for HTA classes, but this difference is not statistically significant ($p = 0.06$) due to the reduced sample size. Ultimately, while humans exhibit a graded and interpretable response to temporal disruption across action categories, the AI treats LTA and HTA classes as fundamentally distinct recognition problems, largely invariant to temporal coherence.

\subsection{Limitations}
While our study provides a detailed characterisation of human--AI differences in ego-centric action recognition, several aspects also point to promising directions for future work. First, although the human behavioural study is large-scale, the high cost of experiments necessarily constrains the number of unique base videos. As a result, our findings are most directly applicable to the balanced subset of ego-centric actions examined here, and future studies could extend this framework to a broader range of activities. Second, our analyses focused on a single domain: ego-centric kitchen activities from EPIC-KITCHENS-100 \cite{EPICKITCHENS}, which are highly structured and object-centric. Extending the present paradigm to other ego-centric scenarios, or to third-person action recognition settings with richer global context, would provide a valuable test of the robustness of the observed human--AI differences across domains. Third, on the model side, we analysed a single state-of-the-art video architecture trained under a specific regime. Applying the same MIRC-based reductions and spatiotemporal manipulations to a broader family of architectures and training objectives, including multimodal or more explicitly predictive models, would help to clarify which of the observed discrepancies reflect general limitations of current video models and which are specific to particular design choices.

%% file: sec/5_conclusion.tex
\section{Discussion and Conclusion}
\label{sec:conclusion}

In this study we compared human and AI performance in ego-centric video action recognition, focusing on the contributions of mid- to high-level visual features to recognition. We used our new Epic-ReduAct dataset \cite{humanvsmachine, Rybansky2026.02.15.705896}, which contained a large number of video clips with systematically reduced spatial and temporal information. This dataset was derived from the EPIC-KITCHENS-100 dataset \cite{EPICKITCHENS}. Our results reveal key differences in how humans and a state-of-the-art video model use spatial and spatiotemporal cues, and provide concrete signals for how future models might be made more efficient and human-aligned.

\subsection{Spatial cues and human--AI divergence}
Our spatial analyses reveal a fundamental divergence between human observers and a state-of-the-art AI model in how minimal visual evidence is used to recognise actions. Humans exhibit a sharp recognition boundary at the transition from Minimal Recognisable Configurations (MIRCs) to sub-MIRCs, indicating a strong reliance on a small set of semantically critical spatial cues, particularly the configuration of the active hand--object interaction. Once these cues are removed, human recognition performance collapses abruptly.

By contrast, the evaluated AI model shows a markedly more gradual degradation and, in many cases, even improved recognition under spatial reduction. This behaviour suggests that the model can benefit from removing potentially distracting information and relies primarily on distributed environmental context and mid-level visual statistics, rather than on the core semantic structure of the action. These findings demonstrate that strong performance on full-resolution benchmarks does not imply human-like spatial reasoning, and that recognition from minimal information is governed by qualitatively different strategies in humans and current models.

At the same time, our results suggest practical directions for improving the efficiency and performance of AI models. In particular, by incorporating either contextual objects or, human-relevant cues, such as active objects identified via human behaviour, as an auxiliary supervisory signal, we can guide models toward more meaningful representations. By explicitly encouraging models to attend to the same critical areas used by humans, it may be possible to shift model reliance away from incidental mid-level statistics and towards more semantic, human-aligned features, ultimately improving both robustness and interpretability.

\subsection{Spatiotemporal cues and temporal robustness}
Extending the MIRC framework to the temporal domain reveals that temporal structure plays different roles for humans and AI models in ego-centric action recognition. Humans remain robust to moderate temporal scrambling when critical spatial cues are preserved, indicating an ability to infer actions from partial, unordered, or fragmented motion information. While temporal scrambling is generally less disruptive than spatial reduction for both humans and the AI model, humans exhibit larger and more variable performance drops across action classes. In contrast, the AI model shows limited sensitivity to temporal disruption and occasionally benefits from scrambling, suggesting reliance on static contextual anchors and short-range motion statistics rather than coherent temporal reasoning. Overall, these results indicate that spatial information constitutes the primary bottleneck for human recognition, whereas current video models do not exploit temporal structure in ways that align with human perception. This insight also points to opportunities for more efficient action recognition by reducing unnecessary temporal redundancy in current approaches. Many existing models rely on large numbers of frames (for example, up to 100) to predict actions, even though such sequences often contain repetitive motion patterns that can be reliably recognised from only a few informative frames. Our proposed categorisation of actions into Low Temporal Actions (LTA) and High Temporal Actions (HTA), therefore, provides a principled guideline for limiting the amount of temporal input fed into the model, improving computational efficiency without sacrificing recognition performance. However, this can be further explored by applying more disruptive or varied forms of temporal scrambling to better understand the effects of different temporal manipulations.

\subsection{Future work}
Building on these findings and limitations, several avenues for future work emerge. A first direction is to use Epic-ReduAct and the MIRC framework as a training signal rather than only as an evaluation tool. For example, AI models could be explicitly optimised to maintain performance across MIRC and sub-MIRC transitions, encouraging them to rely on the same semantically critical cues as human observers. 
A further promising line of work is to extend the proposed reduction and scrambling paradigm to a broader family of architectures and learning setups, including multimodal video language models and more explicitly predictive models. Systematically comparing how such models behave under controlled spatial and spatiotemporal reductions would help to disentangle architectural from data-driven effects and guide the development of architectures whose internal representations and failure modes more closely mirror human perception.

More broadly, we see MIRC-based spatial and spatiotemporal manipulations as a general toolbox for probing and shaping action recognition systems. Applying this toolbox across datasets and domains has the potential to yield models that are not only accurate on benchmarks, but also better aligned with human strategies for recognising actions in complex, ego-centric environments.

%% file: main.bbl
\begin{thebibliography}{}

\bibitem[Ahissar and Hochstein, 2004]{AHISSAR2004457}
Ahissar, M. and Hochstein, S. (2004).
\newblock The reverse hierarchy theory of visual perceptual learning.
\newblock {\em Trends in Cognitive Sciences}, 8(10):457--464.

\bibitem[Ahmadian et~al., 2023]{mofo}
Ahmadian, M., Guerin, F., and Gilbert, A. (2023).
\newblock Mofo: Motion focused self-supervision for video understanding.
\newblock {\em NeurIPS 2023 Workshop Self-Supervised Learning: Theory and Practice}.

\bibitem[Ahmadian et~al., 2024]{ahmadian2024filsselfsupervisedvideofeature}
Ahmadian, M., Guerin, F., and Gilbert, A. (2024).
\newblock Fils: Self-supervised video feature prediction in semantic language space.
\newblock {\em British Machine Vision Conference (BMVC’24)}.

\bibitem[Alcorn et~al., 2019]{8954212}
Alcorn, M.~A., Li, Q., Gong, Z., Wang, C., Mai, L., Ku, W.-S., and Nguyen, A. (2019).
\newblock Strike (with) a pose: Neural networks are easily fooled by strange poses of familiar objects.
\newblock In {\em 2019 IEEE/CVF Conference on Computer Vision and Pattern Recognition (CVPR)}, pages 4840--4849.

\bibitem[Anwyl-Irvine et~al., 2020]{Anwyl-Irvine2020}
Anwyl-Irvine, A.~L., Massonnié, J., Flitton, A., Kirkham, N., and Evershed, J.~K. (2020).
\newblock Gorilla in our midst: An online behavioral experiment builder.
\newblock {\em Behavior Research Methods}, 52:388--407.

\bibitem[Baker et~al., 2018]{Baker2018}
Baker, N., Lu, H., Erlikhman, G., and Kellman, P.~J. (2018).
\newblock Deep convolutional networks do not classify based on global object shape.
\newblock {\em PLOS Computational Biology}, 14(12):e1006613.

\bibitem[Bar, 2003]{089892903321662976}
Bar, M. (2003).
\newblock A cortical mechanism for triggering top-down facilitation in visual object recognition.
\newblock {\em Journal of Cognitive Neuroscience}, 15(4):600--609.

\bibitem[Bar et~al., 2006]{doi:10.1073/pnas.0507062103}
Bar, M., Kassam, K.~S., Ghuman, A.~S., Boshyan, J., Schmid, A.~M., Dale, A.~M., Hämäläinen, M.~S., Marinkovic, K., Schacter, D.~L., Rosen, B.~R., and Halgren, E. (2006).
\newblock Top-down facilitation of visual recognition.
\newblock {\em Proceedings of the National Academy of Sciences}, 103(2):449--454.

\bibitem[Ben-Yosef et~al., 2018]{A_model_for_full_local_image_interpretation}
Ben-Yosef, G., Assif, L., and Ullman, S. (2018).
\newblock Full interpretation of minimal images.
\newblock {\em Cognition}, 171:65 -- 84.

\bibitem[Ben-Yosef et~al., 2020]{Minimal_videos_Trade-off_between_spatial_and_temporal_information_in_human_and_machine_vision}
Ben-Yosef, G., Kreiman, G., and Ullman, S. (2020).
\newblock Minimal videos: Trade-off between spatial and temporal information in human and machine vision.
\newblock {\em Cognition}, 201:104263.

\bibitem[Borghi et~al., 2012]{BORGHI201264}
Borghi, A.~M., Flumini, A., Natraj, N., and Wheaton, L.~A. (2012).
\newblock One hand, two objects: Emergence of affordance in contexts.
\newblock {\em Brain and Cognition}, 80(1):64--73.

\bibitem[Bowers et~al., 2023]{1231244}
Bowers, J.~S., Malhotra, G., Dujmovi{\'c}, M., Llera~Montero, M., Tsvetkov, C., Biscione, V., Puebla, G., Adolfi, F., Hummel, J.~E., Heaton, R.~F., and et~al. (2023).
\newblock Deep problems with neural network models of human vision.
\newblock {\em Behavioral and Brain Sciences}, 46:e385.

\bibitem[Campanella et~al., 2011]{campanella2011visual}
Campanella, F., Sandini, G., and Morrone, M.~C. (2011).
\newblock Visual information gleaned by observing grasping movement in allocentric and egocentric perspectives.
\newblock {\em Proceedings of the Royal Society B: Biological Sciences}, 278(1715):2142--2149.

\bibitem[Carrasco et~al., 2020]{Laconic_Image_Classification_Human_vs_Machine_Performance}
Carrasco, J., Hogan, A., and P\'{e}rez, J. (2020).
\newblock Laconic image classification: Human vs. machine performance.
\newblock In {\em Proceedings of the 29th ACM International Conference on Information \& Knowledge Management}, CIKM '20, page 115–124, New York, NY, USA. Association for Computing Machinery.

\bibitem[Damen et~al., 2018]{EPICKITCHENS}
Damen, D., Doughty, H., Farinella, G.~M., Fidler, S., Furnari, A., Kazakos, E., Moltisanti, D., Munro, J., Perrett, T., Price, W., and Wray, M. (2018).
\newblock Scaling egocentric vision: The epic-kitchens dataset.
\newblock {\em European Conference on Computer Vision (ECCV)}.

\bibitem[Dong et~al., 2022]{dong2022viewfoolevaluatingrobustnessvisual}
Dong, Y., Ruan, S., Su, H., Kang, C., Wei, X., and Zhu, J. (2022).
\newblock Viewfool: Evaluating the robustness of visual recognition to adversarial viewpoints.
\newblock In {\em NeurIPS'22}.

\bibitem[Dosovitskiy et~al., 2021]{vit}
Dosovitskiy, A., Beyer, L., Kolesnikov, A., Weissenborn, D., Zhai, X., Unterthiner, T., Dehghani, M., Minderer, M., Heigold, G., Gelly, S., Uszkoreit, J., and Houlsby, N. (2021).
\newblock An image is worth 16x16 words: Transformers for image recognition at scale.
\newblock In {\em International Conference on Learning Representations}.

\bibitem[Dyck and Gruber, 2020]{Seeing_Eye-to-Eye_A_Comparison_of_Object_Recognition_Performance}
Dyck, L. and Gruber, W. (2020).
\newblock Seeing eye-to-eye?: A comparison of object recognition performance in humans and deep convolutional neural networks under image manipulation.
\newblock Workingpaper, University of Salzburg.
\newblock 19 pages, 7 figures, 3 tables.

\bibitem[Fel et~al., 2022]{Harmonizing_the_object_recognition_strategies_of_deep_neural_networks_with_humans}
Fel, T., Felipe, I., Linsley, D., and Serre, T. (2022).
\newblock Harmonizing the object recognition strategies of deep neural networks with humans.
\newblock {\em Advances in neural information processing systems}, 35:9432--9446.

\bibitem[Fonaryov and Lindenbaum, 2020]{On_the_Minimal_Recognizable_Image_Patch}
Fonaryov, M. and Lindenbaum, M. (2020).
\newblock On the minimal recognizable image patch.
\newblock {\em CoRR}, abs/2010.05858.

\bibitem[Garbe, 2012]{Garbe2012}
Garbe, W. (2012).
\newblock Symspell: Spelling correction \& fuzzy search - 1 million times faster through symmetric delete spelling correction algorithm.

\bibitem[Geirhos et~al., 2020a]{DBLP:journals/corr/abs-2004-07780}
Geirhos, R., Jacobsen, J.-H., Michaelis, C., Zemel, R., Brendel, W., Bethge, M., and Wichmann, F.~A. (2020a).
\newblock Shortcut learning in deep neural networks.
\newblock {\em Nature Machine Intelligence}, 2(11):665--673.

\bibitem[Geirhos et~al., 2020b]{DBLP:journals/corr/abs-2006-16736}
Geirhos, R., Meding, K., and Wichmann, F.~A. (2020b).
\newblock Beyond accuracy: quantifying trial-by-trial behaviour of cnns and humans by measuring error consistency.
\newblock In {\em Proceedings of the 34th International Conference on Neural Information Processing Systems}, NIPS '20, Red Hook, NY, USA. Curran Associates Inc.

\bibitem[Geirhos et~al., 2021]{DBLP:journals/corr/abs-2106-07411}
Geirhos, R., Narayanappa, K., Mitzkus, B., Thieringer, T., Bethge, M., Wichmann, F.~A., and Brendel, W. (2021).
\newblock Partial success in closing the gap between human and machine vision.
\newblock In Ranzato, M., Beygelzimer, A., Dauphin, Y., Liang, P., and Vaughan, J.~W., editors, {\em Advances in Neural Information Processing Systems}, volume~34, pages 23885--23899. Curran Associates, Inc.

\bibitem[Geirhos et~al., 2019]{DBLP:journals/corr/abs-1811-12231}
Geirhos, R., Rubisch, P., Michaelis, C., Bethge, M., Wichmann, F.~A., and Brendel, W. (2019).
\newblock Imagenet-trained {CNN}s are biased towards texture; increasing shape bias improves accuracy and robustness.
\newblock In {\em International Conference on Learning Representations}.

\bibitem[Giese and Poggio, 2003]{Neural_mechanisms_for_the_recognition_of_biological_movements}
Giese, M.~A. and Poggio, T. (2003).
\newblock Neural mechanisms for the recognition of biological movements.
\newblock {\em Nature Reviews Neuroscience}, 4(3):179--192.

\bibitem[Goodale and Milner, 1992]{GOODALE199220}
Goodale, M.~A. and Milner, A. (1992).
\newblock Separate visual pathways for perception and action.
\newblock {\em Trends in Neurosciences}, 15(1):20--25.

\bibitem[Gruber et~al., 2021]{Oculo-retinal_dynamics_can_explain_the_perception_of_minimal_recognizable_configuration}
Gruber, L.~Z., Ullman, S., and Ahissar, E. (2021).
\newblock Oculo-retinal dynamics can explain the perception of minimal recognizable configurations.
\newblock {\em Proceedings of the National Academy of Sciences}, 118.

\bibitem[Güçlü and van Gerven, 2015]{10005}
Güçlü, U. and van Gerven, M. A.~J. (2015).
\newblock Deep neural networks reveal a gradient in the complexity of neural representations across the ventral stream.
\newblock {\em Journal of Neuroscience}, 35(27):10005--10014.

\bibitem[Harari et~al., 2020]{What_Takes_the_Brain_so_Long_Object_Recognition_at_the_Level_of_Minimal_Images_Develops_for_up}
Harari, D., Benoni, H., and Ullman, S. (2020).
\newblock Object recognition at the level of minimal images develops for up to seconds of presentation time.
\newblock {\em Journal of Vision}, 20(11):266--266.

\bibitem[Harel et~al., 2006]{NIPS2006_4db0f8b0_Graph_Based_Visual_Saliency}
Harel, J., Koch, C., and Perona, P. (2006).
\newblock Graph-based visual saliency.
\newblock In Sch\"{o}lkopf, B., Platt, J., and Hoffman, T., editors, {\em Advances in Neural Information Processing Systems}, volume~19. MIT Press.

\bibitem[He et~al., 2016]{resnet}
He, K., Zhang, X., Ren, S., and Sun, J. (2016).
\newblock {Deep Residual Learning for Image Recognition}.
\newblock In {\em Proceedings of 2016 IEEE Conference on Computer Vision and Pattern Recognition}, CVPR '16, pages 770--778. IEEE.

\bibitem[Henderson et~al., 2008]{HENDERSON200840}
Henderson, J.~M., Larson, C.~L., and Zhu, D.~C. (2008).
\newblock Full scenes produce more activation than close-up scenes and scene-diagnostic objects in parahippocampal and retrosplenial cortex: An fmri study.
\newblock {\em Brain and Cognition}, 66(1):40--49.

\bibitem[Hochreiter and Schmidhuber, 1997]{hochreiter1997long}
Hochreiter, S. and Schmidhuber, J. (1997).
\newblock Long short-term memory.
\newblock {\em Neural computation}, 9(8):1735--1780.

\bibitem[Hochstein and Ahissar, 2002]{HOCHSTEIN2002791}
Hochstein, S. and Ahissar, M. (2002).
\newblock View from the top: Hierarchies and reverse hierarchies in the visual system.
\newblock {\em Neuron}, 36(5):791--804.

\bibitem[Hu et~al., 2022]{If_a_Human_Can_See_It_So_Should_Your_System_Reliability_Requirements_for_Machine_Vision_Component}
Hu, B.~C., Marsso, L., Czarnecki, K., Salay, R., Shen, H., and Chechik, M. (2022).
\newblock If a human can see it, so should your system: reliability requirements for machine vision components.
\newblock In {\em Proceedings of the 44th International Conference on Software Engineering}, ICSE '22, page 1145–1156, New York, NY, USA. Association for Computing Machinery.

\bibitem[Hubel and Wiesel, 1959]{HubelD.H.1959Rfos}
Hubel, D.~H. and Wiesel, T.~N. (1959).
\newblock Receptive fields of single neurones in the cat's striate cortex.
\newblock {\em The Journal of physiology}, 148(3):574--591.

\bibitem[Jacob et~al., 2021]{Jacob2021}
Jacob, G., Pramod, R.~T., Katti, H., and Arun, S.~P. (2021).
\newblock Qualitative similarities and differences in visual object representations between brains and deep networks.
\newblock {\em Nature Communications}, 12(1):1872.

\bibitem[Jang et~al., 2021]{jang2021noise}
Jang, H., McCormack, D., and Tong, F. (2021).
\newblock Noise-trained deep neural networks effectively predict human vision and its neural responses to challenging images.
\newblock {\em PLoS Biology}, 19(12):e3001418.

\bibitem[Jhuang et~al., 2007]{A_Biologically_Inspired_System_for_Action_Recognition}
Jhuang, H., Serre, T., Wolf, L., and Poggio, T. (2007).
\newblock A biologically inspired system for action recognition.
\newblock In {\em 2007 IEEE 11th International Conference on Computer Vision}, pages 1--8.

\bibitem[Kheradpisheh et~al., 2016]{Bio_inspired_unsupervised_learning_of_visual_features_leads}
Kheradpisheh, S.~R., Ganjtabesh, M., and Masquelier, T. (2016).
\newblock Bio-inspired unsupervised learning of visual features leads to robust invariant object recognition.
\newblock {\em Neurocomputing}, 205:382--392.

\bibitem[Krizhevsky et~al., 2012]{alexnet}
Krizhevsky, A., Sutskever, I., and Hinton, G.~E. (2012).
\newblock Imagenet classification with deep convolutional neural networks.
\newblock In Pereira, F., Burges, C., Bottou, L., and Weinberger, K., editors, {\em Advances in Neural Information Processing Systems}, volume~25. Curran Associates, Inc.

\bibitem[Lecun et~al., 1998]{726791}
Lecun, Y., Bottou, L., Bengio, Y., and Haffner, P. (1998).
\newblock Gradient-based learning applied to document recognition.
\newblock {\em Proceedings of the IEEE}, 86(11):2278--2324.

\bibitem[Levi, 2008]{LEVI2008635}
Levi, D.~M. (2008).
\newblock Crowding—an essential bottleneck for object recognition: A mini-review.
\newblock {\em Vision Research}, 48(5):635--654.

\bibitem[Libby and Eibach, 2011]{libby2011self}
Libby, L.~K. and Eibach, R.~P. (2011).
\newblock How the self affects and reflects the content and subjective experience of autobiographical memory.
\newblock In {\em The Self}, pages 75--91. Psychology Press.

\bibitem[Lindsay, 2015]{DBLP:journals/corr/Lindsay15}
Lindsay, G.~W. (2015).
\newblock Feature-based attention in convolutional neural networks.
\newblock {\em CoRR}, abs/1511.06408.

\bibitem[Livingstone and Hubel, 1988]{doi:10.1126/science.3283936}
Livingstone, M. and Hubel, D. (1988).
\newblock Segregation of form, color, movement, and depth: Anatomy, physiology, and perception.
\newblock {\em Science}, 240(4853):740--749.

\bibitem[Loshchilov and Hutter, 2019]{adam}
Loshchilov, I. and Hutter, F. (2019).
\newblock Decoupled weight decay regularization.
\newblock In {\em International Conference on Learning Representations}.

\bibitem[Loucks and Baldwin, 2009]{LOUCKS200984}
Loucks, J. and Baldwin, D. (2009).
\newblock Sources of information for discriminating dynamic human actions.
\newblock {\em Cognition}, 111(1):84--97.

\bibitem[Maaz et~al., 2024]{VideoChatGPT}
Maaz, M., Rasheed, H., Khan, S., and Khan, F.~S. (2024).
\newblock Video-chatgpt: Towards detailed video understanding via large vision and language models.
\newblock In {\em Proceedings of the 62nd Annual Meeting of the Association for Computational Linguistics (ACL 2024)}.

\bibitem[Malhotra et~al., 2022]{Malhotra2022}
Malhotra, G., Dujmovi{\'c}, M., and Bowers, J.~S. (2022).
\newblock Feature blindness: a challenge for understanding and modelling visual object recognition.
\newblock {\em PLOS Computational Biology}, 18(5):e1009572.

\bibitem[Masquelier et~al., 2010]{Learning_simple_and_complex_cells_like_receptive_fields_from_natural}
Masquelier, T., Serre, T., Thorpe, S., and Poggio, T. (2010).
\newblock Learning simple and complex cells-like receptive fields from natural images: a plausibility proof.
\newblock {\em Journal of Vision - J VISION}, 7:81--81.

\bibitem[McIsaac and Eich, 2002]{mcisaac2002vantage}
McIsaac, H.~K. and Eich, E. (2002).
\newblock Vantage point in episodic memory.
\newblock {\em Psychonomic bulletin \& review}, 9(1):146--150.

\bibitem[M{\'e}ly and Serre, 2017]{Mély2017}
M{\'e}ly, D.~A. and Serre, T. (2017).
\newblock {\em Towards a Theory of Computation in the Visual Cortex}, pages 59--84.
\newblock Springer Singapore, Singapore.

\bibitem[Mikolov et~al., 2013]{MikolovSCCD13}
Mikolov, T., Sutskever, I., Chen, K., Corrado, G.~S., and Dean, J. (2013).
\newblock Distributed representations of words and phrases and their compositionality.
\newblock In Burges, C., Bottou, L., Welling, M., Ghahramani, Z., and Weinberger, K., editors, {\em Advances in Neural Information Processing Systems}, volume~26. Curran Associates, Inc.

\bibitem[Müller et~al., 2024]{Do_Humans_and_Convolutional_Neural_Networks_Attend_to_Similar_Areas_during_Scene_Classification}
Müller, R., Dürschmidt, M., Ullrich, J., Knoll, C., Weber, S., and Seitz, S. (2024).
\newblock Do humans and convolutional neural networks attend to similar areas during scene classification: Effects of task and image type.
\newblock {\em Applied Sciences}, 14(6):2648.

\bibitem[Oosterhof et~al., 2012]{10.1162/jocn_a_00195}
Oosterhof, N.~N., Tipper, S.~P., and Downing, P.~E. (2012).
\newblock Viewpoint (in)dependence of action representations: An mvpa study.
\newblock {\em Journal of Cognitive Neuroscience}, 24(4):975--989.

\bibitem[Orban et~al., 2021]{PMID:33745819}
Orban, G., Lanzilotto, M., and Bonini, L. (2021).
\newblock From observed action identity to social affordances.
\newblock {\em Trends in cognitive sciences}, 25(6):493—505.

\bibitem[Qi et~al., 2019]{8954364_amodal_KINS_dataset}
Qi, L., Jiang, L., Liu, S., Shen, X., and Jia, J. (2019).
\newblock Amodal instance segmentation with kins dataset.
\newblock In {\em 2019 IEEE/CVF Conference on Computer Vision and Pattern Recognition (CVPR)}, pages 3009--3018.

\bibitem[Radford et~al., 2021]{clip}
Radford, A., Kim, J.~W., Hallacy, C., Ramesh, A., Goh, G., Agarwal, S., Sastry, G., Askell, A., Mishkin, P., Clark, J., Krueger, G., and Sutskever, I. (2021).
\newblock Learning transferable visual models from natural language supervision.
\newblock {\em CoRR}, abs/2103.00020.

\bibitem[Rahmaniboldaji et~al., 2024]{dear}
Rahmaniboldaji, S., Rybansky, F., Vuong, Q., Guerin, F., and Gilbert, A. (2024).
\newblock Dear: Depth-enhanced action recognition.
\newblock {\em European Conference of Computer Vision 2024 WS}.

\bibitem[Rahmaniboldaji et~al., 2025]{humanvsmachine}
Rahmaniboldaji, S., Rybansky, F., Vuong, Q., Guerin, F., and Gilbert, A. (2025).
\newblock Human vs. machine minds: Ego-centric action recognition compared.
\newblock In {\em 2025 IEEE/CVF Conference on Computer Vision and Pattern Recognition Workshops (CVPRW)}, pages 2943--2953.

\bibitem[Ravi et~al., 2024]{ravi2024sam2}
Ravi, N., Gabeur, V., Hu, Y.-T., Hu, R., Ryali, C., Ma, T., Khedr, H., R{\"a}dle, R., Rolland, C., Gustafson, L., Mintun, E., Pan, J., Alwala, K.~V., Carion, N., Wu, C.-Y., Girshick, R., Doll{\'a}r, P., and Feichtenhofer, C. (2024).
\newblock Sam 2: Segment anything in images and videos.
\newblock {\em arXiv preprint arXiv:2408.00714}.

\bibitem[Reimers and Gurevych, 2019]{abs-1908-10084}
Reimers, N. and Gurevych, I. (2019).
\newblock Sentence-{BERT}: Sentence embeddings using {S}iamese {BERT}-networks.
\newblock In Inui, K., Jiang, J., Ng, V., and Wan, X., editors, {\em Proceedings of the 2019 Conference on Empirical Methods in Natural Language Processing and the 9th International Joint Conference on Natural Language Processing (EMNLP-IJCNLP)}, pages 3982--3992, Hong Kong, China. Association for Computational Linguistics.

\bibitem[Riesenhuber and Poggio, 1999]{hmax}
Riesenhuber, M. and Poggio, T. (1999).
\newblock Hierarchical models of object recognition in cortex.
\newblock {\em Nature Neuroscience}, 2(11):1019 – 1025.
\newblock Cited by: 2565.

\bibitem[Rizzolatti et~al., 2001]{rizzolatti2001neurophysiological}
Rizzolatti, G., Fogassi, L., and Gallese, V. (2001).
\newblock Neurophysiological mechanisms underlying the understanding and imitation of action.
\newblock {\em Nature Reviews Neuroscience}, 2(9):661--670.

\bibitem[Roche and Chainay, 2013]{roche2013visually}
Roche, K. and Chainay, H. (2013).
\newblock Visually guided grasping of common objects: effects of priming.
\newblock {\em Visual Cognition}, 21(8):1010--1032.

\bibitem[Roe, 2004]{Roe2004}
Roe, A.~W. (2004).
\newblock Modular complexity of area v2 in the macaque monkey.
\newblock {\em Annual Review of Neuroscience}, 27:237--260.

\bibitem[Roe and Ts'o, 1995]{Roe1995}
Roe, A.~W. and Ts'o, D.~Y. (1995).
\newblock Visual topography in primate v4: multiple maps of visual space.
\newblock {\em Journal of Neuroscience}, 15(5):3689--3715.

\bibitem[Rybansky et~al., 2026]{Rybansky2026.02.15.705896}
Rybansky, F., Rahmaniboldaji, S., Gilbert, A., Guerin, F., Hurlbert, A.~C., and Vuong, Q.~C. (2026).
\newblock Seeing just enough: The contribution of hands, objects and visual features to egocentric action recognition.
\newblock {\em bioRxiv}.

\bibitem[Serre, 2006]{10.5555/1195185}
Serre, T. (2006).
\newblock {\em Learning a dictionary of shape-components in visual cortex: comparison with neurons, humans and machines}.
\newblock PhD thesis, Massachusetts Institute of Technology, USA.

\bibitem[Serre et~al., 2007]{A_feedforward_architecture_accounts_for_rapid_categorization}
Serre, T., Oliva, A., and Poggio, T. (2007).
\newblock A feedforward architecture accounts for rapid categorization.
\newblock {\em Proceedings of the National Academy of Sciences}, 104(15):6424--6429.

\bibitem[Shmuelof and Zohary, 2005]{SHMUELOF2005457}
Shmuelof, L. and Zohary, E. (2005).
\newblock Dissociation between ventral and dorsal fmri activation during object and action recognition.
\newblock {\em Neuron}, 47(3):457--470.

\bibitem[Simonyan and Zisserman, 2014]{DBLP:journals/corr/SimonyanZ14}
Simonyan, K. and Zisserman, A. (2014).
\newblock Two-stream convolutional networks for action recognition in videos.
\newblock In {\em Proceedings of the 28th International Conference on Neural Information Processing Systems - Volume 1}, NIPS'14, page 568–576, Cambridge, MA, USA. MIT Press.

\bibitem[Simonyan and Zisserman, 2015]{vgg}
Simonyan, K. and Zisserman, A. (2015).
\newblock Very deep convolutional networks for large-scale image recognition.
\newblock In {\em 3rd International Conference on Learning Representations (ICLR 2015)}, pages 1--14. Computational and Biological Learning Society.

\bibitem[Soomro et~al., 2012]{ucf101}
Soomro, K., Zamir, A.~R., and Shah, M. (2012).
\newblock {UCF101:} {A} dataset of 101 human actions classes from videos in the wild.
\newblock {\em CoRR}, abs/1212.0402.

\bibitem[Srivastava et~al., 2019]{Minimal_Images_in_Deep_Neural_Networks_Fragile_object_recognition_in_Natural_Images}
Srivastava, S., Ben{-}Yosef, G., and Boix, X. (2019).
\newblock Minimal images in deep neural networks: Fragile object recognition in natural images.
\newblock {\em CoRR}, abs/1902.03227.

\bibitem[Torralba et~al., 2006]{torralba2006contextual}
Torralba, A., Oliva, A., Castelhano, M.~S., and Henderson, J.~M. (2006).
\newblock Contextual guidance of eye movements and attention in real-world scenes: the role of global features in object search.
\newblock {\em Psychological review}, 113(4):766.

\bibitem[Vaswani et~al., 2017]{transformer}
Vaswani, A., Shazeer, N., Parmar, N., Uszkoreit, J., Jones, L., Gomez, A.~N., Kaiser, L.~u., and Polosukhin, I. (2017).
\newblock Attention is all you need.
\newblock In Guyon, I., Luxburg, U.~V., Bengio, S., Wallach, H., Fergus, R., Vishwanathan, S., and Garnett, R., editors, {\em Advances in Neural Information Processing Systems}, volume~30. Curran Associates, Inc.

\bibitem[Vuong and Tarr, 2004]{VUONG20041717_Rotation_direction_affects_object}
Vuong, Q.~C. and Tarr, M.~J. (2004).
\newblock Rotation direction affects object recognition.
\newblock {\em Vision Research}, 44(14):1717--1730.

\bibitem[Wichmann and Geirhos, 2023]{031739}
Wichmann, F.~A. and Geirhos, R. (2023).
\newblock Are deep neural networks adequate behavioral models of human visual perception?
\newblock {\em Annual Review of Vision Science}, 9(Volume 9, 2023):501--524.

\bibitem[Yamins and DiCarlo, 2016]{Yamins2016}
Yamins, D.~L. and DiCarlo, J.~J. (2016).
\newblock Using goal-driven deep learning models to understand sensory cortex.
\newblock {\em Nature Neuroscience}, 19(3):356--365.

\bibitem[Yamins et~al., 2014]{doi:10.1073/pnas.1403112111}
Yamins, D. L.~K., Hong, H., Cadieu, C.~F., Solomon, E.~A., Seibert, D., and DiCarlo, J.~J. (2014).
\newblock Performance-optimized hierarchical models predict neural responses in higher visual cortex.
\newblock {\em Proceedings of the National Academy of Sciences}, 111(23):8619--8624.

\bibitem[Yao et~al., 2023]{side4video}
Yao, H., Wu, W., and Li, Z. (2023).
\newblock Side4video: Spatial-temporal side network for memory-efficient image-to-video transfer learning.
\newblock {\em arXiv preprint arXiv:2311.15769}.

\bibitem[Zhang et~al., 2025]{videollama3}
Zhang, B., Li, K., Zesen~Cheng, Z.~H., Yuan, Y., Chen, G., Leng, S., Jiang, Y., Zhang, H., Li, X., Jin, P., Zhang, W., Wang, F., Bing, L., and Zhao, D. (2025).
\newblock Videollama 3: Frontier multimodal foundation models for image and video understanding.
\newblock {\em arXiv preprint arXiv:2501.13106}.

\bibitem[Zhu et~al., 2017]{zhu2017semantic_amodal_segmentation}
Zhu, Y., Tian, Y., Mexatas, D., and Dollár, P. (2017).
\newblock Semantic amodal segmentation.
\newblock In {\em Conference on Computer Vision and Pattern Recognition (CVPR)}, pages 4542--4550.

\end{thebibliography}
